# Generative Adversarial Networks (GANs): Challenges, Solutions, and Future Directions

Divya Saxena

University Research Facility in Big Data Analytics (UBDA), The Hong Kong Polytechnic University, Hong Kong, divya.saxena.2015@ieee.org

Jiannong Cao

Department of Computing and UBDA, The Hong Kong Polytechnic University, Hong Kong, csjcao@comp.polyu.edu.hk

### ABSTRACT

Generative Adversarial Networks (GANs) is a novel class of deep generative models which has recently gained significant attention. GANs learns complex and high-dimensional distributions implicitly over images, audio, and data. However, there exist major challenges in training of GANs, i.e., mode collapse, non-convergence and instability, due to inappropriate design of network architecture, use of objective function and selection of optimization algorithm. Recently, to address these challenges, several solutions for better design and optimization of GANs have been investigated based on techniques of re-engineered network architectures, new objective functions and alternative optimization algorithms. To the best of our knowledge, there is no existing survey that has particularly focused on the broad and systematic developments of these solutions. In this study, we perform a comprehensive survey of the advancements in GANs design and optimization solutions proposed to handle GANs challenges. We first identify key research issues within each design and optimization technique and then propose a new taxonomy to structure solutions by key research issues. In accordance with the taxonomy, we provide a detailed discussion on different GANs variants proposed within each solution and their relationships. Finally, based on the insights gained, we present promising research directions in this rapidly growing field.

## 1. INTRODUCTION

Deep generative models (DGMs), such as Restricted Boltzmann Machines (RBMs), Deep Belief Networks (DBNs), Deep Boltzmann Machines (DBMs), Denoising Autoencoder (DAE), and Generative Stochastic Network (GSN), have recently drawn significant attention for capturing rich underlying distributions of the data, such as audio, images or video and synthesize new samples. These deep generative models are modelled by Markov chain Monte Carlo (MCMC) based algorithms [1][2]. MCMC-based approaches calculate the gradient of log-likelihood where gradients vanish during training advances. This is the major reason that sample generation from the Markov Chain is slow as it could not mix between modes fast enough. Another generative model, variational autoencoder (VAE), uses deep learning with statistical inference for representing a data point in a latent space [3] and experiences the complexity in the approximation of intractable probabilistic computations. In addition, these generative models are trained by maximizing training data likelihood where likelihood-based methods go through the curse of dimensionality in many datasets, such as image, video. Moreover, sampling from the Markov Chain in high-dimensional spaces is blurry, computationally slow and inaccurate.

To handle the abovementioned issues, Goodfellow, et al. [4] proposed Generative Adversarial Nets (GANs), an alternative training methodology to generative models. GANs is a novel class of deep generative models in which backpropagation is used for training to evade the issues associated with MCMC training. GANs training is a minimax zero-sum game between a generative model and a discriminative model. GANs has gained a lot of attention recently for generating realistic images as it avoids the difficulty related to maximum likelihood learning [5]. Figure 1 shows an example of progress in GANs capabilities from year 2014 to 2018.

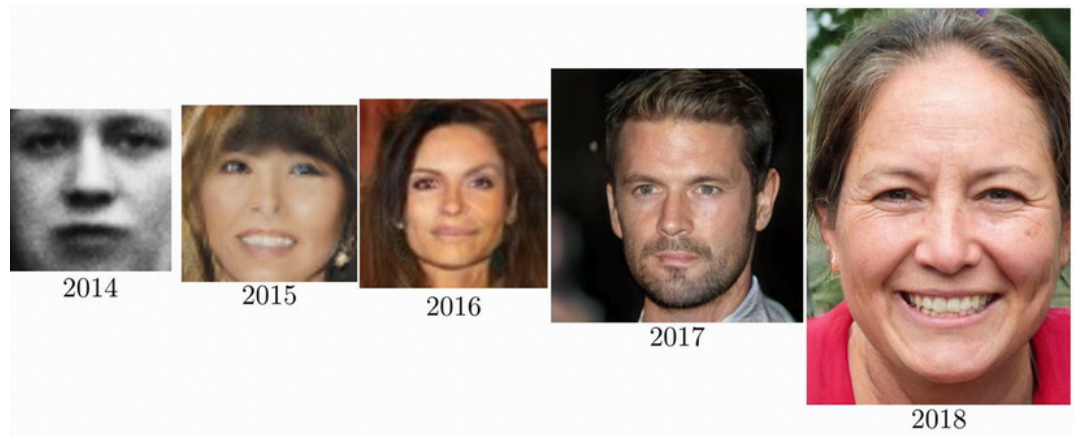

**Figure 1. Progress in the GANs capabilities for image generation from year 2014 to 2018. Figure from [4][6][7][8][9]**

GANs are a structured probabilistic model which comprises of two adversarial models: *a generative model*, called Generator (G) for capturing the data distribution and *a discriminative model*, called Discriminator (D) for estimating the probability to find whether a data generated is from the real data distribution or generated by G's distribution. A two-player minimax game is played by D and G until Nash equilibrium using a gradient-based optimization technique (Simultaneous Gradient Descent), i.e., G can generate images like sampled from the true distribution, and D cannot differentiate between the two sets of images. To update G and D, gradient signals are received from the loss induced by calculating divergences between two distributions by D. We can say that the three main GANs design and optimization components are as follows: (i) network architecture, (ii) objective (loss) function, and (iii) optimization algorithm.

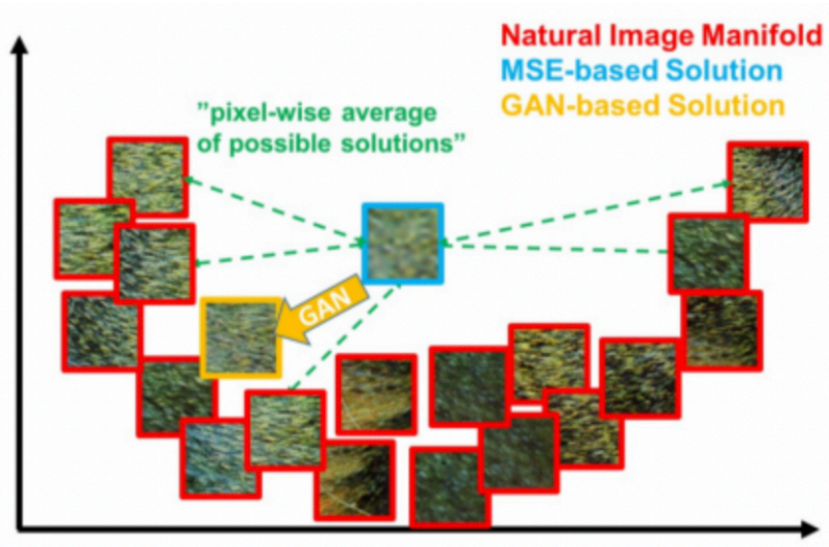

**Figure 2. Patches from the natural image manifold (red) and super-resolved patches obtained with MSE (blue) and GANs (Yellow). Figure from [10]**

For a task that models multi-modal data, a particular input can be related to several different correct and acceptable answers. Figure 2 shows an illustration having several natural image manifolds (in red color), result achieved by basic machine learning model using mean squared error (MSE), which computes pixel-wise average over numerous a little bit different possible answers in the pixel space (i.e., causes the blurry image) and result achieved by GANs which drives reconstruction towards the natural image manifolds. Because of this advantage, GANs has been gaining huge attention and the applicability of GANs is growing in many fields.

GANs has worked well on several realistic tasks, such as image generation [8][9], video generation [11], domain adaptation [12], and image super-resolution [10], etc. Despite its success in many applications, traditional GANs is highly unstable in training because of the unbalanced D and G training. D utilizes a logistic loss which saturates quickly. In addition, if D can easily differentiate between real and fake images, D's gradient vanishes and when D cannot provide gradient, G stops updating. In recent times, many improvements have been introduced for handling the mode collapse problem as G produces samples based on few modes rather than the whole data space. On the other hand, several objective (loss) functions have been introduced to minimize a divergence different from the traditional GANs formulation. Further, several solutions have been proposed to stabilize the training.

## 1.1. Motivation and Contributions

In recent times, GANs has achieved outstanding performance in producing natural images. However, there exist major challenges in training of GANs, i.e., mode collapse, non-convergence and instability, due to inappropriate design of network architecture, use of objective function and selection of optimization algorithm. Recently, to address these challenges, several solutions for better design and optimization of GANs have been investigated based on techniques of re-engineered network architectures, new objective functions and alternative optimization algorithms. To study GANs design and optimization solutions proposed to handle GANs challenges in contiguous and coherent way, this survey proposes a novel taxonomy of different GANs solutions. We define taxonomic classes and sub-classes addressing to structure the current works in the most promising GANs research areas. By classifying proposed GANs design and optimization solutions into different categories, we analyze and discuss them in a systematic way. We also outline major open issues that can be pursued by researchers further.

There are a limited number of existing reviews on the topic of GANs. [13] discussed how GANs and state-of-the-art GANs works. [14]–[16] provided a brief introduction of some of the GANs models, while [16] also presented development trends of GANs, and relation of GANs with parallel intelligence. [17] reviewed various GANs methods from the perspectives of algorithms, theory, and applications. [18] categorized GANs models into six fronts, such as architecture, loss, evaluation metric, Normalization and Constraint, image-to-image translation, conditional techniques and discussed them in brief. [19] presented a summary of GANs models addressing the GANs challenges. On the other hand, several researchers reviewed specific topics related to GANs in detail. [20] reviewed GANs based image synthesis and editing approaches. [21] surveyed threat of adversarial attacks on deep learning. [22] discussed various types of adversarial attacks and defenses in detail.

Despite reviewing the state-of-the-art GANs, none of these surveys, to the best of our knowledge, has particularly focused on broad and systematic view of the GANs developments introduced to address the GANs challenges. In this study, our main aim is to comprehensively structure and summarize different GANs design and optimization solutions proposed to alleviate GANs challenges for the researchers that are new to this field.

**Our Contributions.** Our paper makes notable contributions summarized as follows:

**New taxonomy.** In this study, we identify key research issues within each design and optimization technique and present a novel taxonomy to structure solutions by key research issues. Our proposed taxonomy will facilitate researchers to enhance the understanding of the current developments handling GANs challenges and future research directions.

**Comprehensive survey.** In accordance with the taxonomy, we provide a comprehensive review of different solutions proposed to handle the major GANs challenges. For each type of solution, we provide detailed descriptions and systematic analysis of the GANs variants and their relationships. But still, due to the wide range of GANs applications, different GANs variants are formulated, trained, and evaluated in heterogenous ways and direct comparison among these GANs is complicated. Therefore, we make a necessary comparison and summarize the corresponding approaches w.r.to their novel solutions to address GANs challenges. We provide a detailed investigation into the numerous research domains where GANs has been broadly explored. This survey can be used as a guide for understanding, using, and developing different GANs approaches for various real-life applications.

**Future directions.** This survey also highlights the most promising future research directions.

## 1.2. Organization

In this paper, we first discuss three main components for designing and training GANs framework, analyze challenges with GANs framework, and present a detailed understanding of the current developments handling GANs challenges from the GANs design and optimization perspective.

Figure 3 shows the organization of the paper. Section 2 explains the GANs framework from the designing and training perspective. In Section 3, we present the challenges in the training of GANs. In Section 4, we identify key issues related to the design and training of GANs and present a novel taxonomy of GANs solutions handling these key issues. In accordance with the taxonomy, Section 5, 6 and 7 summarizes GANs design and optimization solutions, their pros and cons, and relationships. Section 8 summarizes the challenges addressed by different GANs design solutions. Section 9 discusses the research domains where GANs has been widely explored. Section 10 discusses the future directions and Section 11 summarizes the paper.

# 2. GENERATIVE ADVERSARIAL NETWORKS

Before discussing in detail about solutions for better design and optimization of GANs in the proposed taxonomy, in this section, we will provide an overview of GANs framework and main GANs design and optimization components.

## 2.1. Overview

In recent years, generative models are continuously growing and have been applied well for a broad range of real applications. Generative models' compute the density estimation where model distribution $p_{model}$ is learned to approximate the true and new data distribution $p_{data}$. Methods to compute the density estimation have two major concerns: selection of suitable objective (loss) function and appropriate selection of formulation for the density function of $p_{model}$. The selection of objective functions for generative model's training plays an important role for better learning behaviors and performance [23][24]. The de-facto standard of the most widely used objective is based on the maximum likelihood estimation theory in which model parameters maximize the training data likelihood.

Researchers have shown that maximum likelihood is not a good option as training objective because a model trained using maximum likelihood mostly overgeneralise and generate unplausible samples [23]. In addition, marginal likelihood is intractable which requires a solution to overcome this for learning the model parameters. One possible solution to handle the marginal likelihood intractability issue is not to compute it ever and learn model parameters via a tool indirectly [25].

GANs achieves this by having a powerful D which have a capability to discriminate samples from $p_{data}$ and $p_{model}$. When D is unable to discriminate samples from $p_{data}$ and $p_{model}$, then model has learned to generate samples similar to the samples from the real data. A possible solution for formulating density function of $p_{model}$ is to use an *explicit density function* in which maximum likelihood framework is followed for estimating the parameters. Another possible solution is to use an implicit density function for estimating the data distribution excluding analytical forms of $p_{model}$, i.e., train a G where if real and generated data are mapped to the feature space, they are enclosed in the same sphere [23][24]. However, GANs is the most notably pioneered class of this possible solution.

GANs is an expressive class of generative models as it supports exact sampling and approximate estimation. GANs learns high-dimensional distributions implicitly over images, audio, and data which are challenging to model with an explicit likelihood. Basic GANs are algorithmic architectures of two neural networks competing with each other to capture the real data distribution. Both neural nets try to optimize different and opposing objective (loss) function in the zero-sum game to find (global) the Nash equilibrium. The three main components for design and optimization of GANs are: (i) network architecture, (ii) objective (loss) function, and (iii) optimization algorithm. There has been a large amount of works towards improving GANs by re-engineering architecture [5][6][25], better objective functions [29]–[31], and alternative optimization algorithms [29][30].

In the following sections, we shall discuss three main components for the GANs design and optimization, namely network architecture, loss function and the optimization algorithm followed by the minimax optimization for Nash equilibrium in detail.

## 2.2. Network Architecture

GANs learns to map the simple latent distribution to the more complex data distribution. To capture the complex data distribution $p_{data}$, GANs architecture should have enough capacity. GANs is based on the concept of a non-cooperative game of two networks, a generator G and a discriminator D, in which G and D play against each other. GANs can be part of deep generative models or generative neural models where G and D are parameterized via neural networks and updates are made in parameter space.

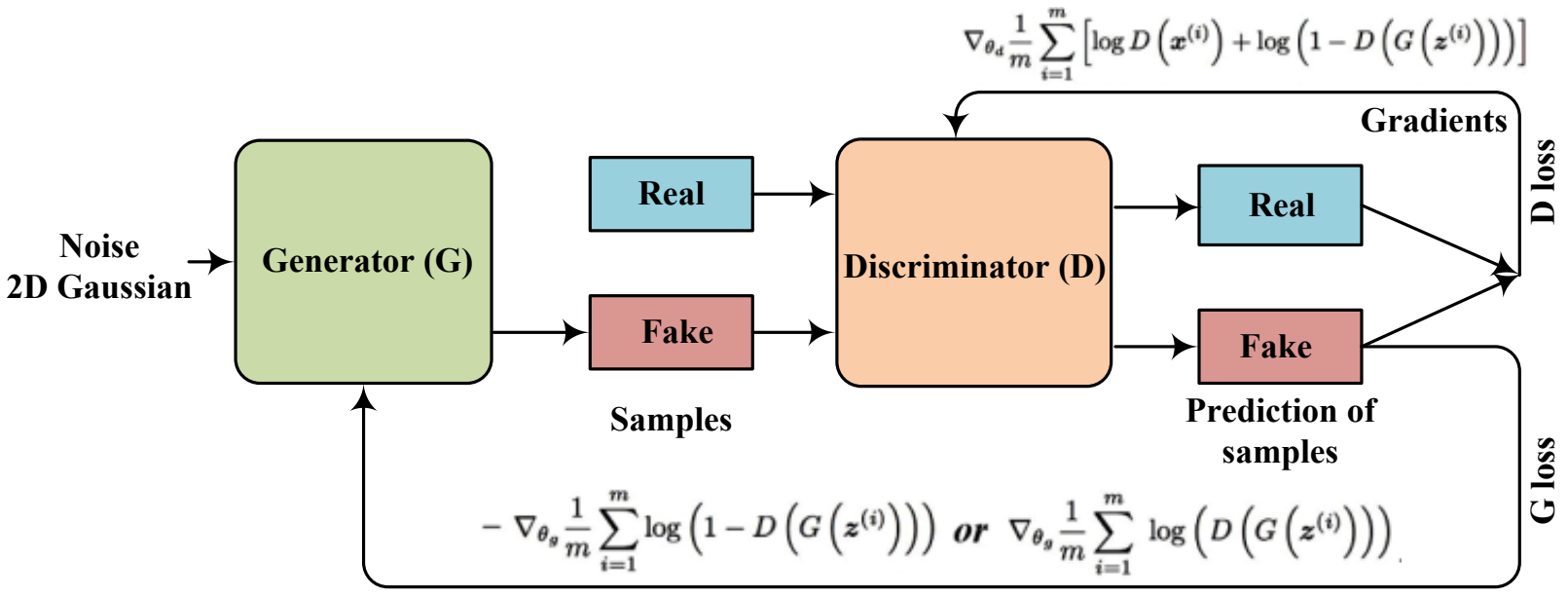

**Figure 3. Basic GANs Architecture**

Both G and D play a minimax game where G's main aim is to produce samples similar to the samples produced from real data distribution and D's main goal is to discriminate the samples generated by G and samples generated from the real data distribution by assigning higher and lower probabilities to samples from real data and generated by G, respectively. On the other hand, the main target of GANs training is to keep moving the generated samples in the direction of the real data manifolds through the use of the gradient information from D.

In GANs, $x$ is data extracted from the real data distribution, $p_{data}$, noise vector $z$ is taken from a Gaussian prior distribution with zero-mean and unit variance $p_z$, while $p_g$ refers the G's distribution over data $x$. Latent vector $z$ is passed to G as an input and then G outputs an image G($z$) with the aim that D cannot differentiate between G($z$) and D($x$) data samples, i.e., G($z$) resembles with D($x$) as close as possible. In addition, D simultaneously tries to restrain itself from getting fooled by G. D is a classifier where D($x$) = 1 if $x \sim p_{data}$ and D($x$) = 0 if $x \sim p_g$, i.e., $x$ is from $p_{data}$ or from $p_g$.

The basic GANs architecture for the above discussion is given in Figure 3. For the given basic GANs architecture, the game setup between D and G during training is discussed below.

## 2.3. Objective (Loss) Function

Objective function tries to match real data distribution $p_{data}$ with $p_g$. Basic GANs use two objective functions: (1) D minimizes the negative log-likelihood for binary classification; (2) G maximizes the probability of generated samples for being real. D parameters are denoted by $\theta_D$, which are trained to maximize the loss to distinguish between the real and fake samples. G parameters are denoted by $\theta_G$ which are optimized such that the D is not able to distinguish between real and fake samples generated by G. $\theta_G$ is trained to minimize the same loss that $\theta_D$ is maximizing. Hence, it is a zero-sum game where players compete with each other. The following minimax objective applied for training G and D models jointly via solving:

$$\min_{\theta_G} \max_{\theta_D} V(G, D) = \min_{G} \max_{D} \mathbb{E}_{x \sim p_{data}}[\log D(x)] + \mathbb{E}_{z \sim p_z}\left[\log\left(1 - D\big(G(z)\big)\right)\right] \quad (1)$$

$V(G, D)$ is a binary cross entropy function, commonly used in binary classification problems [34]. In Eq. 1, for updating the model parameters, training of G and D are performed by backpropagating the loss via their respective models. In practice, Eq. 1 is solved by alternating the following two gradient updates:

$$\theta_D^{t+1} = \theta_D^t + \lambda^t \nabla_{\theta_D} V(D^t, G^t) \text{ and } \theta_G^{t+1} = \theta_G^t + \lambda^t \nabla_{\theta_G} V(D^{t+1}, G^t)$$

where $\theta_G$ is the parameter of G, $\theta_D$ is the parameter D, $\lambda$ is the learning rate, and $t$ is the iteration number.

In practice, second term in Eq. 1, $\log\left(1 - D\left(G(z)\right)\right)$ saturates and makes insufficient gradient flow through G, i.e., gradients value gets smaller and stop learning. To overcome the vanishing gradient problem, the objective function in Eq. 1 is reframed into two separate objectives:

$$\max_{\theta_D} \mathbb{E}_{x \sim p_{data}}[\log D(x)] + \mathbb{E}_{z \sim p_z}\left[\log\left(1 - D\left(G(z)\right)\right)\right] \text{ and } \max_{\theta_G} \mathbb{E}_{z \sim p_z}\left[\log\left(D\left(G(z)\right)\right)\right] \tag{2}$$

Moreover, G's gradient for these two separate objectives have the same fixed points and are always trained in the same direction. After cost computation in Eq. 2, Backpropagation can be used for updating the model parameters. Because of these two different objectives, the update rule is given as:

$$\{\theta_D^{t+1}, \theta_G^{t+1}\} \leftarrow \begin{cases} \text{Update} & \text{if } D(x) \text{ predicts wrong} \\ \text{Update} & \text{if } D\left(G(z)\right) \text{ predicts wrong} \\ \text{Update} & \text{if } D\left(G(z)\right) \text{ predicts correct} \end{cases}$$

If D and G are given enough capability with sufficient training iterations, G can convert a simple latent distribution $p_g$ to more complex distributions, i.e., $p_g$ converges to $p_{data}$, such as $p_g = p_{data}$.

## 2.4. Optimization Algorithm

In GANs, optimization is to find (global) equilibrium of the minmax game, i.e., saddle point of min-max objective. Gradient-based optimization methods are widely used to find the local optima for classical minimization and saddle point problems. Any traditional gradient-based optimization technique can be used for minimizing each player's cost simultaneously which leads to Nash equilibrium. Basic GANs uses the *Simultaneous Gradient Descent* for finding Nash-equilibria and update D's and G's parameters by simultaneously using gradient descent on D's and G's utility functions. Each player D and G tries to minimize its own cost/objective function for finding a Nash equilibrium $(\theta_D, \theta_G)$, $J_D(\theta_D, \theta_G)$ for the D and $J_G(\theta_D, \theta_G)$ for the G, i.e., $J_D$ is at a minimum w.r.to $\theta_D$ and $J_G$ is at a minimum w.r.to $\theta_G$.

**Minimax optimization for Nash equilibrium.** The total cost of all players in a zero-sum game is always zero. In addition, a zero-sum game is also known as a minimax game because solution includes minimization and maximization in an outer and inner loop, respectively. Therefore,

$$J_G + J_D = 0$$

i.e.,

$$J_G = -J_D$$

All the GANs game designed earlier, apply the same cost function to the D, $J_D$, but they vary in G's cost, $J_G$. In these cases, D uses the same optimal strategy. D's cost function is as follows w.r.to $\theta_D$:

$$J_D(\theta_D, \theta_G) = -\frac{1}{2}\mathbb{E}_{x \sim p_{data}}[\log D(x)] - \frac{1}{2}\mathbb{E}_z\left[\log\left(1 - D\left(G(z)\right)\right)\right] \tag{3}$$

Eq. 3 represents the standard cross-entropy cost which is minimized during the training of a standard binary classifier with a sigmoid output. In the minimax game, G attempts to minimize and maximize the log-probability of D being correct and being mistaken, respectively. In GANs, training of a classifier is performed on two minibatches of data: a real data minibatch having examples' label 1 and another minibatch from G having examples' label 0, i.e., the density ratio between true and generated data distribution is represented as follows:

$$D(x) = \frac{p_{data}(x)}{p_g(x)} = \frac{p(x|y=1)}{p(x|y=0)} = \frac{p(y=1|x)}{p(y=0|x)} = \frac{D^*(x)}{1 - D^*(x)}$$

Here, $y$=0 means generated data and $y$=1 means real data. $p(y=1) = p(y=0)$ is assumed.

Through training D, we get an estimate of the ratio $p_{data}(x)/p_g(x)$ at every point $x$ which allows the computation of divergences and their gradients and sets approximation technique of GANs differ from VAEs and DBMs as they generate lower bounds or Markov Chains based approximations. D learns to distinguish samples from data for any given G. The optimal D for a fixed G is given by $D_G^*(x) = \frac{P_{data}(x)}{P_{data}(x) + P_g(x)}$. After sufficient training steps, G and D with enough capacity will converge to $p_g = p_{data}$, i.e., D cannot discriminate between two distributions. For optimal D, the minimax game in Eq. 1 can now be reformulated as follows.

$$J_D = -2\log 2 + 2\left(D_{JS}(p_{data} \parallel p_g)\right) \tag{4}$$

where $D_{JS}$ denotes Jensen-Shannon divergence. When D is optimal, G minimizes Jensen-Shannon divergence (JSD) mentioned in Eq. 4 which is an alternative similarity measure between two probability distributions, bounded by [0,1]. In the basic GANs, it is feasible to estimate neural samplers through approximate minimization of the symmetric JSD.

$$D_{JS}(p_{data} \parallel p_g) = \frac{1}{2}D_{KL}\left(p_{data} \parallel \frac{1}{2}(p_{data} + p_g)\right) + \frac{1}{2}D_{KL}\left(p_g \parallel \frac{1}{2}(p_{data} + p_g)\right),$$

where $D_{KL}$ denotes the KL divergence.

JSD is based on KL divergence, but it is symmetric, and it always has a finite value. Because $D_{JS}(p_{data} \parallel p_g)$ is an appropriate divergence measure between distributions, i.e., real data distribution $p_{data}$ can be approximated properly when sufficient training samples exist and model class $p_g$ can represent $p_{data}$. The JSD between two distributions is always non-negative and zero when two distributions are equivalent, $J_D^* = -2\log 2$ is the global minimum of $J_D$ which shows $p_g = p_{data}$.

Given the minimax objective, $p_g = p_{data}$ occurs at a saddle point $\theta_G, \theta_D$. Saddle point of a loss function occurs at a point which is minimal w.r.to one set of weights and maximal w.r.to another. GANs training interchanges between minimization and maximization steps. If one step is powerful than another, then solution path "slides off" the loss surface as shown in Figure 4. Due to this, training becomes unstable and network collapses [35].

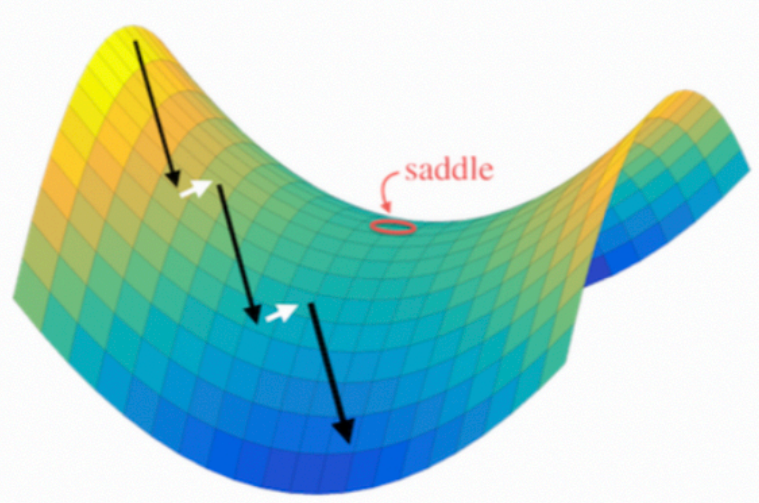

**Figure 4. An illustration of gradient method of Adversarial Net** [35]

Finding Nash equilibrium in GANs is a very challenging problem as objective functions are non-convex, parameters are continuous, and parameter space is high-dimensional [36], e.g., an update to $\theta_D$ that decreases $J_D$ can increase $J_G$, and an update to $\theta_G$ that decreases $J_G$ can increase $J_D$, i.e., training fails to converge. GANs are still hard to train and training suffers from the following problems, such as difficulty converging and instability, and mode collapse. In the next section, we shall discuss major reasons of these problems in the GANs training in detail.

# 3. CHALLENGES IN TRAINING GANs

GANs suffer from the limitation of generating samples with little diversity, even trained on multi-model data. E.g., when GANs is trained on hand-written digits' data with ten modes, G may unable to generate some digits [37]. This condition is identified as mode collapse problem and several recent advances in GANs has focused to resolve this problem.

It may also possible that G and D oscillate during training, instead of a fixed-point convergence. When a player gets more powerful than another player, then it may possible that system does not learn and suffers from the vanishing gradients, i.e., instability. When the generated samples are initially very poor, D learns to differentiate easily between real and fake samples. This causes D(G($z$)), probability of the generated samples being real, will be close to zero, i.e., gradient of log(1 − D(G($z$)) will be very small [38]. This shows that when D fails to provide gradients, G will stop updating. Also, hyperparameters selection, such as batch size, momentum, weight decay, and learning rate is an utmost important factor for GANs training to converge [6].

In this section, we shall discuss about the main challenges in the GANs training in detail.

### 3.1. Mode Collapse

Mode collapse problem can occur as the max-min solution to the GANs works in a different way from the min-max solution. Therefore, in $G^* = \min_{\theta_G} \max_{\theta_D} V(G, D)$, G* generates samples from the data distribution. In case of $G^* = \max_{\theta_D} \min_{\theta_G} V(G, D)$, G maps every $z$ value to the single $x$ coordinate that D believes them real instead of fake. Simultaneous gradient descent does not clearly benefit min max over max min or vice versa.

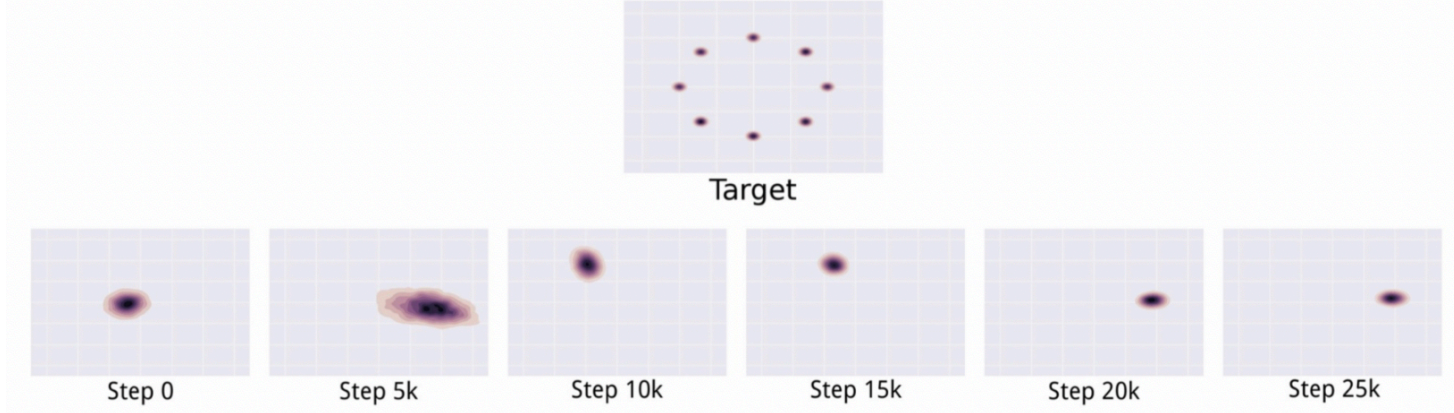

**Figure 5. An example for mode collapse problem on a 2D toy dataset. Target distribution is a mixture of Gaussians in 2D space. During GANs training, G generates only single mode at each time step and keep moving among different modes as D is trained to discard separate modes. Figure from** [39]

Figure 5 shows GANs training on a toy dataset in which G produces only single mode instead of having multi-model training dataset and keeps cycling between different modes as D continuously rejects samples generated by G. This shows GANs oscillates and have difficulty in achieving Nash equilibrium. Mode collapse is one of the key reasons for the cause of unstable GANs training – another big challenge in GANs.

The main drawback of GANs is that they could not focus on the whole data distribution as GANs' objective function has some similarity with the JSD. Experiments have shown that even for the bi-model distribution, minimizing JSD only produces a good fit to the principal mode and could not generate quality images [24].

Figure 6 illustrates the missing mode problem where G's gradient pushes G towards major mode $M_1$ for most $z$ while G's gradient pushes G towards $M_2$ only when G($z$) is very close to mode $M_2$. However, it is likely that in the prior distribution, such $z$ is of low or zero probability. As Figure 6 shows one of the main reasons of mode missing is that the G visits missing modes area rarely, i.e., provides very few examples for improving G around those areas.

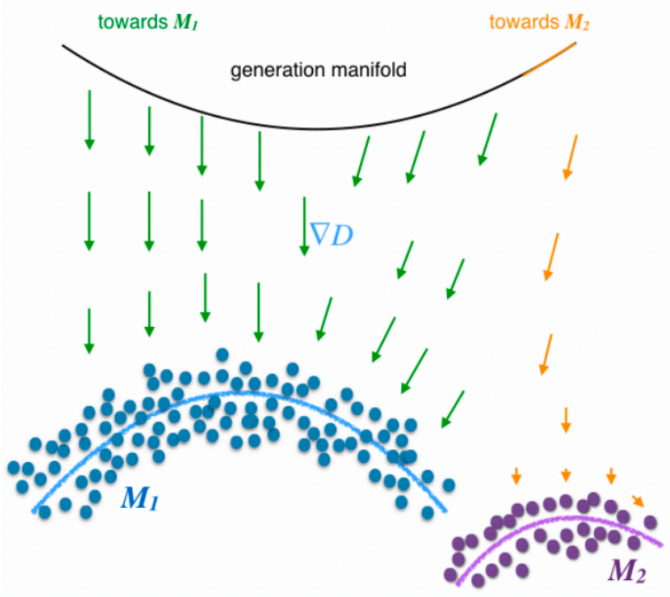

**Figure 6. Illustration of missing mode problem. Figure from** [40]

Generally, mode collapse is a consequence of poor generalization. There can be two types of mode collapse: (1) most of the modes from the input data are absent from the generated data, (2) only a subset of particular modes is learned by G. An ill-suited objective function can be a major reason for the mode collapse problem where several GANs variants, including modifying D's objective [27][30], modifying G's objective [41] have been proposed. In these variants, G is shown, at equilibrium, and able to learn the whole data distribution, but convergence is elusive in practice. To handle this issue, several recent studies have introduced new network architectures with new objective functions or alternative training schemes.

In the next-subsection, we shall discuss another major challenge in the GANs training.

### 3.2. Non-convergence and Instability

In the traditional GANs, G uses two loss function as already discussed: $\mathbb{E}_Z[\log \mathrm{D}(\mathrm{G}(z))]$ and $\mathbb{E}_Z\left[\log\left(1-\mathrm{D}(\mathrm{G}(z))\right)\right]$. But, unfortunately, G's loss can lead to potential issues in GANs training.

The former loss function $\mathbb{E}_Z[\log \mathrm{D}(\mathrm{G}(z))]$ can be the cause of gradient vanishing problem when D can easily differentiate between real and fake samples. For an optimal D, G loss minimization is similar to the minimization of the JSD between real image distribution and generated image distribution. As already discussed, in this case, the JSD will be 2log2. This allows optimal D to assign probability 0 to fake samples, and 1 to real ones and causes gradient of G loss towards 0 which is called vanishing gradients on G. Figure 7 shows that as D gets better and the gradient of G vanishes.

To minimize the cross-entropy between a target class and a classifier's predicted distribution, classifier requires to choose the correct class. In GANs, D tries to minimize a cross-entropy while G tries to maximize the same cross-entropy. When D confidence is quite high, D rejects the samples generated by G, and then G's gradient vanishes. One possible solution to alleviate this issue is to reverse the target employed for constructing the cross-entropy cost. Thus, G's cost is as follows:

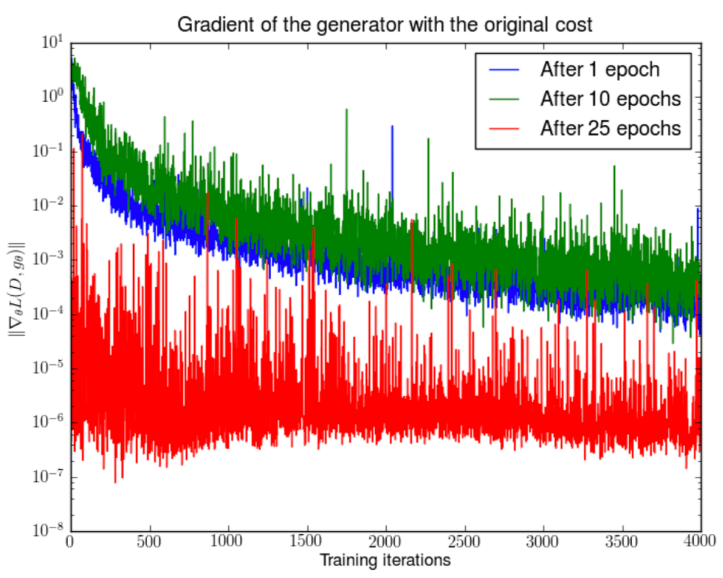

**Figure 7. DCGANs for 1, 10 and 25 epochs. G is fixed while D is trained from scratch and using original cost function to compute the gradients** [42]

**Figure 8. DCGANs for 1, 10 and 25 epochs. G is fixed, D is trained from scratch and using − log D cost function to compute the gradients** [42]

The second one is considered as the −log D trick [4][42][29]. The minimization of the G's loss function $\mathbb{E}_Z\left[\log\left(1-\mathrm{D}(\mathrm{G}(z))\right)\right]$ is equal to minimize $D_{KL}(p_g \parallel p_{data}) - 2\left(D_{JS}(p_{data} \parallel p_g)\right)$ which causes the unstable gradients as it minimizes the KL divergence and maximizes JSD simultaneously. This situation is called instability of G's gradient updates. Figure 8 shows that gradients of the G are growing rapidly. This figure also shows that variance of the gradients growing, i.e., these gradients updates will lead generation of low sample quality.

$$\mathrm{J}_G = -\frac{1}{2}\mathbb{E}_Z[\log D(G(z))], \text{ or}$$

$$\mathrm{J}_G = -\frac{1}{2}\mathbb{E}_Z\left[\log\left(1 - D(G(z))\right)\right],$$

Several GANs design and optimization solutions have been proposed to cope up with the non-convergence and instability problems. We shall discuss key solutions in the subsequent sections.

### 3.3. Evaluation Metrics

GANs model has been used for the wide applications of the unsupervised representation learning, supervised and semi-supervised learning, inpainting, denoising and many more. For these extensive applications, loads of heterogeneity exists in models' formulation, training and evaluation. Despite the availability of lots of GANs model, the evaluation is still qualitative, (i.e., visual examination of samples by human) even though several approaches and measures have been introduced to evaluate GANs performance. Visual inspection is time-consuming, subjective and cannot capture distributional characteristics, which is an utmost important factor for the unsupervised learning. As the selection of appropriate model is important for getting good performance for an application, the selection of appropriate evaluation metric is important for drawing right conclusion. For designing better GANs model, it is required to overcome the limitations of the qualitative measure by developing or using proper quantitative metrics. Recently, multiple GANs evaluation metrics have been introduced with the emergence of new models.

In this paper, we restrict our focus on GANs design and optimization solutions proposed for handling first two GANs challenges, mode collapse, non-convergence and instability. In the next section, we shall discuss the proposed taxonomy of GANs design and optimization solutions proposed to mitigate these challenges and improve the GANs performance.

# 4. A TAXONOMY

Table 1 illustrates our proposed taxonomy of GANs designing and optimization solutions proposed to handle two major GANs training challenges as discussed earlier. In recent times, several solutions (S) have been proposed for better design and optimization of basic GANs based on three main techniques, namely re-engineering network architecture ($S_1$), new loss function ($S_2$), and alternative optimization algorithm ($S_3$). *Re-engineered network architecture* ($S_1$) focuses on the re-engineering GANs network architecture [6][43][8][44][45], *new loss function* ($S_2$) covers modified or new loss functions for GANs [36][38][40][46], while *alternative optimization algorithm* ($S_3$) includes modified or regularized optimization algorithms for GANs [39].

**Table 1 Proposed taxonomy and different variants proposed within the GANs design and optimization techniques**

| Techniques | Solutions (S) | | Variants |
|---|---|---|---|
| **Re-engineered Network Architecture ($S_1$)** | **Conditional generation ($S_{11}$)** | | cGANs [28], FCGAN [47], IRGAN [47], GRANs [48], LAPGAN [5], SGAN [43], IcGAN [49], BiCoGAN [50], MatAN [51], Self-conditioned GANs [52], AC-GANs [53], TripleGAN [54], KDGAN [55], ControlGAN [56] |
| | **Generative-discriminative network pair ($S_{12}$)** | **Training of Single G ($S_{12}(i)$)** | DCGAN [6], ProgressGAN [8], PacGAN [37], BayesianGAN [57], CapsNets [58], QuGANs [59], SAGAN [60] |
| | | **Training of Multiple Gs ($S_{12}(ii)$)** | cGAN [44], AdaGAN [45], MAD-GAN [61], MGAN [62], MPM GAN [63], FictitiousGAN [64], MIX+GAN [65] |
| | | **Training of Multiple Ds ($S_{12}(iii)$)** | D2GAN [66], GMAN [67], StabGAN [68], Dropout GAN [69], MicroBatchGANs [70], SGAN [71] |
| | **Joint Architecture ($S_{13}$)** | **Data space autoencoder ($S_{13}(i)$)** | VAE-GAN [72], AAE [73], AVB [74], ASVAE [75], MDGAN [40], Dist-GAN [76], α-GAN [25] |
| | | **Latent space autoencoder ($S_{13}(ii)$)** | ALI [77], BiGAN [78], DALI [79], CV-BiGAN [80], MV-BiGAN [80], HALI [81], AGE [82], VEEGAN [83], MGGAN [84] |
| | **Improved D ($S_{14}$)** | | EBGAN [85], BEGAN [86], MAGAN [87], [88], Max-Boost-GAN [89] |
| | **Memory Networks ($S_{15}$)** | | MemoryGAN [90] |
| | **Latent space engineering ($S_{16}$)** | | DeLiGAN [91], NEMGAN [92], MultiplicativeNoise [93], DE-GAN [94], InfoGAN [95] |
| **New Loss Function ($S_2$)** | **New probability distance and divergence ($S_{21}$)** | | WGAN [29], LS-GAN [96],RWGAN [31], f-GAN [97], [98], χ2-GAN [99], OT-GAN [100], LSGAN [34], SoftmaxGAN [101], GAN-RL [102], GoGAN [103], IGAN [36], McGAN [104], MMD GAN [46], MMGAN [105], CramerGAN [106] |
| | **Regularization ($S_{22}$)** | | WGAN-GP [107], BWGAN [30], [108] , CT-GAN [109], SN-GAN [110], [111], FisherGAN [112], [32], Unrolled GANs [39], [113], [114], DRAGAN [115] |
| **Alternative Optimization Algorithm ($S_3$)** | | | [32], [116], [117], [33], [35] |

For each technique, we point out number of key research issues and the corresponding solutions (S) to address the GANs challenges.

1. **Re-engineered network architecture ($S_1$).** GANs are hard to train as G could not learn the complex data distribution and generates low variety of samples. Therefore, GANs requires better designs of model architectures. In recent times, several architectural solutions have been proposed to handle GANs challenges in different ways, such as conditional generation, generative-discriminative network pair, using strong discriminator, memory network, and encoder-decoder architecture, engineering noise. We classify existing architectural solutions into six categories where each category represents an issue with the existing architecture and its possible solution to improve the GANs performance. The key research issues and their solutions are as follows:

    a. **Conditional generation ($S_{11}$).** An unconditional generative model cannot control the modes generation. To control the generation process, a generative model can be conditioned on additional information [28]. In this way, conditional GANs learn conditional probability distribution where a condition can be any auxiliary information about the data.

    b. **Generative-discriminative network pair ($S_{12}$).** Intermediate representation of GANs can only handle the generation of smaller images [6]. It requires architectural changes in the layers and networks for improving the generating ability of GANs. But still, a single pair of G and D in minimax game fluctuates and do not converge as discriminatively trained networks do [44]. To handle this issue, multiple Gs and Ds can be trained to increase the generation capacity of Gs and to get more constructive gradient signals for G, respectively.

    c. **Joint architecture ($S_{13}$).** Some research works have proposed to use most common approach to address mode collapse, encoder-decoder architecture in GANs, in which features are learned from the latent space or image space.

    d. **Improved Discriminator ($S_{14}$).** Weak discriminators can also be the reason of mode collapse, because of either low capacity or a poorly-chosen architecture. Therefore, researchers proposed to use autoencoder to define the D's objectives.

    e. **Memory networks ($S_{15}$).** Basic GANs encounters instability and divergence in the unsupervised GANs training due to two key issues, namely the structural discontinuity in a latent space and the forgetting problem of GANs. Basic GANs use unimodal continuous latent distribution to embed multiple classes; hence, structural discontinuity between classes is not clear in the generated samples. In addition, Ds forget about the previously generated samples. To handle these issues, [90] proposed a solution to incorporate a memorization module with unsupervised GAN models.

    f. **Latent-space engineering ($S_{16}$).** Basic GANs cannot generate diverse samples in case of limited data availability. Researchers introduce to reparametrize latent space $z$ like a mixture model and then learn parameters of mixture model with GANs [91].

2. **New loss function ($S_2$).** The model parameters oscillate, destabilize and never converge. A use of good loss function improves GANs learning to reach better optima. Several researchers proposed to tackle the training instability problem by finding better distance measures [29] or regularizers [28][45]. The issues and their solutions are as follows:

    a. **New probability distance and divergence ($S_{21}$).** Basic GANs training is unstable as it uses JS divergence (JSD) which is neither continuous nor provide a usable gradient [29]. There is need of new probability distance and divergence for getting usable gradients everywhere. New probability distance and divergence can solve the mode collapse problem by stabilizing GAN training.

    b. **Regularization ($S_{22}$).** GANs training is unstable when model distribution and data distribution manifolds do not overlap in the high-dimensional space, i.e., dimensionality misspecification [111]. Regularizing D gives an efficient direction to convolve the distributions.

3. **Alternative optimization algorithm ($S_3$).** Basic GANs uses *Simultaneous gradient Descent* for finding Nash-equilibria which often fails to find local Nash-equilibria [32]. Some works have suggested to use another gradient descent optimization technique, while some have suggested modifications to optimization or training technique.

In the following sections (Section 5, 6 and 7), we introduce and compare existing GANs designing and optimization solutions and their variants according to the proposed taxonomy. We review and critically discuss these solutions with their advantages and drawbacks. Furthermore, we provide a supplementary material which contains the loss function(s) used in the re-engineering network architecture taxonomic class and a comparison among new loss functions proposed for GANs. Table 1 lists the different variants proposed within the GANs design and optimization solutions.

# 5. RE-ENGINEERED NETWORK ARCHITECTURE

As we have already discussed in Section 4, several network architectures have been proposed to scale up the GANs performance, such as hierarchical GANs structure, and engineering on the latent space for generating high quality samples for the limited data, use of autoencoders for D for more stable training, and combination of any of these structures.

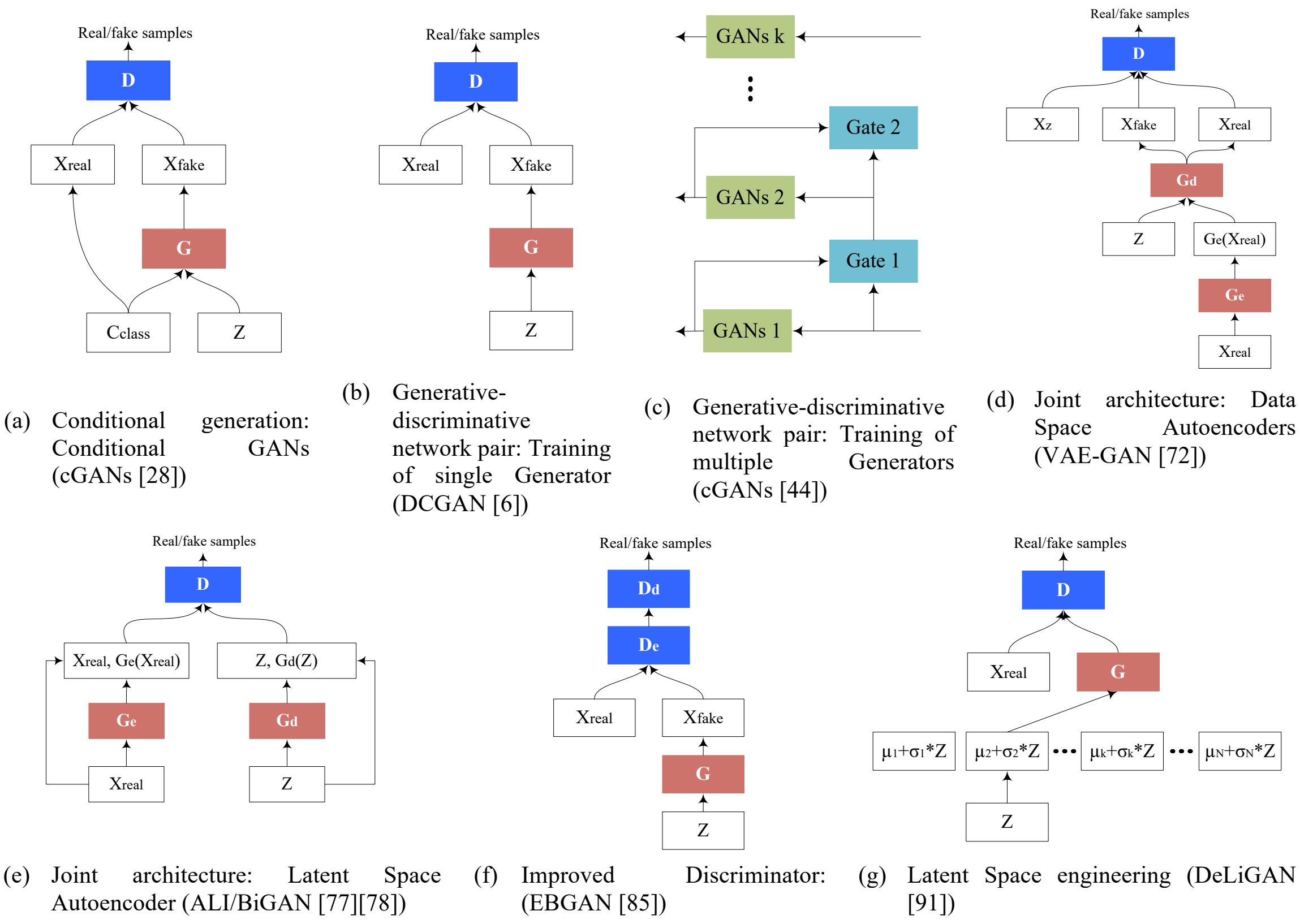

(a) Conditional generation: Conditional GANs (cGANs [28])

(b) Generative-discriminative network pair: Training of single Generator (DCGAN [6])

(c) Generative-discriminative network pair: Training of multiple Generators (cGANs [44])

(d) Joint architecture: Data Space Autoencoders (VAE-GAN [72])

(e) Joint architecture: Latent Space Autoencoder (ALI/BiGAN [77][78])

(f) Improved Discriminator: (EBGAN [85])

(g) Latent Space engineering (DeLiGAN [91])

D: Discriminator, G: Generator, Z: Latent (noise) space, Xreal: Real data, Xfake: Fake data generated by G, Cclass: an auxiliary information, Ge: Encoder for G, Gd: Decoder for G, De: Encoder for D, Dd: Decoder for D.

**Figure 9. Schematic view of most representative GANs variants within the taxonomic class of network architecture GANs**

Figure 9 shows the schematic illustrations of the most representative solutions within the taxonomic class of re-engineered network architecture. In conditional generation sub-class, conditional GANs use an auxiliary information on G and/or D during the generation to control the mode generation (see Figure 9 (a)). In the sub-class of generative-discriminative network pair, single and multiple pair of G and D are trained for better

GANs performance. DCGANs uses fully convolutional downsampling/upsampling layers instead of Fully connected layers used in traditional GANs (see Figure 9 (b)), while cascade of GANs (cGANs) consists of multiple GANs and gates where each GANs redirects badly modelled part of training data to next GANs for capturing whole distribution of the data (see Figure 9 (c)). In addition, within the sub-class of joint architecture, VAE-GAN (see Figure 9 (d)) and ALI/BiGAN (see Figure 9 (e)) have introduced an efficient inference mechanism which includes features from the latent space and data space, respectively. In the sub-class of improved discriminator, energy-based GANs replaced encoder architecture for D with an autoencoder architecture for the stable training where D's objective is to match autoencoder loss distribution instead of data distribution (see Figure 9 (f)). In the sub-class of memory networks, memory within the network is introduced to handle the mode collapse and instability problem in unsupervised GANs framework. Within the sub-class of latent space engineering, latent space is reparameterized using a mixture of Gaussian model to get samples in the high probability regions in the latent space which supports better GANs performance (see Figure 9 (g)).

In this section, we shall thoroughly study the different designs of GANs network architectures with their strengths and drawbacks.

## 5.1. Conditional Generation

Conditional generation (e.g., cGANs [28]) have shown a significant improvement in generating good sample quality. Conditional GANs has shown a potential application for the image synthesis and image editing applications. In addition, a class-conditional model of images is not significant if it produces only one image per class. In conditional GANs (cGANs), a condition $c$ is induced on both G and D. The main aim of cGANs is to generate realistic images instead of making difference between generated samples based on input conditions. On the other hand, in conditional GANs, condition vectors are concatenated into some layers of G and D (see Figure 10). For example, in fully conditional GANs (FCGAN), each layer of D including $x$ is conditioned. This joint representation cannot capture the complex relationships between two distinct modalities [118]. To handle this issue, [47] introduced a Spatial Bilinear Pooling (SBP) approach where each pixel of an image is conditioned, i.e., multiplicative interaction between all elements of two vectors. Moreover, an Information Retrieving GANs (IRGAN) [47] is also proposed in which latent codes are conditioned explicitly. IRGAN explicitly put condition information in the latent codes.

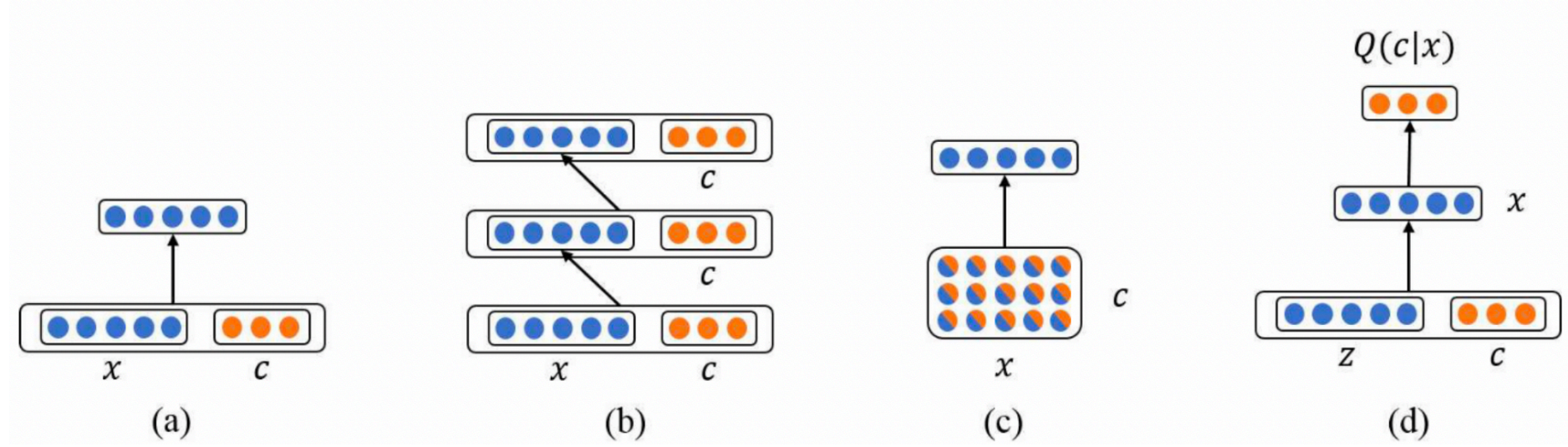

**Figure 10. Conditional GANs: (a) cGANs, (b) FCGAN, (c) SBP, and(d) IRGAN. Figure from [47]**

Im, et al. [48] introduced a new class of GANs based on the Recurrent Neural Networks (RNNs) instead of CNNs as Gs for GANs, called Generative Recurrent Adversarial Networks (GRANs). GRANs generate an image in a sequence of structurally identical steps without imposing a coarse-to-fine (or any other) structure on the generation process. The main advantage of sequential modelling in which generated output is conditioned on previous states repetitively is that it makes the complex data distributions modelling problem simple by mapping them to a sequence of easy problems. Moreover, authors also proposed new metric, called Generative Adversarial Metric (GAM) for evaluating adversarial networks quantitatively.

As image data have high variations, generation of diverse images with sufficient details is quite challenging for GANs. Recently, some works have focused to decompose GANs into a sequence of GANs for breaking difficult task into tractable sub-tasks. Denton, et al. [5] integrated a conditional model on the cascade of convolutional GANs with the framework of Laplacian pyramid (LAPGAN) with $k$ levels. LAPGAN model generates and upsamples images in multiple steps and have shown the generation of the higher quality images. But they suffered from the objects looking wobbly as noise added in chaining multiple models. In particular, to generate an image, LAPGAN model explicitly decomposes task into a sequence of conditional generations of levels of a Laplacian pyramid. [43] proposed another model, called Stacked GANs (SGAN) which is composed of a top-down stack of GANs and each stack is trained to produce low-level representations conditioned on high-level representations. The problem of estimating image distribution is decomposed into small tasks and intermediate supervision is provided by representation D at each training hierarchy. Authors introduced conditional loss to let G employ the high-level conditional information and an entropy loss to encourage each G for generating diverse representations in addition to adversarial loss (i.e., divergence measure). In addition, a representation Ds is proposed to be used at each hierarchy for providing the intermediate supervision to G at that level. Previous approaches introduced several loss terms for regularizing G's output without regularizing its internal representations. SGAN architecture is similar to LAPGAN as both consist of a sequence of GANs but LAPGAN generates *multi-resolution* images from coarse-to-fine (i.e., a sequential adversarial network) while SGAN models *multi-level representations* from abstract-to-specific. SGAN also works similar to InfoGAN [95] w.r.to the variational mutual information maximization technique but InfoGAN is used to forecast simply a small set of latent code, whereas SGAN predicts noise variables in each stack. Furthermore, InfoGAN maximizes mutual information between the output and the latent code, whereas SGAN maximizes the entropy of the output $h_i$ conditioned on $h_{i+1}$.

On the other hand, A cGANs must be capable to disentangle the intrinsic (latent variables) and extrinsic factors (known as auxiliary information), and also disentangle extrinsic factors' components from each other, in the generation process. Inverse cGANs produces disentangled information-rich representation of data which can be employed for some downstream tasks, such as classification. To achieve such optimal framework, [49] proposed an Invertible cGANs (IcGAN) to learn inverse mappings to intrinsic and extrinsic factors for pretrained cGANs by using two standalone encoders (Es) trained post-hoc, one for each task. However, IcGAN suffers from the following two limitations: IcGAN prevents E from having an effect on factors' disentanglement during the generation process and avoids E from learning the inverse mapping to intrinsic factors effectively.

In addition, existing encoder-based cGANs models suffer from the two major shortcomings: (1) extrinsic factors are not encoded [73], (2) if encoded, then encode them in fixed-length continuous vectors which do not have an explicit form [119]. This avoids data generation with arbitrary combinations of extrinsic attributes. To alleviate encoder-based cGANs issues, [50] proposed a network architecture, called Bidirectional cGANs (BiCoGAN) which learns inverse mappings of data samples to both intrinsic and extrinsic factors, and is trained simultaneously with G and D. BiCoGAN is similar to Bidirectional GANs (BiGANs) [78] w.r.to implicit regularization, mode coverage and robustness against mode collapse. Unlike conditional extension of ALI model (cALIM) [77], BiCoGAN involves both data samples and extrinsic attributes as inputs to encode the intrinsic features. However, it does not involve extrinsic attributes encoding from data samples. BiCoGAN could not produce better results as E could not perform the inverse mapping to extrinsic attributes and G does not include the extrinsic factors during the data samples generation. To handle these problems, training techniques are also introduced.

Existing researches show that cGANs could not perform well for the supervised tasks, such as semantic segmentation, instance segmentation, line detection, etc. The possible reason can be that G is optimized by minimizing a loss function that does not depend directly on the real data labels. To handle the aforementioned issue, [51] proposed to replace D with a siamese network working on both the real data and the generated samples to allow G's loss function to depend on the targets of the training examples. This approach is called Matching Adversarial Networks (MatAN) which can be utilized as a D network for supervised tasks. In addition, [52] proposed a class conditional GANs which is trained without manually annotated class labels. While, labels are derived automatically by applying clustering in the D's feature space. Clustering step finds diverse modes automatically and requires G to cover them explicitly.

On the other hand, some researchers proposed to use the Classifier (C), reconstructing side information to increase the performance of cGAN. [53] proposed auxiliary classifier GANs (AC-GAN) where a C is used as D of GANs architecture and a condition is categorial class label Cclass to only G instead of additional condition c to both G and D. In AC-GAN, D estimates a probability distribution over both sources and Cclass. To generate realistic images, AC-GAN demonstrated that addition of more structure to latent space with a particular loss function works well. But this variant cannot perform well for the semi-supervised learning as D suffers from two incompatible convergence points: discrimination of real and fake data and prediction of class label, and G cannot produce data in a particular class. Moreover, GANs, a two-player game consumes high time to reach equilibrium due to high-variance gradient updates. Also, G cannot control the semantics of generated samples.

To handle this issues, [54] proposed a three-player adversarial game to drive G to match the conditional distribution p($x$|$y$), called Triple-GAN for both classification and class-conditional image generation in semi-supervised learning where C results is passed to D as input. D identifies data-label pair is from the real dataset or not, while G and C characterize the conditional distributions between images and labels. But, TripleGAN does not have constraints to guarantee that semantics of interest will be captured by $y$ and also does not have a structure for achieving posterior inference for $z$.

Wang, et al. [55] also introduced a three-player game, called KDGAN, consists of a C, a teacher T, and a D in which C and T are trained from each other using distillation losses and are adversarially trained against D using adversarial losses. C learns the true data distribution at the equilibrium through the simultaneously optimization of distillation and adversarial losses. To achieve the low-variance gradient updates for speed up the training, samples are generated from the concrete distribution. Moreover, KDGAN achieves stable training when C impeccably models the true data distribution. But KDGAN suffers from G collapse when the class count increases. KDGAN idea is inspired by [120] in which D is used to train C for learning the data distribution produced by T, while in KDGAN, D trains C to learn the real data distribution directly. In addition, even though both Triple-GAN and KDGAN introduced three-players game, both models have some differences as follows: (1) In KDGAN, C and T (i.e., G) learn conditional distribution over labels given features, while in Triple-GAN, C and G learn a conditional distribution over labels given features and a conditional distribution over features given labels, respectively, (2) In KDGAN, generated samples from Gs are discrete data, while in Triple-GAN, generated samples include both discrete and continuous data.

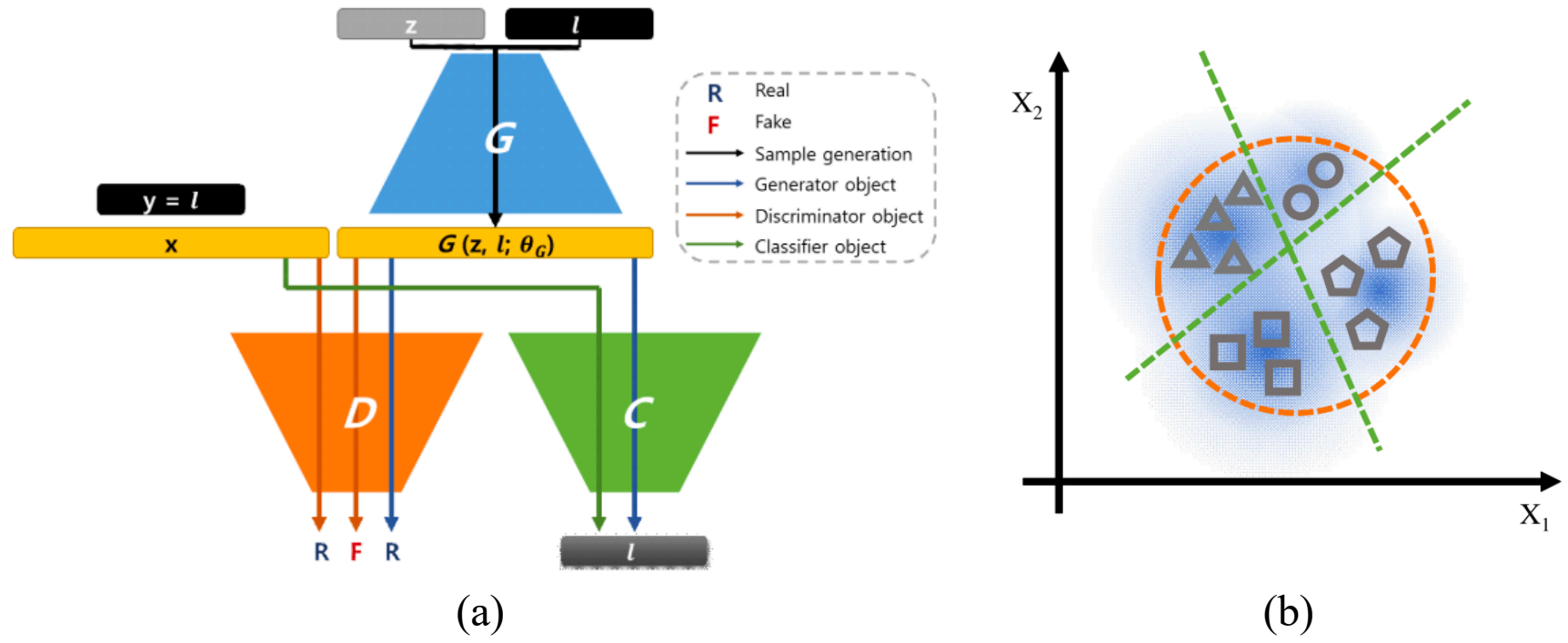

**Figure 11. (a) ControlGAN architecture** [56]. **(b) ControlGAN concept. Green dashed line represents C and orange line represents D. Grey boxes represents samples labelled with different class. Blue region, i.e., represents G which tries to learn data distribution and classify samples to correct labels, simultaneously.**

Furthermore, AC-GAN, TripleGAN and KDGAN use a classifier C connected to D, i.e., D chooses the samples condition. Therefore, these solutions can hardly deal with cGANs limitation, i.e., cGANs can generate images of different styles but it is not possible to generate images from two different domains, such as color and depth image domains. In addition, cGANs generates specific images, such as, face conditioned on the attribute vector, but cannot model image distribution conditioned on a part of that image or on previous frames [121], i.e., for image generation, major features, such as smile can be conditioned instead of detailed features, such as pointy nose. This occurs because D decides whether the condition/label is correct or not.

Due to this, limitation of cGANs cannot be handled by [54][53] as if very few examples are available for a particular condition/label in a dataset and/or condition/label is far from the data distribution's center where the samples densely exist, D distinguishes samples with such conditions fake. To handle this limitation, Lee, et al. [56] proposed to combine GANs with a decoder-encoder structure based architecture for controlling generated samples with detailed feature, called Controllable Generative Adversarial Network (ControlGAN) in which GANs game is formulated as three players game of G/De, D and C/E (see Figure 11). In ControlGAN, G tries to make fool D and be classified correctly by C. In ControlGAN, an independent network is used to map the features into corresponding input labels which dedicate D only for distinguishing real and fake samples and enhances the quality of generated data. This also allows ControlGAN to generate data beyond the training data. In addition, ControlGAN also uses an equilibrium parameter to balance between GANs and a decoder-encoder structure for stable training.

## 5.2. Generative-discriminative Network Pair

In recent years, GANs has achieved great success due to its ability to generate realistic samples, but traditional single-generator GANs worked well for only small images, such as MNIST but could not model the large images. To handle this issue, some researchers proposed to use the multiple Gs rather than a single one for generating high quality images by increasing the generation capacity of Gs

In addition, some researchers advocated to use the multiple Ds while some formulated the minimax game using multiple Ds and Gs. We classify these approaches into three categories: (1) training of single generator (G), (2) training of multiple generators (Gs), and (3) training of multiple discriminators (Ds). We shall discuss these approaches in this section in detail.

### 5.2.1. Training of Single Generator

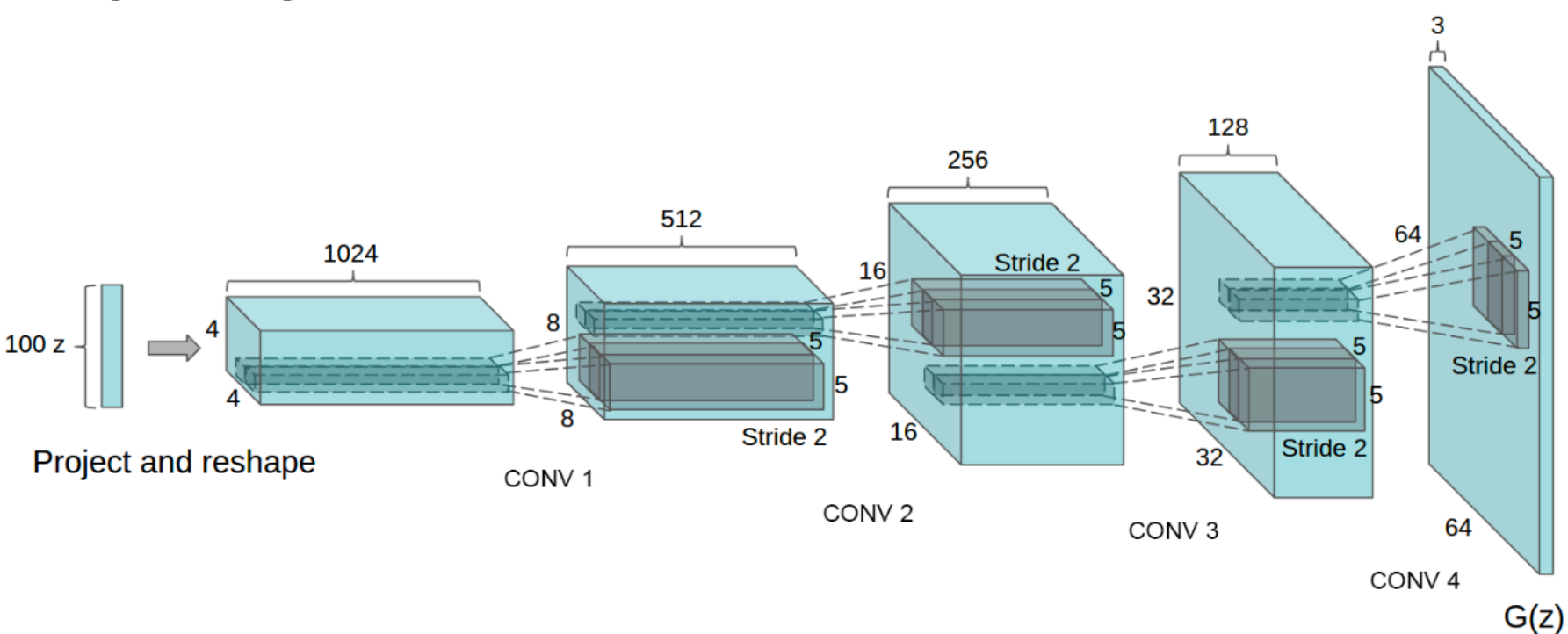

**Figure 12. DCGANs Generator. A sequence of four fractionally-strided convolutions for converting 100-dimensional uniform distribution $z$ into a 64 × 64 pixel image. Figure from [6]**

Basic GANs generated images are noisy and incomprehensible. GANs are unstable to train. Also, generation of high-resolution images is difficult. Therefore, there is need of a set of architectures resulting stable training for a range of datasets and allowing training of higher resolution and deeper generative models.

Radford et al. [6] successfully designed a class of architecturally constrained deep convolutional GANs, called DCGANs, which has shown substantial advancements on unsupervised image representation learning, i.e., more stable training and generates superior quality images. The architectural changes for stable DCGANs are as follows: (1) Use strided convolutions (D) and fractional-strided convolutions (G) instead of pooling layers (see Figure 12), (2) exploit batchnorm in both G and D, (3) do not use fully connected hidden layers for deep architectures, (4) apply ReLU activation in all G's layers excluding the output which uses Tanh, and (5) Use LeakyReLU activation in D for all layers. Even though DCGANs performed well in compared to basic GANs, DCGANs still suffered from some form of model instability – authors observed that as models training time is extended, DCGANs used to collapse a subset of filters to a single oscillating mode at times. It requires further investigation to handle this form of instability.

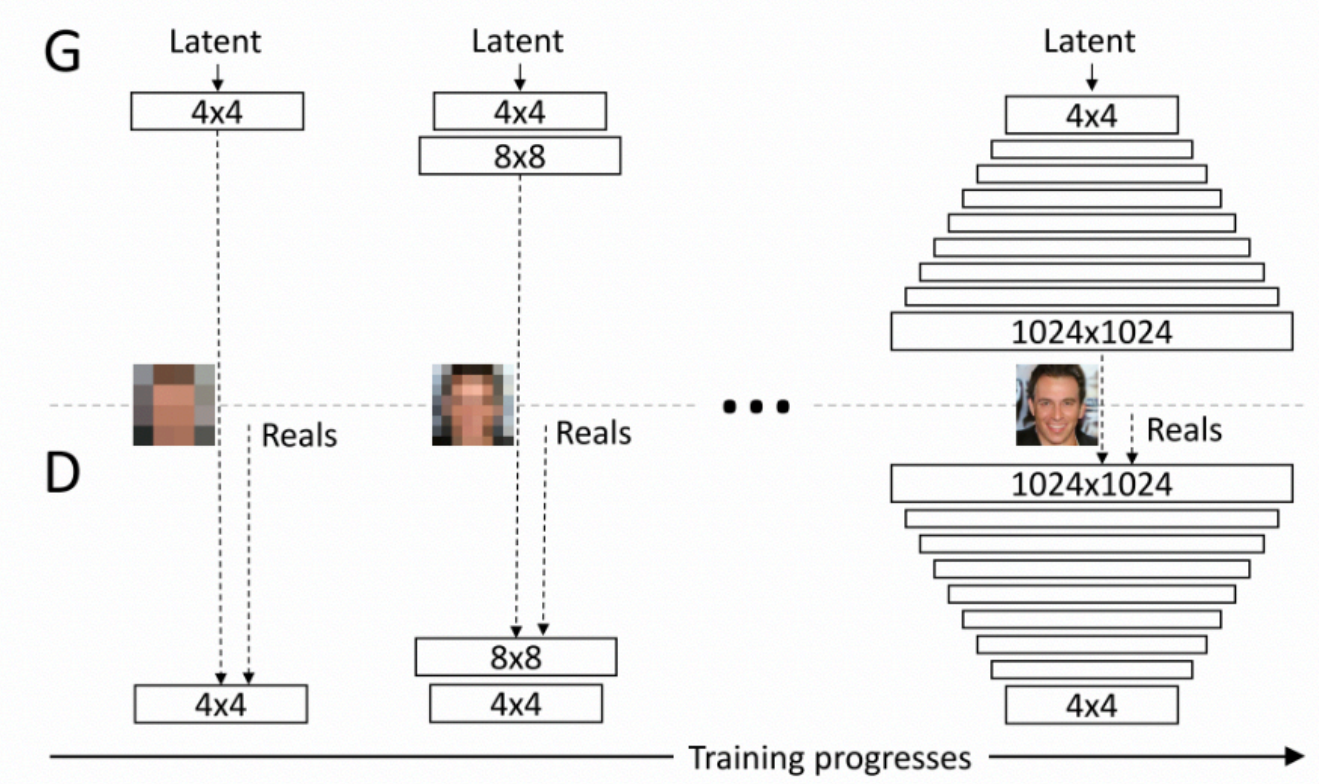

**Figure 13. Illustration of training process of ProgressGAN. G and D start with the low spatial resolution of 4×4 pixels and it keeps growing as the training advances. Figure from [8].**

Karras, et al. [8] introduced a novel training strategy for GANs, called ProgressGAN, in which low-resolution images are passed as input and then resolution of images is increased gradually by adding layers at each stage. The main idea of the proposed methodology is to grow the network of G and D step-by-step as it allows learning of large-scale structure of the image distribution first and then other finer details. Figure 13 demonstrates the training process of ProgressGAN This speedup training and improves stability. On the other hand, new layers are fade in smoothly to avoid the unexpected shocks to the well-trained, smaller-resolution layers. The proposed GANs training strategy of increasing resolutions performed well on CelebA dataset with size 1024 × 1024 and generated most realistic looking faces. A layer of minibatch standard deviation is also added at the end of D for capturing the diversity in the minibatch. This idea is quite simple to capture minibatch statistics and resolves the issue of sensitivity to hyperparameter tuning of the original minibatch idea of [36]. Lin, et al. [37] presented a new GANs framework, called PacGAN, which can be used by any existing GANs with a slight change to D. The main aim is to pass *m* packed/concatenated samples from the same class to D to be jointly classified as either real or generated. Packing penalizes Gs with mode collapse, therefore, favors distribution of G with less mode collapse during training. PacGAN needs no hyperparameter tuning but induces a little architecture's overhead.

On the other hand, some researchers proposed to integrate the GANs framework with other frameworks to improve the generating ability of the GANs. As a result, a strong generative model with better generation capability is achieved instead of two separate models. [57] introduced a Bayesian framework for unsupervised and semi-supervised learning with GANs to improve the generative ability. In GANs, updates are implicitly conditioned on a set of noise samples *z*, but in Bayesian GANs, *z* is marginalized from posterior updates using simple Monte Carlo. Jaiswal, et al. [58] introduced a change in D's architecture and argued to use a capsule networks (CapsNets) instead of standard CNNs. Results show that the learned images by CapsNets are more robust to changes in pose and spatial relationships of parts of objects in images. Authors also updated GANs objective function to *CapsNets margin loss* for training. CapsNets achieved good performance for the datasets, MNIST and CIFAR-10, while could not perform better for the complex datasets, such as ImageNet.

Lloyd, et al. [59] proposed a quantum adversarial game, called quantum generative adversarial networks (QuGANs) to show that quantum adversarial networks may have a great benefit over classical adversarial networks for high-dimensional data scenarios. In QuGANs, D optimizes strategy with the G's fixed strategy and G optimizes strategy with the D's fixed strategy over a fixed number of trials. QuGANs achieves Nash equilibrium when G finds the correct statistics while D cannot find difference between true and generated data. Most of the GANs models [6][36][105] for the image generation have used convolutional layers where convolution processes the information in a local neighborhood only and could not model long-range dependencies in images efficiently. Zhang, et al. [60] introduced a method for both G and D to model spatial relationship among separated regions in the images, called Self Attention GAN (SAGAN). Authors also presented two techniques, such as use of spectral normalization [110] in the G and D for stable GANs training (spectral normalization does not require extra hyperparameter tuning) and use of two-timescale update rule (TTUR) [33] for addressing slow learning in regularized Ds. In addition, to handle mode collapse and support stable training, [88] introduced explicit manifold learning as prior for GANs. A new target of Minimum Manifold Coding is further enforced for manifold learning to find simple and unfolded manifolds which works even in the case of the sparsely or unevenly distributed data.

### 5.2.2. Training of Multiple Generators

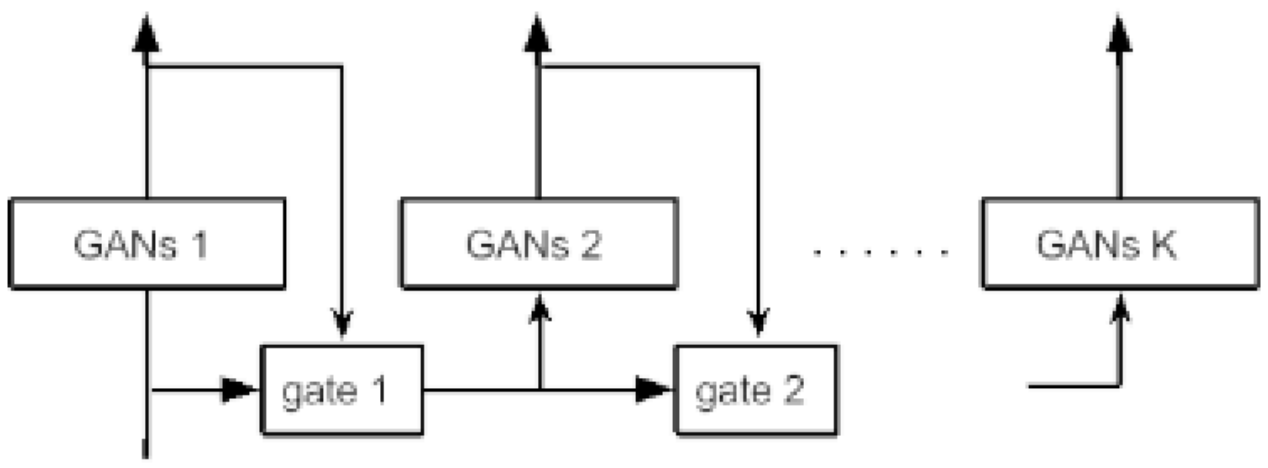

**Figure 14. cGANs framework consisting of multiple GANs [44].**

Ensembles have already shown potential for improving the results of discriminative CNNs. Wang, et al. [44] explored the usage of ensembles of GANs and proposed a framework having the cascades of the GANs, called cGANs, in which G focuses on capturing the whole data distribution instead of the principal mode of the data. Figure 14 shows the cGANs framework which consists of multiple GANs and gates. cGANs consists of multiple GANs where a G in each GANs tries to capture current data distribution which previous GANs could not captured efficiently. For selecting data, which is passed to the next GANs, it is considered that for badly modeled data *x*, the D value D(*x*) should be high, i.e., D is confident that *x* is real data. Gate function is used to re-direct the data to the next GANs. If D value D(*x*) is greater than a pre-determined threshold value *tr*, then *x* will be used in the next GANs. However, additive procedure of cGANs is not motivated by the theoretical analysis of optimality conditions. Further, Tolstikhin, et al. [45] have shown through the empirical analysis that cGANs heuristic fails to address the mode collapsing problem. They introduced a meta-algorithm, called Adaptive GAN (AdaGAN) similar to AdaBoost [123] in which each iteration corresponds to learning a weak generative model w.r.to a re-weighted data distribution. The assigned weight keeps changing to handle the hard samples. AdaBoost reweighs the training data and trains new Gs incrementally to get a mixture covering the whole data space. AdaGAN can be used on the top of a G architecture, such as a Gaussian mixture model, VAE [3], WGAN [29], UnrolledGAN [39] or mode-regularized GANs [40], which have been developed to handle the mode collapse problem. However, training multiple Gs in an iterative manner is computationally expensive. In addition, AdaBoost assumes that GANs based on single-generator can produce realistic images for some modes, such as dogs or cats but cannot capture other modes, such as giraffe [45]. So, by removing the images of dogs or cats manually from the training data and train a next GANs over this data can create a better mixture. But, in practice, this assumption does not work as single-generator GANs produces images of unrecognizable objects for diverse datasets, such as ImageNet.

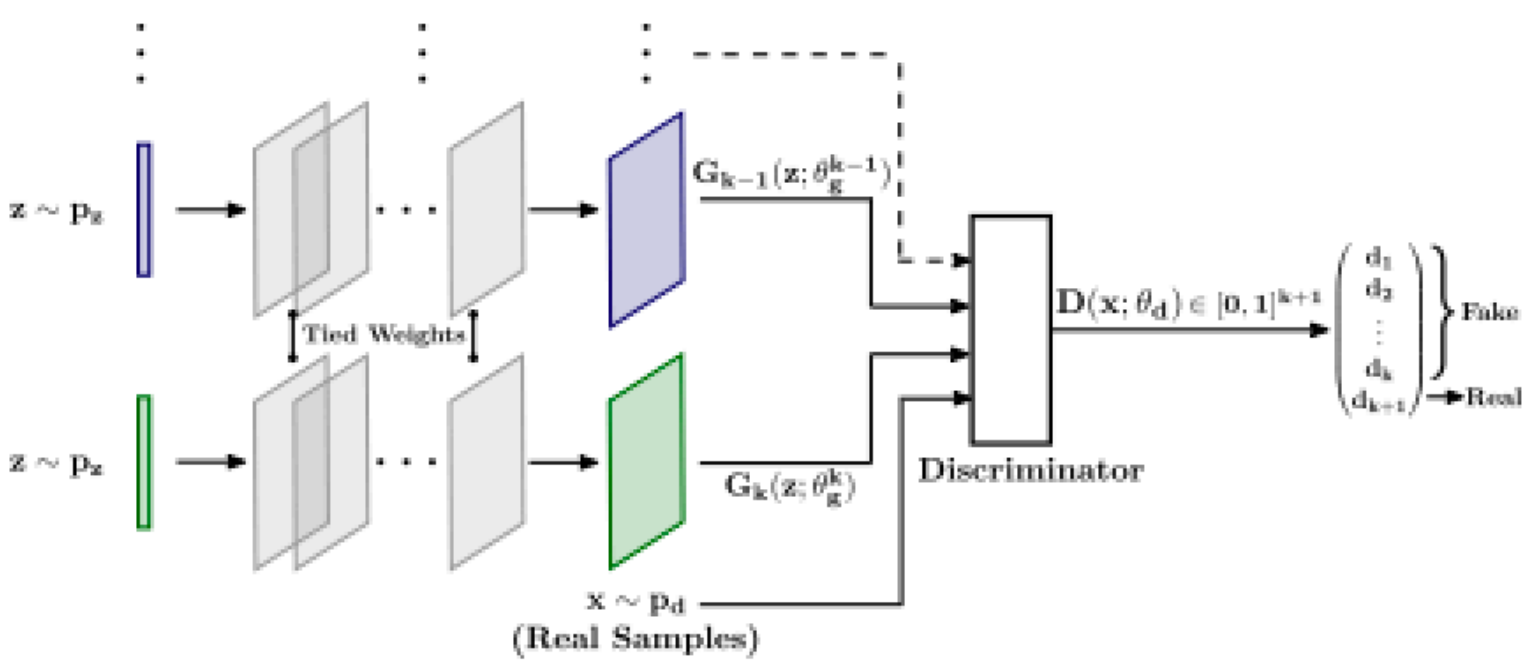

**Figure 15. Multi-Agent Diverse GAN (MAD-GAN). The D outputs *k* + 1 softmax scores signifying the probability of its input sample being from either one of the *k* Gs or the real distribution** [61]**.**

Ghosh, et al. [61] introduced a novel adversarial architecture consisting of a C and multiple Gs trained by a multi-class D which differentiate among fake and real samples with the identification of the G generated fake sample, called MAD-GAN. The idea is to approximate data distribution using a mixture of multiple distributions where each distribution captures a subset of data modes separately from those of others (see Figure 15). The idea of using multiple Gs is same as a mixture model where each G learns one data mode. In addition, objective function penalizes Gs for generating fake samples while does not encourage Gs to focus in generating variety of modes. Experimental results have shown that MAD-GAN performed extremely well for the highly challenging diverse-class dataset. Hoang, et al. [62] extended the concept of multi-generator GANs, called MGAN, where a set of Gs are trained simultaneously (share parameters) to approximate data distribution using a mixture of multiple distributions with the encouragement to specialize in different modes in the distribution. MGAN uses an additional model to classify (rather than modifying D) which G generated fake input where output of the classifier is further used as penalty term to force diversity among Gs. The objective function of MGAN focus to minimize the JSD between the mixture of distributions induced by the Gs and the data distribution and to maximize the JSD among Gs. Ghosh, et al. [63] presented a multi-agent GANs framework of multiple Gs communicating through message passing, called MPM GANs, for better image generation. Objectives are to pass the messages among Gs for capturing different data modes and promote competition among the Gs and tries to make other G better than the current G. Ge, et al. [64] proposed a training algorithm, called Fictitious GANs in which D is trained on the mixed outputs from a series of trained Gs. Fictitious GANs is a meta-algorithm which can be used on the top of existing GANs variants.

Another direction for handling mode collapse and support more stable training is to train multiple Gs and Ds in GANs as the use of mixture guarantees existence of approximate equilibrium. Arora, et al. [65] proposed to train multiple Gs and Ds with different parameters, called MIX+GAN and optimize the minimax game with the weighted average reward function over all pairs of G and D. But, training of MIX+GAN is computationally expensive as solution lacks parameter sharing and enforcing divergence among Gs is missing.

### 5.2.3. Training of Multiple Discriminators

Nguyen, et al. [66] proposed to use two Ds for yielding constructive gradient signals for G. In single generator dual discriminator architecture (D2GAN), one D works on the KL divergence to reward samples from the true data distribution while another D works on the reverse KL divergence to reward samples generated by G where G tries to fool both two Ds. This helps to avoid mode collapse problem. The combined use of KL divergence and reverse KL divergence into a unified objective function attempts to diversify the estimated density in learning multi-modes effectively. However, they tend to increase instability because the goals of two antithetical Ds conflict. Durugkar, et al. [67] also used several Ds for boosting the learning of G and stabilize GANs, called Generative Multi-Adversarial Network (GMAN). In GMAN, either average loss of all Ds or a D with minimum loss is used to train G. The main idea is to accelerate training of G to a more robust state irrespective of the choice of cost function.

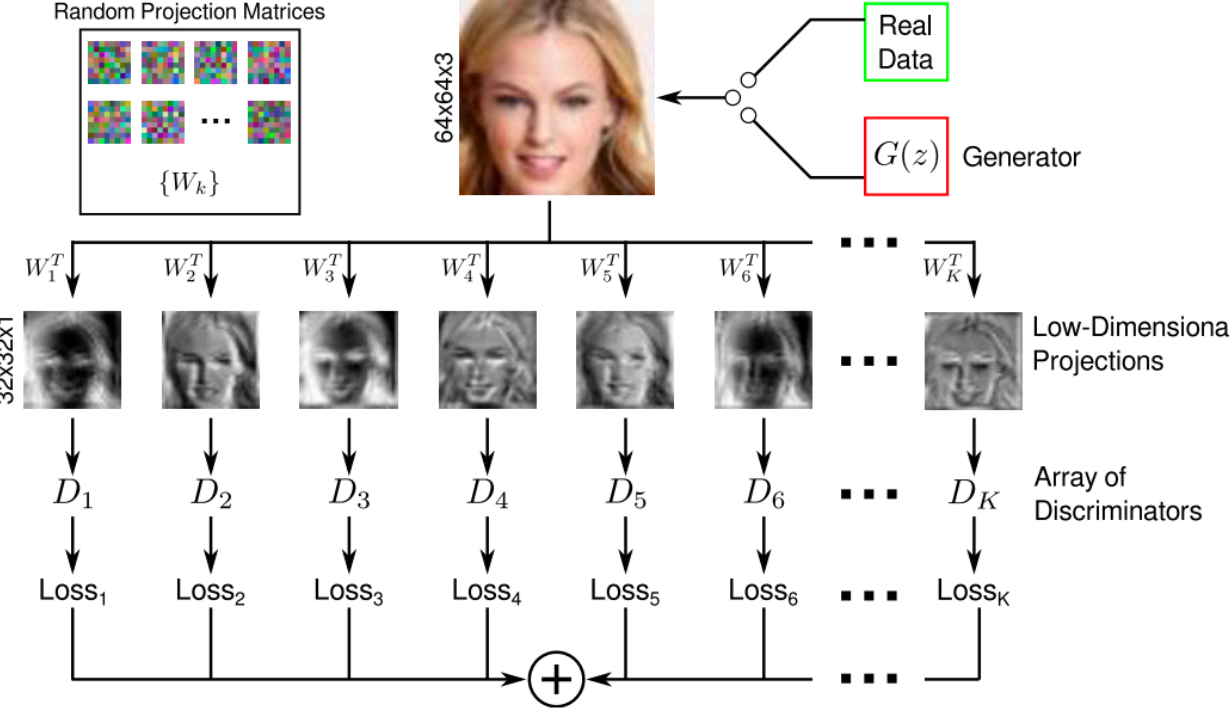

**Figure 16. Overview of StabGAN. Figure from** [68].

Despite the GANs progress, GANs is unstable in the case of high-dimensional data as real data distribution can have the possibility to be focused in a small fraction of the ambient space. Due to this, D can easily classify almost all generated samples as fake and does not provide meaningful gradients to D. Neyshabur, et al. [68] proposed a different approach to handle the instability in which G is trained against an array of Ds where each D handles different, randomly-chosen, low-dimensional projection of the data, (StabGAN)(see Figure 16). Existing similar approaches either train ensemble of GANs [44] or ensemble of Ds [67] while in this, low-dimensional projection of the data is provided to D so that D cannot reject generated samples perfectly, i.e., provides meaningful gradients to G during training. Instead of combining each D's output directly, each D's loss is calculated individually and then an average of all D's loss is calculated. Moreover, G learns real data distribution to fool all Ds simultaneously.

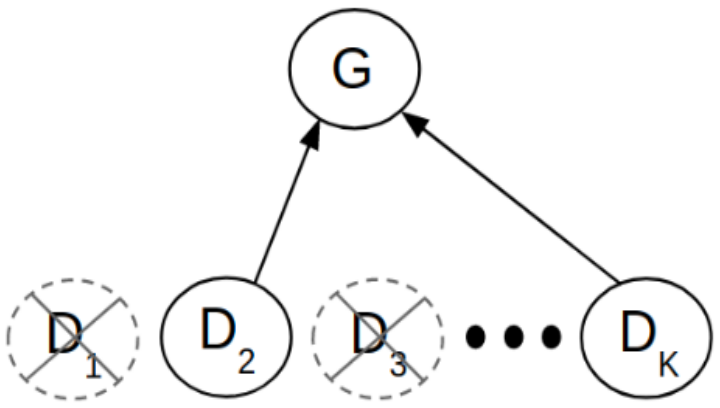

**Figure 17. Dropout-GAN. Some Ds are dropped depending on some probability. Figure from** [69]

In the previous two works, G's output depends on the feedback given by a specific set of Ds which is not a robust solution for handling mode collapsing and compromises the extensibility of framework. As a solution, Mordido, et al. [69] proposed to apply dropout mechanism by dynamically selecting feedback from ensemble of Ds that change at each batch for G's learning and to promote variety in its output, called Dropout-GAN (see Figure 17). On the other hand, the drop of a particular D's loss selected with a probability *d* before updating parameters of G, supports variability in the solution. In this case, G fools a dynamic set of Ds at every batch instead of one or a static set of Ds. Therefore, this solution can be known as regularization as it targets to encourage more generalizability on the fake samples produced by G. Durugkar et al. [67], and Nguyen, et al. [66] approaches limit the extensibility as they condition the D's architecture for promoting variety either by having convolutional architecture for D or by having different architectures for each D, while, the use of dropout approach does not compromise extensibility of the solution. In addition, to handle the issue of mode collapse in GANs, [70] proposed to assign a different portion of each minibatch, called microbatch, to a D. D's task of distinguishing between real and fake data is also gradually changed to discriminating samples from inside or outside its assigned microbatch with the use of a diversity parameter α. G is asked to generate diversity in each minibatch so that it is hard for each D to discriminate microbatch.

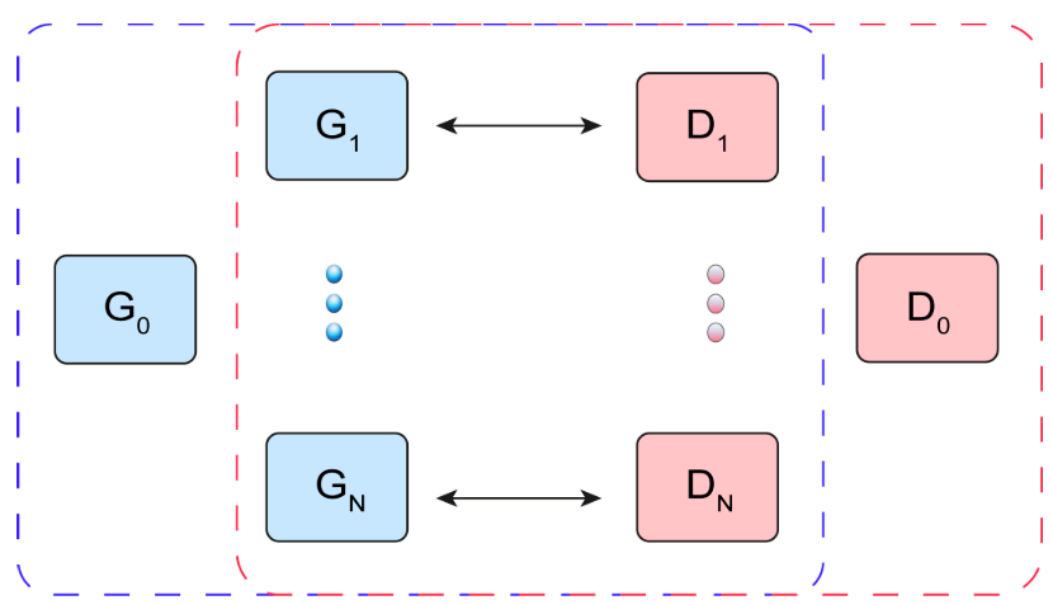

**Figure 18. Illustration of SGAN approach** [71]**.**

Chavdarova and Fleuret [71] proposed an alternative way for GANs training, called SGAN, in which a global pair of $G_0$ and $D_0$ is trained indirectly, i.e., $D_0$ is trained with $G_i$, $i = 1, \ldots, N$, and $G_0$ is trained with $D_i$, $i = 1, \ldots, N$ (see Figure 18). The main advantage of such training procedure is as follows: (1) global networks continue to learn with higher probability even if a particular local pair's training worsens; and (2) High computation can be performed in parallel which makes the time overhead less important factor. Compared to [45], SGAN can be used by any existing GANs framework as it runs in parallel and yields a single G.

### 5.3. Joint Architecture

GANs can generate more visually compelling sample images in compared to (V)AEs but in GANs, more complex loss functions are used than (V)AEs. However, basic GANs lacks an efficient inference mechanism. On the other hand, even though GANs produces more natural-looking images, instabilities in optimization induce mode collapse problem. To alleviate the above-mentioned issues with the GANs, recently, some research works have introduced to use the feature learning which includes features from the latent space and data space for improving GANs. The main idea to employ feature learning is that features from different spaces are complementary for producing realistic images. Combining the strengths and weaknesses of both (V)AE and GANs approach have provided a promising research direction for the unsupervised, supervised and reinforcement learning.

The combination of two architectures (AEs and GANs), i.e., encoder-decoder architecture, supports numerous benefits, such as can be used to reconstruct data (i.e., inpainting [111][112]), can be used for representation learning. AE-GANs can be primarily grouped into two methodologies: (1) combining AEs and GANs as data space AEs to learn a mapping from the data to the latent space and back to the data space; (2) combining AEs and GANs as latent space AEs, i.e., autoencoding the latent/noise space. In this section, we shall discuss both encoder-decoder architectures in detail.

#### 5.3.1. Data space autoencoders

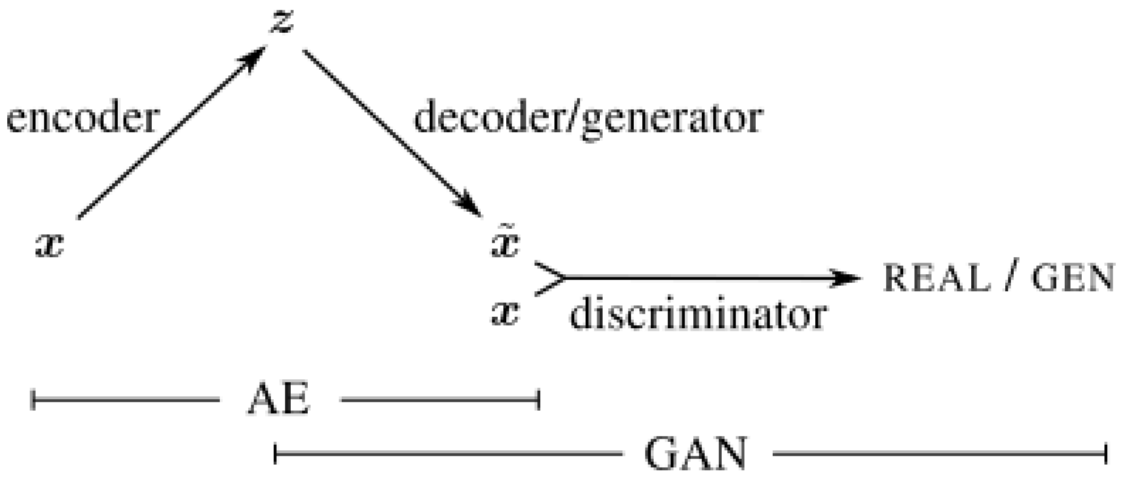

**Figure 19. Overview of VAE-GAN. VAE is combined with a GAN by collapsing Dec and G into one. Figure from** [72]

Larsen, et al. [72] trained VAE and GANs jointly, called VAE-GAN that uses learned representations to measure similarities in data space as element-wise metrics do not generate high resolution and realistic images (see Figure 19). In VAE-GAN, VAE decoder and the GAN's G are merged and trained jointly. The replacement of element-wise metric with feature-wise metric generates better image samples. VAE-GAN uses the combination of KL divergence and reconstruction loss (i.e., distance measure) for training the inference model. For this, optimization via backpropagation requires exact form of prior distribution and re-parameterization tricks. As GANs updates G with the reconstruction loss, GANs can successfully handle the mode collapse issue as G will able to reconstruct every input $x$. Both VAE and GANs cannot be applied for unsupervised conditional generation tasks as both cannot find disentangled mappings.

[73] introduced an adversarial autoencoder (AAE) that uses the GANs for performing variational inference. Matching the aggregated posterior to the prior guarantees that samples generated from prior space will be more realistic. In AAE, encoder (E) is used to map the data distribution to the prior distribution while decoder (Dec) is used to learn a deep generative model for mapping the imposed prior to the data distribution. Both adversarial and autoencoder network in AAE are trained together in two parts: *reconstruction phase* in which AE updates E and Dec for minimizing the reconstruction error of the inputs, and *regularization phase* in which adversarial network updates its discriminative network to discriminate real and fake data. This regularization does not guarantee that G will have capability to approximate data distribution accurately and handle the mode missing problem. Further, Mescheder, et al. [74] proposed a new training technique, called Adversarial Variational Bayes (AVB), to train VAEs adversarially for having the flexible inference model. Another line of research is the new adversarial learning for VAE: joint distribution of data and codes [75]. One possible solution is to feed observed data through E to produce codes while another solution is to draw latent codes from a prior and propagate through Dec to manifest data.

[40] introduced a mode regularized GANs (MDGAN) related to VAE-GAN [72] in terms of training an VAE jointly with the GANs model. However, VAE in VAE-GAN is used to generate samples whereas MDGAN's autoencoder based losses used as a regularizer to penalize missing modes which improves GAN's training stability and sample qualities. For getting more stable gradients, MDGAN used distance between real data and reconstructed data as a regularizer. MDGAN uses two Ds, one to discriminate between data and reconstructions (i.e., manifold) for learning a good AE and one to distinguish between two distributions, i.e., diffusion. Authors also proposed a set of evaluation metrics for evaluating diversity of modes and distribution fairness of the probability mass. Training of MDGAN is composed of two steps, a mode regularization step to reduce the model's variance and a diffusion step to reduce instability. However, MDGAN uses an additional autoencoder as well as a two-step training procedure which can be computationally expensive.

Tran, et al. [76] proposed two novel distance constraints for improving the G's training, called Dist-GAN, by an AE. First, latent-data distance constraint to impose compatibility between latent sample distances and the corresponding data sample distances so that G does not generate samples close to each other. Second, D-score distance constraint to align the distribution of fake data with the real data so that G can generate data similar to real ones. Dist-GAN is applicable to any prior distribution as it limits AE by the data and latent sample distances. Dist-GAN using only a D network while MDGAN [40] requires two Ds. To alleviate the mode collapse, VAE-GAN [72] considers the reconstructed data as fake, MDGAN also uses this similarly in the manifold step, while Dist-GAN employs them as real data, which is crucial to control D for avoiding gradient vanishing. Moreover, both VAE-GAN and MDGAN use reconstruction loss regularization for G, while Dist-GAN uses additional regularization for AE.

Furthermore, in GANs, the effective likelihood is not known and intractable, and GANs are based on density ratio estimation [41], [98], [126], [127] which provides a tool to overcome intractable distributions. One possible solution to handle the intractability of the marginal likelihood issue is not to compute it ever, and learn about the model parameters via a tool indirectly [25]. To implement this solution, Rosca, et al. [25] introduced α-GANs using the variational inference for training in which the synthetic likelihood is used instead of intractable likelihood function of basic GANs and an implicit function used instead of the unknown posterior distribution. This study combines variational lower bound on the data likelihood with the density ratio trick to understand the VAEs and GANs connection. The use of the density ratio allows GANs for the marginal likelihood intractability through its relative behavior w.r.to the real distribution. This trick is very useful to deal with implicit distributions or likelihood-free models as it only needs data from the two distributions without accessing to their analytical forms. Moreover, α-GANs uses adversarial loss with a data reconstruction loss to handle both the issue of samples blurriness and mode collapse.

Unlike α-GANs, VAE-GAN applies the analytical KL loss for minimizing the distance between prior and posterior of the latent and does not elaborate its connection to density ratio estimation. In addition, α-GANs is end-to-end unsupervised model which maximizes a lower bound on the real data likelihood and do not use any pre-trained classifier or a feature matching loss.

### 5.3.2. Latent space autoencoders

Data space autoencoders autoencode data points in which a reconstruction loss is calculated on input and encoded images. But selecting appropriate loss function is a challenging task. Otherwise, combining autoencoder with adversarial learning could be enough for better image generation. While selecting a loss function for autoencoder on noise vector $z$ is an easy task as $z$ are drawn from a standard normal distribution. In addition, the biggest question in GANs is "can GANs be used for unsupervised learning of rich feature representations for random data distributions?". The possible reason is that G supports the mapping of latent space to generated data, i.e., an inverse mapping from data to latent representation is not supported by GANs. Inverse mapping is extremely important as it presents an information-rich representation of $x$, that can be employed as input for downstream tasks (such as, classification) [77][78]. To handle this issue, a new autoencoder-type adversarial architecture is required which supports unsupervised learning with both generation and inference.

Unlike prior AEs and adversarial networks hybrids, it is required to setup the adversarial game directly between the E and G with no trained external mappings in the process of learning. To handle the above-mentioned issues, another class of methods proposed in which methods learn to map latent space to the data space and backward. The main advantage of these methods over data space autoencoders is that the noise vectors can be reconstructed easily as the distribution from which they are chosen is known.

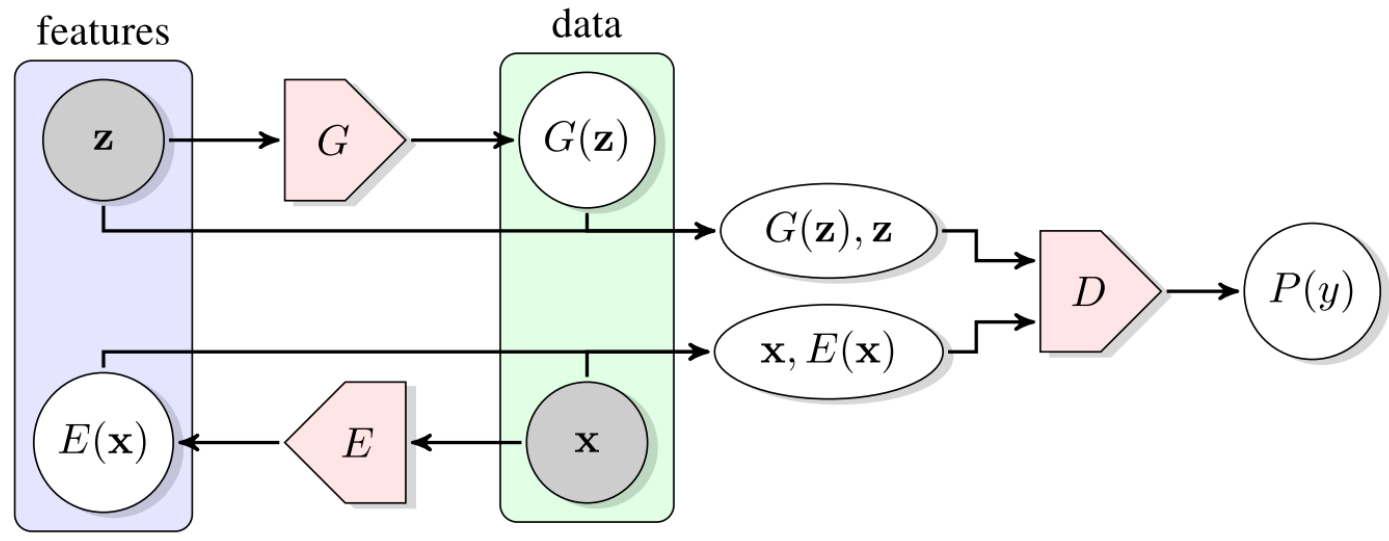

**Figure 20. Bidirectional Generative Adversarial Networks (BiGAN)** [78].

Dumoulin, et al. [77] introduced an approach similar to adversarial autoencoder, called Adversarially Learned Inference (ALI), in which inference is performed by starting an adversarial game between E and Dec/G through a D that works on *x* and *z* space. D, Dec/G, and E are parametric deep (multi-layer) network. In ALI, both G and E are trained together, and D is trained to discriminate between two joint distributions over image and latent spaces produced either by E on the real data or by G to the latent prior. Donahue, et al. [78] proposed the same model as ALI, called Bidirectional GANs (BiGAN), and explored the benefits of learned features for supervised and semi-supervised tasks. In addition, unlike adversarial autoencoder, ALI does not optimize explicit reconstruction loss and D receives joint pairs of samples ($x$, $z$) rather than $z$ samples (see Figure 20). ALI and BiGAN learn bidirectional mapping between the data and latent space while basic GANs learns a unidirectional mapping from latent to data space. In this case, gradients should propagate from D to E and Dec which can be achieved through reparameterization trick [3][128][129]. For the sampling, a random variable is computed as a deterministic transformation of some $z$ such that its distribution is the desired distribution. As gradient propagation into the E and Dec depend on the reparametrization trick, ALI and BiGAN are not directly applicable to either applications with discrete data or to models with discrete latent variables. ALI and BiGAN can match joint distributions of Dec and E and performs inference by sampling from E's conditional that also matches the Dec's posterior. However, achieving equilibrium of the jointly adversarial game is difficult as the dependency structure between data and codes is not explicitly specified [79]. Therefore, inference in ALI is not always effective. Unlike ALI, decomposed adversarial learned inference (DALI) [79] breaks the problem of matching the joint distributions into two sub-problems - matching priors on the latent codes and conditionals on the data explicitly. Due to this constraint, DALI gets better generation capability.

On the other hand, most of the applications require to handle multiple source of information, i.e., multiple views. For example, in traffic prediction, multiple views, such as traffic data, weather data, Point-of-Interest data, etc., are needed for efficient prediction. Multi-view machine learning is a long-studied problem where most of the existing studies have focused on the classification viewpoint and assume that all the views are present all the time. To handle this issue, Chen, et al. [80] proposed an extended architecture of BiGAN, called conditional-views BiGAN (CV-BiGAN) which supports to model a conditional distribution P($y$|.). Authors proposed another model on the top of the CV-BiGAN architecture, called multi-view BiGANs (MV-BiGAN) which can predict in case of one or few views availability and updates its prediction if new views are available. For the stable MV-BiGAN model, authors have also proposed a regularization term to add new views to existing views for controlling the uncertainty over the outputs. Belghazi, et al. [81] attempts to extend ALI [77], named HALI, by achieving better perceptual matching in the reconstructions, and by being capable to reduce the observables using a sequence of composed features maps. Unlike VAE-GAN, HALI offers more meaningful reconstructions with different levels of fidelity and minimizes perceptual reconstruction error implicitly during adversarial training. [82] proposed a novel autoencoder-type architecture, called Adversarial Generator-Encoder (AGE) Network which comprises two parametric mappings: E and G. E maps data $x$ to latent space $z$ while G maps $z$ to $x$. E and G game is useful in matching distributions. In AGE, E encodes data into codes taken from the prior. During the time of encoding, Dec tries to generate samples where encoded samples will match to the prior distribution. As E and Dec do not depend on D, it reduces the computational complexity and the converge time in comparison to the previous models utilizing a bidirectional mapping.

[83] has shown that latent space-based encoding, which is learned independently of the generated data, does not support good quality reconstruction and enhances the chance of mode collapse in the generated data space. To address this, authors proposed a Variational Encoder Enhancement to GANs, called VEEGAN, to map both the real and generated data space to a fixed distribution in a variational framework. VEEGAN proposed to use an additional reconstructor network (Rn) where G and Rn are trained jointly using an implicit variational principle to encourage Rn for mapping the data distribution to a Gaussian and for approximately inverting G's act. As a result, this forces G to map from the noise distribution to the true data distribution. Like VEEGAN, [130] also tried to enforce diversity in latent space. [130] used the last layer of D as a feature map for studying real and fake data distribution. Authors used the Bures distance between covariance matrices in feature space for matching real and fake batch diversity. The results show that matching the diversity handles the issue of mode collapse and also generate good sample quality.

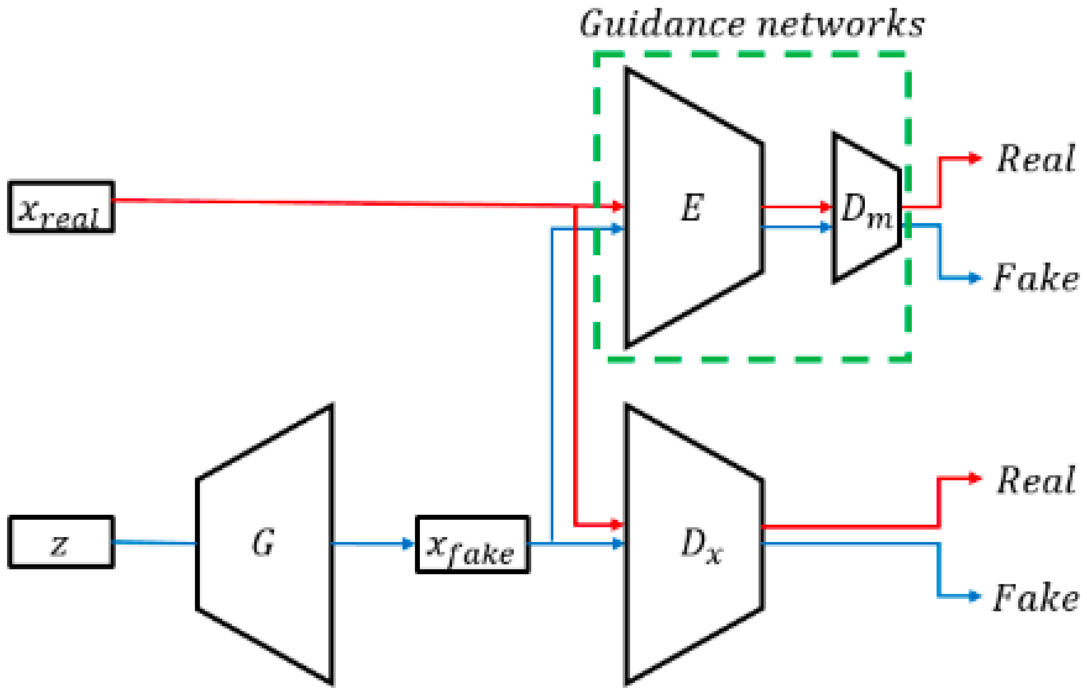

**Figure 21. MGGAN model structure. $x_{real}$and $x_{fake}$ are sample of $p_{data}$and $p_{model}$, respectively; $z$ is latent vector; The guidance network consists of *e* and D*m*, where *m* means manifold space. MGGAN is similar to basic GANs except the guidance network. Figure from** [84]

Unlike MDGAN [40], VEEGAN uses the reconstruction loss in the latent domain instead of the data domain and generates good quality images. Also, MDGAN and VEEGAN improve inference mapping through this reconstruction loss in compared to ALI [77] and BiGAN. Even though, these methods perform well in compared to data space autoencoders, still they do not support the learning of disentangled relation between latent space and data as data is inverted to a semantically useless fixed noise distribution. To handle the issue of bidirectional mapping, Bang, et al. [84] introduced a naïve weakly bidirectional mapping approach, called manifold guided generative adversarial network (MGGAN), in which a guidance network is induced with the basic GANs framework for learning all modes of data distribution (see Figure 21). In MGGAN, E and G are not connected, i.e., they are trained separately. This separation is assumed effective w.r.to performance as both can focus on their own objectives. In MGGAN, bidirectional mapping is used to regularize G training so that G cannot fool D by producing the same or similar samples corresponding to a single major mode of true data distribution. [77][78][40][83] also proposed encoder based architecture to support the bidirectional mapping and mapped $p_{data}$ into low dimensional manifold space. Even though network architecture is similar to MGGAN, existing bidirectional mapping is designed to map $p_{data}$ into $p_z$ (i.e., inference mapping), while MGGAN maps $p_{data}$ onto meaningful manifold space; called manifold mapping. Moreover, authors claimed that MGGAN learns the meaningful landscape in latent space; therefore, MGGAN does not overfit the training data.

## 5.4. Improved Discriminator

To handle the bad gradients problem during the GANs training, D is replaced with autoencoder where D assigns low energy to training data while high energy to samples generated by G. G and D objectives are generalized to consider real-valued "energies" as input instead of probabilities [85].

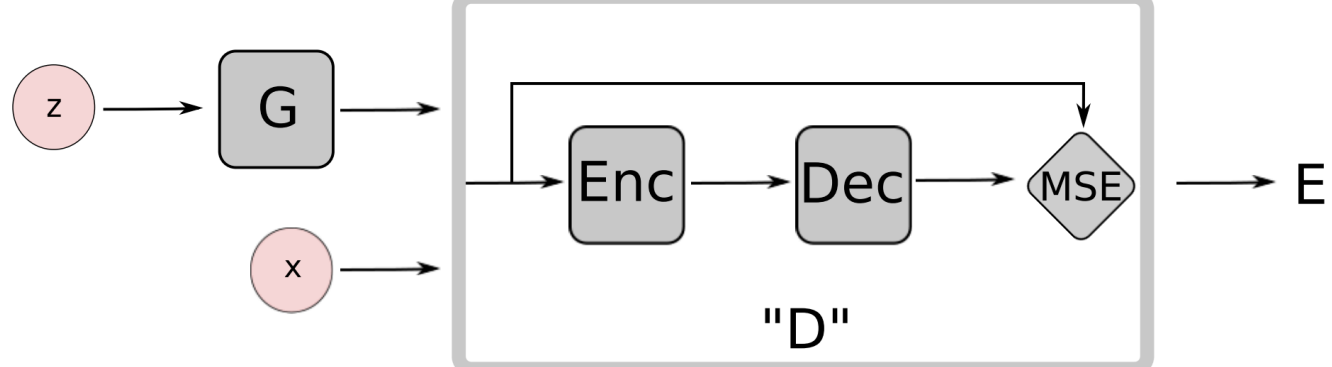

**Figure 22. EBGAN architecture which replaces D with an autoencoder. Figure from** [85]

Some researchers proposed to project GANs as energy-based model in which a function is built to map [131] each point of an input space to a single scalar, called energy. In energy-based models, energy surface is modelled like preferred patterns get assigned low energies and undesirable patterns are getting assigned high energies. Zhao, et al. [85] presented an Energy-based GANs (EBGAN) in which an autoencoder is used for D and implemented an energy function on D to assign low energies to regions near the data manifold and higher energies to other regions (see Figure 22). Moreover, to stabilize the EBGAN training, a hinge loss is used as objective function so that D can remove synthetic samples with energy over a margin $m$. Obtaining an appropriate value of $m$ is a vital factor for effective training and it highly depends on both architecture selection and data complexity. On the other hand, EBGAN has proposed to reduce total variation (TV) distance between the real and generated data distributions. TV distance has same regularity as JSD. Therefore, EBGAN can also have the same problems as basic GANs, not able to train D till optimality which causes bad gradients. On a contrary, to restrict D from degenerating to uniform prediction, [132] proposed a regularized EBGAN with entropy loss to generate many possible outputs from the same conditional input. Berthelot, et al. [86] extended EBGAN, called BEGAN by introducing a loss function matching the fake data's energy to a fraction of the true data's energy.

Then, Wang, et al. [87] proposed a new robust training procedure, called Margin Adaptation for Generative Adversarial Networks (MAGANs), to maintain the equilibrium between D and G through hinge loss margin using the expected energy of the target distribution. EBGANs, MAGANs and BEGANs take care of the expected energy of real and fake data for generating realistic samples and for controlling training. But, BEGANs is more stable, easy to train and robust to hyperparameter variations. Unlike EBGANs, MAGANs do not use any new hyperparameter and eliminate the dependence on the margin hyperparameter. Convergence of MAGAN to global optima is easier than both EBGANs and BEGANs. Moreover, MAGAN converges to its global optimum when distribution of real and generated data match exactly, while BEGANs provide no such guarantee. In addition, to handle mode collapse and support stable training, [88] introduced explicit manifold learning as prior for GANs in which a manifold preserving reconstruction loss is used during G's training. Reconstruction loss is one of the efficient ways to guarantee not missing information in GANs. Even EBGAN replaced D loss with reconstruction loss for showing other energy functions performance. A new target of Minimum Manifold Coding is further enforced for manifold learning to find simple and unfolded manifolds which works even in the case of the sparsely or unevenly distributed data.

Another line of research is to improve the generative power of G when a huge size of training data is unavailable. In this case, existing energy-based models cannot perform as GANs requires to learn enough features for unsupervised learning. To handle the above-mentioned issue, Deepearthgo, et al. [89] proposed to train a G by learning two mapping functions simultaneously, called Max-Boost-GAN. This improves the power of G without expanding the network's size. Besides this, cost of GANs does not increase. In addition, to get better gradients when G is distant from convergence, authors have introduced a margin loss in GANs objective.

### 5.5. Memory Networks

One of the architecture solutions introduced memory within the network to handle the mode collapse and instability problem in unsupervised GANs framework.

[90] discussed two main causes of instability in the unsupervised GANs framework, namely structural discontinuity problem and forgetting problem. First, basic GANs use a unimodal continuous latent space (e.g. Gaussian distribution). Due to this, it cannot handle structural discontinuity between different classes and suffers from the mode collapse problem. E.g., in basic GANs, both building and cats are embedded into a common continuous latent distribution while both do not share any intermediate structure. Second, D forgets the past generated samples by G during the training as D often focuses only on latest input images. This forgetting behavior causes instability as loss functions of G and D computed on the basis of each other's performance.

To handle the aforementioned problems, [90] introduced a memory network for the unsupervised GANs training, called MemoryGAN. To handle the structure discontinuity problem, memory network learns representation of training samples and helps G to better understand the underlying class. The forgetting problem can be handled by learning to memorize distribution of the data generated by G, including rare ones. MemoryGAN introduces a life-long memory into D to increase the model's memorization capacity explicitly instead of adding regularization terms [122] or modifying GANs algorithms [83]. In addition, MemoryGAN is quite similar to one of the unsupervised GANs frameworks, InfoGAN [95], but InfoGAN learns latent cluster information of data into small-sized model parameters implicitly.

The objective of MemoryGAN depends on InfoGAN, where MemoryGAN also adds a mutual information loss between $K_i$ and $G(z, K_i)$. This loss computes structural similarity between the sampled memory information and generated sample. $\hat{I}$ is the expectation of negative cosine similarity.

$$L_D = -\mathbb{E}_{x\sim p_{data}}[\log D(x)] + \mathbb{E}_{(z,c)\sim p_{(z,c)}}\left[\log\left(1 - D\left(G(z, K_i)\right)\right)\right] + \lambda\hat{I}$$
$$L_G = \mathbb{E}_{(z,c)\sim p_{(z,c)}}\left[\log\left(1 - D\left(G(z, K_i)\right)\right)\right] + \lambda\hat{I}$$

### 5.6. Latent Space Engineering

Existing works to handle the issues of GANs focused on the use of a fixed latent distribution, either a unimodal Normal [4][6][29][77] or a factored multimodal distribution [95][133] with uniform mode priors. But still, these distributional selections for the latent space leads a dissimilarity between the modes of generated and real data distributions (has a non-uniform mode prior). This case is possible when every data mode denotes a class and class distribution is imbalanced [92]. Therefore, for better performance, some researchers have proposed approaches for learning a better distribution of noise.

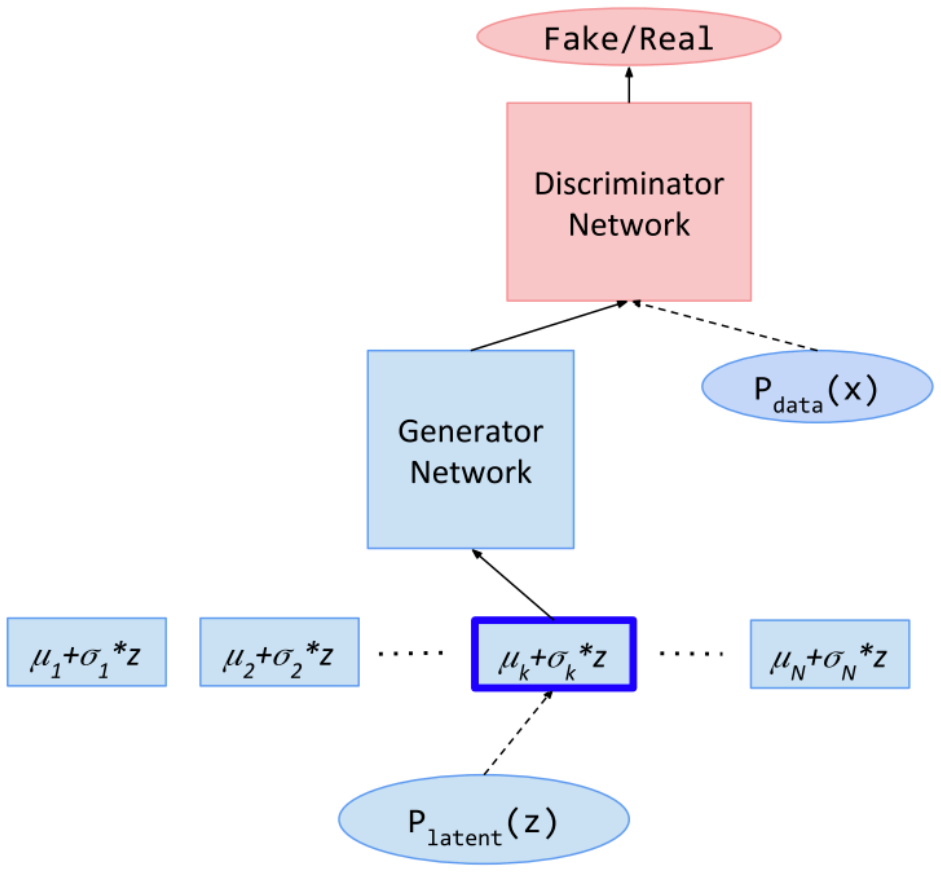

**Figure 23. In DeLiGAN, one of the Gaussian components is randomly selected (as dark blue box) and then "reparameterization trick"** [134] **is used to get a sample from the chosen Gaussian. Figure from** [91]

Gurumurthy, et al. [91] proposed an architecture to handle the mode collapse problem in the case of diverse and limited training data, called DeLiGAN (see Figure 23). DeLiGAN reparametrizes the latent generative space as a mixture model and learns the parameters of mixture model along with GANs. The primary aim of this work is to increase modeling power of the prior distribution rather than increasing the model depth. Mishra, et al. [92] introduced to construct the latent space (additive noise) in such a way that it matches to its target distribution w.r.to number of modes and modal mass or mode priors, called NEMGAN. Authors provided a solution for "*How to learn the latent space distribution in a GANs framework such that modal properties of the true and the generated space are matched?*". Furthermore, [93] introduced to use the multiplicative noise instead of additive noise for better visual quality and diversity of generated features, and explored for the unsupervised learning of features.

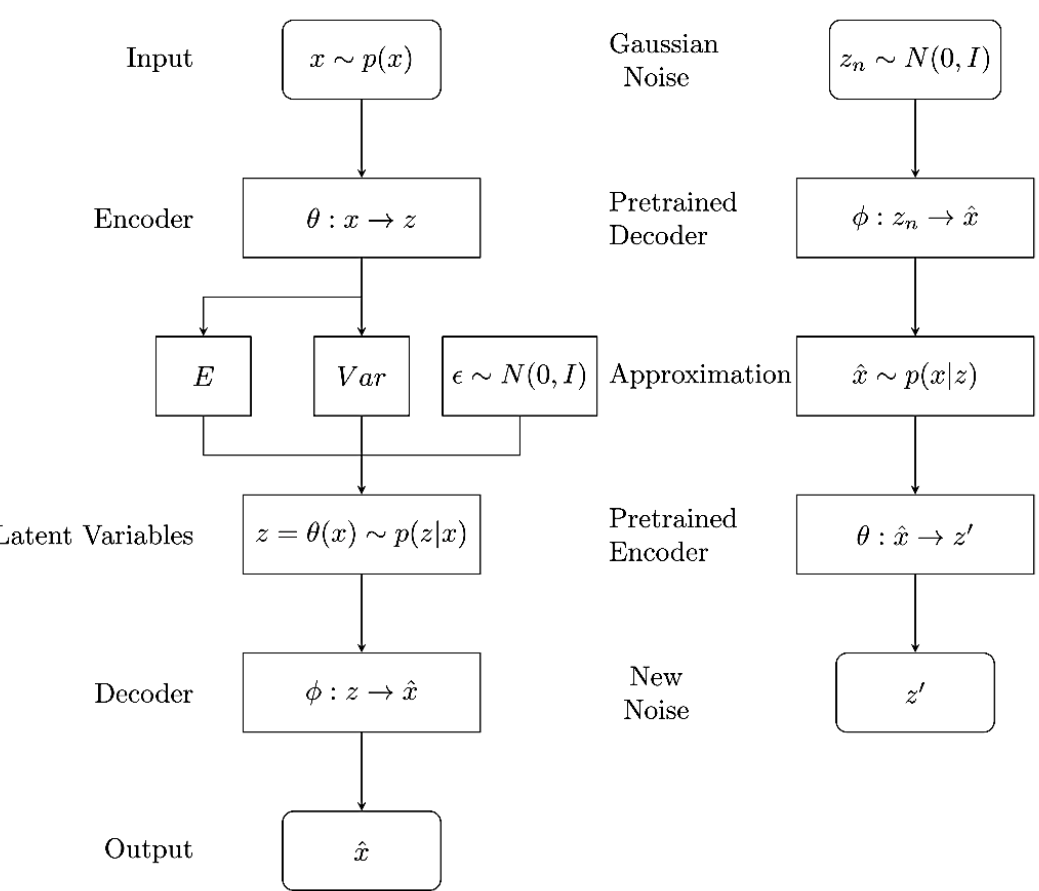

**Figure 24. Standard VAEs model (on left side) and Decoder-encoder (DE) structure (on right side). In DE structure, pre-trained Dec and E in VAEs are swapped. Figure from** [94]

In DCGANs, *z* is sampled from a Gaussian distribution and then G maps whole normal distribution to the images. Due to this, DCGANs cannot reflect the inherent structure of the training data. To handle this issue, Zhong, et al. [94] proposed a Decoder-Encoder structure by transforming the original Gaussian noise *z* to an informative one *z'* (see Figure 24). Authors have shown the analysis that proposed solution accelerate the training process and enhance the quality of generated images. Authors also proposed hidden-space loss function to make model robust. On the other hand, some researchers proposed that the noise vector *z* used in GANs formulation can be used by G in any manner. Due to this, it is feasible that *z* can be utilized in highly entangled way which can cause individual dimension of *z* to not correspond to semantic features of the data. Therefore, *z* can be divided into two parts instead of the use of a single unstructured noise vector [95]. First, *z* as a source of incompressible noise and second, c is a latent code to target the salient structured semantic features of the data distribution. This is a simple modification to GANs objective to learn interpretable and disentangled representations. Authors have also claimed that InfoGAN is able to disentangle both discrete and continuous latent factors without any extra training time. Further, [95] proposed an information-theoretic regularization to maximize mutual information between a small subset of the *z* and the G's distribution, called Information Maximizing Generative Adversarial Networks (InfoGAN). The unsupervised disentangled representation learned by InfoGAN outperforms existing supervised label information works [135][136]–[138][73]. In addition, a latent-code reconstruction based penalty is used in the cost function. Like VEEGAN [83], InfoGAN uses an AE over the latent codes, while in InfoGAN, reconstructor network are not trained on the true data distribution. Unlike VEEGAN, in InfoGAN, a latent code part is reconstructed. Unlike VEEGAN, InfoGAN requires some stabilization tricks as basic GANs.

# 6. NEW LOSS FUNCTION

Learning in implicit generative models, or likelihood-free models comprises of two steps: *comparison* and *estimation*. The comparison step uses the density ratio or difference estimators, while in estimation step, parameters of generative models are learned. For learning, two sets of samples drawn from the true data distribution and the model distribution are compared, this process is known as *density estimation-by-comparison*. To compare the true data distribution p($x$) with model distribution q($x$), two methods are used density difference p($x$)-q($x$)) and the density ratio p($x$)/q($x$). The density ratio estimation can be further categorized into three general approaches: *class-probability estimation*, *divergence minimisation* and *ratio matching*, while density difference estimation involves moment matching. The density ratio trick is widely used [4][126][73], [74], [139], [140][141] where *class probability estimation* is the most popular approach.

Basic GANs is based on the class probability estimation where density ratio can be computed by building a classifier to distinguish observed data from that generated by the model. Density ratio includes samples only from two distributions, therefore, makes it suitable for handling with implicit distributions or likelihood-free models. Divergence minimization minimizes the divergence between the p($x$) and q($x$) and uses it as an objective to drive learning in the generative model. Ratio matching directly minimises the error between the true density ratio and an estimate of it. Moment matching compares the moments of the two distributions by minimising their distance. Several works have been introduced based on the divergence minimisation [97], ratio matching [98], and moment matching [46].

Statistical divergences, such as KL divergence, Reverse-KL divergence and JSD, measure the difference between two given probability distributions and belong to f-divergence class of probability distance metrics (used in basic GANs [4], f-GAN [97], b-GAN [98]). Integral Probability Metrics (IPMs) is another class which involves Wasserstein distance (used in WGAN, WGAN-GP [29][107]) and Maximum Mean Discrepancy (MMD) (used in MMDGAN [46]). The difference between f-divergences and IPMs is that f-divergences computes distance using p($x$)/q($x$), i.e., density ratio, while IPMs uses the difference, p($x$) - q($x$), i.e., density difference.

Several methods have been proposed for improving the quality of gradients and providing additional supervision to G and D. New probability distance and divergence are introduced to handle the issue of *vanishing gradients* and non-convergence in training. In addition, some techniques have been proposed to introduce different kinds of regularization on weights or gradients for stable training, including gradient regularization, spectral-norm regularization [49], etc.

In this section, we shall discuss more stable alternatives of the objective functions for G and D and objective function with the regularization in detail.

## 6.1. New Probability Distance and Divergence

In this sub-section, we shall discuss popular probability distances and divergences used in the context of learning distributions to improve the GANs training stability and mode collapse problem.

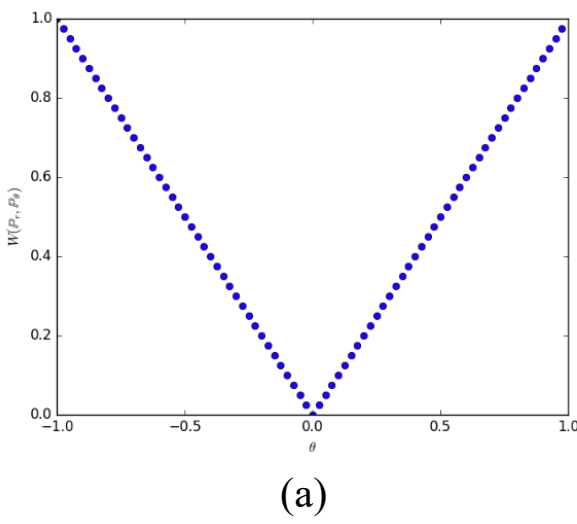

(a)

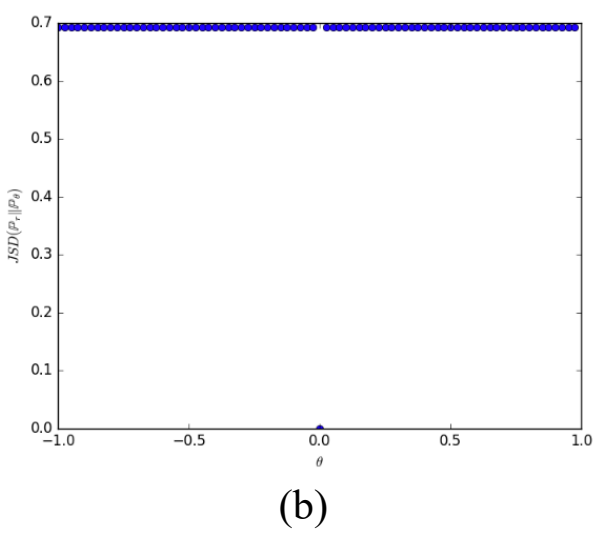

(b)

**Figure 25. $\rho$(P$\theta$, P0) as a function of $\theta$, (a) $\rho$ is the EM distance (b) $\rho$ is the JSD. The EM plot is continuous and gives a usable gradient everywhere, whereas JS plot is not continuous and does not give a usable gradient. Figure from** [29].

Arjovsky, et al. [29] proposed a loss function which also acts as a measure of convergence, called Wasserstein Generative Adversarial Networks (WGAN). WGAN has non-zero gradients everywhere and the implementation includes removing the sigmoid function in the objective and adding weight clipping to the D's network. To accelerate the GANs training and make the training process stable, WGAN introduced to reduce the JSD with an efficient approximation of Earth-Mover (EM) distance [142] as shown in Figure 25.

EM distance is continuous and differentiable. This new metric is effective in solving mode collapse by stabilizing GANs training. WGAN allows to train the critic till optimality which supports the equilibrium between G and D. A well-trained critic provides high quality gradients which are used to train G. However, WGAN can be unstable when gradients of the loss function are large. Thus, too large weights are clipped after each SGD update. Even though WGAN supports stability and better mode coverage, it suffers from slow training. Moreover, tuning weight clipping and hyperparameters is a tedious task.

[96] discussed that assuming model should have infinite capacity in basic GANs causes training issues as it constrains a model to lie in Lipschitz continuous function space. In the case of WGAN, Lipschitz condition is from the Kantorovich-Rubinstein duality and only the critic is constrained. Loss sensitive GAN (LS-GAN) [96] used a weight-decay regularization technique so that weights of a model to lie in a bounded area to guarantee the Lipschitz function condition. Like WGAN, LS-GAN also uses a Lipschitz constraint in which it assumes that the density of real samples is Lipschitz continuous, i.e., nearby data do not suddenly change.

Guo, et al. [31] proposed a new class of statistical divergence, Relaxed Wasserstein (RW) divergence for the large-scale computations. RW divergence is the Wasserstein divergence parametrized by the class of strictly convex and differentiable functions containing different curvature information. RWGANs are robust and converge faster than WGAN. [97] introduced f-GAN to minimize the variational estimate of f-divergence where it formulates training D as a density ratio estimation (like GANs [4]). The main objective of f-GAN is to minimize the f-divergence between the true data distribution p($x$) and the model distribution q($x$). In addition, [98] proposed to use density ratio estimation based on the Bregman divergence, called b-GAN. The objective of b-GAN is the direct estimation of the true density ratio without estimating p($x$) and q($x$) independently. Basically, GANs are trained using a distribution discrepancy measure, such as information-theoretic divergences, integral probability metrics (IPM), and Hilbert space discrepancy metrics. Tao, et al. [99] studied these training metrics in connection and introduced a novel metric, called $\chi$2- GAN, for stable training and reduced mode collapse.

In Original GANs, D uses a sigmoid cross entropy loss for differentiating real samples from the fake samples. If D classifies a generated sample as real, then G will stop updating even though generated sample is far from the real data distribution. In this case, sigmoid cross entropy loss will not push generated samples towards real data distribution as classification role is done. To handle this issue, Mao, et al. [34] proposed the Least Squares Generative Adversarial Networks (LSGANs) that uses least squares loss function (l2 loss) instead of sigmoid cross entropy loss to stabilize the learning process and to reduce the possibility of mode collapse. For handling vanishing gradients problem, samples are penalized on the basis of their distances to decision boundary to generate more gradients to update G. Authors have also shown that minimizing loss function of LSGANs minimizes the Pearson $\chi$2 divergence. However, LSGANs could not achieve good performance for generating diverse images with real datasets. [101] introduced to use Softmax cross-entropy loss, called Softmax GANs, instead of the classification loss in the basic GANs. In basic GANs, D uses a logistic loss which saturates quickly, and its gradient vanishes if it is simple to differentiate between real samples and generated samples. As gradient vanishes, G stops updating. In Softmax GANs, gradient is always non-zero unless the Softmax distribution matches the target distribution. In addition, Saliman, et al. [100] introduced a GANs variant, called Optimal Transport GANs (OT-GAN), for minimizing a newly proposed metric, mini-batch energy distance to compute the distance between the G's generated and real data distribution. However, OT-GAN suffers from large amounts of computation and memory.

Traditional GANs is related to Noise Contrastive Estimation (NCE) in which a binary classification task is defined between true and noise samples with a logistic loss, while Softmax GANs is related to Importance Sampling version of GANs in which Importance Sampling replaces the logistic loss with a multi-class Softmax and cross-entropy loss. Furthermore, [102] introduced to exploit the features learned by D to handle the issue of mode collapse and instability. A reconstruction loss is added with GANs objective function in which real data features generated by D is fed into G for real data generation. Authors claimed that in order to enhance the generated sample quality, proposed reconstruction loss can be applied with other regularization loss function, such as gradient penalty. [103] proposed a GANs training approach in which D is trained using a maximum margin formulation to improve the D's capability, i.e., a better G.

On the other hand, previous GANs training approaches have employed gradient descent on the cost of each player simultaneously in which gradient descent enters a stable orbit instead of converging to the desired equilibrium point [143]. Feature matching can handle GANs training stability by having G's new objective as it avoids GANs from overtraining on current D [36]. The new objective allows G to produce data equivalent to the statistics of the real data instead of directly maximizing the output of D. D is used only to state the statistics that are worth matching. Feature matching is similar to techniques that use *maximum mean discrepancy* [144]–[146] for G's network training [128][129]. Several feature matching based GANs have been proposed with an aim to improve convergence.

Salimans, et al. [36] proposed three techniques for fast convergence of GANs game, called IGAN: *feature matching*, *minibatch features* and *virtual batch normalization* (VBN). A new objective function in feature matching does not work on directly boosting D's output, while it involves G to generate data that matching statistics of the real data and D is used only to identify the statistics worth matching. Feature matching's objective performs well for classification but could not generate indistinguishable samples. In addition, the idea of minibatch features is equivalent to batch normalization [149] while VBN is a direct extension of batch normalization. Minibatch discrimination allows D to process the correlations between training data points in one batch, rather than handling them separately. Minibatch discrimination works well for generating realistic images, but it could not predict labels accurately. Moreover, minibatch discrimination is computationally complex and highly sensitive to the selection of hyperparameters [37]. VBN can handle this issue, but it has high computational complexity as it involves running forward propagation on two minibatches of data which can be used only in G network. In addition, IGAN uses the same network architecture as DCGAN, but training of IGAN is more advanced.

IGAN works only for semi-supervised learning, so for unsupervised learning, [41] proposed to attach G's training criterion with a second training objective which directs G in the direction of data by explicitly modeling the data density in addition to D. The second training objective is calculated in the space of features learned by D to make it computationally inexpensive by not having separate convolutional network for it. Then, in that space, denoising autoencoder is trained to estimate the energy gradient of the data on which it is trained.

The parametrization used in WGAN suffers from *l1* mean feature matching problem. To handle this issue, [104] extended the concept of WGAN by matching second order moment feature through the singular value decomposition concept. [104] used the mean feature matching and covariance matching GANs (McGAN) for stable training and reduced mode collapse and used the Integral Probability Metrics (IPM) loss as it correlates with the generated samples' quality. McGAN also maximizes an embedding covariance discrepancy between real samples and generated samples. McGAN matches second order moment from the primal-dual norm perspective which requires matrix (tensor) decompositions because of exact moment matching [150] and hard to scale to higher order moment matching. In contrast, Maximum Mean Discrepancy GANs (MMDGAN) [46] can match high-order moments with kernel tricks by giving up exact moment matching. MMD distance is derived from the Generative moment matching networks (GMMN) [147] and is an alternative to Wasserstein distance. MMDGAN uses the adversarial kernel learning techniques instead of Gaussian kernels which can better represent the feature space. But still, the computational complexity of MMDGAN increases as number of sample increases. [105] introduced a new manifold-matching GANs (MMGAN) to stabilize GANs training by discovering two manifolds for the vector representations of real and fake images in which real and fake images are statistically identical, if they are equal. To match the two manifolds, authors have proposed a loss function to train G. In addition, for enhancing diversity in generated samples, authors have proposed kernel tricks for getting better manifold structures, moving-averaged manifolds across mini-batches, and a correlation matrix based regularizer.

Bellemare, et al. [106] has pointed that WGAN suffers from the biased gradients. As a solution, they proposed an energy function without the biased gradients, called CramerGAN, where Cramer distance is related to the kernel embedded space distance of MMDGAN.

### 6.2. Regularization

In this sub-section, we shall discuss regularization strategies proposed to stabilize GAN training.

[107] proved that the usage of weight clipping in WGAN for imposing a Lipschitz constraint on the critic generates lower-quality samples or fails to converge. Authors proposed another solution to WGAN's weight clipping which is to penalize the norm of gradient of the critic w.r.to its input, called Wasserstein Generative Adversarial Networks – gradient penalty (WGAN-GP). Further, [30] extended WGAN-GP concept to any separable complete normed space, called Banach Wasserstein GANs (BWGAN). The main difference is that the l2 norm is replaced with a dual norm. In addition, authors generalized the WGAN-GP theory to Banach spaces to allow features selection for G.

On the other hand, [108] have present a theoretical claim that BWGAN is not good for GANs training and then propose a less restrictive regularization. While, [109] have shown that gradient penalty in WGAN-GP cannot support stable training as GP often does not check the continuity of region near the real data. To handle this issue, authors proposed training in WGAN-GP which explicitly checks the continuity condition using two perturbed version of $x$, near any observed real data point $x$.

[110] have shown that in WGAN-GP, weight clipping reduces the rank of the weight matrix which reduces the features used by D to distinguish the distributions. To handle this issue, authors proposed spectral normalization technique which does not affect the rank of the weight matrix and stabilize the training of D networks. Authors have an analysis that spectral normalization for GANs outperforms other regularization techniques, such as weight normalization [151], weight clipping [29], and gradient penalty [107].

A geometric mismatch is not beneficial for f-GAN as the resulting f-divergence is not finite [42]. As a solution [143][144], it is suggested to use an alternative family of distance functions, IPMs which includes Wasserstein distance and RKHS-induced maximum mean discrepancies. On the other hand, [111] has proposed noise-induced regularization scheme to make f-GAN models robust against dimensional misspecifications. [112] proposed a GANs variant for stable training which builds on the IPM framework and normalizes the critic with its second moment, like χ2-GAN. Fisher GANs relies on a more sophisticated augmented Lagrangian to optimize the same objective for both the critic and G, while χ2-GAN decouples the critic and G objectives, requiring simpler (unconstrained) SGD type updates. Fisher GANs reduce the distance between two distributions as well as in-class variance. Moreover, it does not add any weight clipping or gradient penalty.

On the other hand, some researchers introduced to add regularization term to objective function for easily converging the objective function to the global optima.

[32] revisited the basic GANs algorithm for finding the Nash-equilibrium and proposed to add a regularization term to the loss function for handling the non-convergence of simultaneous gradient ascent based on the Jacobian of the gradients for both D and G. Authors have identified theoretically, key elements responsible for the failure of the SGD to achieve Nash-equilibria. Then, authors introduced a new design for the GANs training on the basis of insights driven from the SGD failure analysis. Further, to address the convergence issue in the training of GANs, [39] introduced a surrogate objective for G's update which unrolls D to regularize its updates. But, computational cost of unrolling step is high, which makes it unsuitable for the large-scale datasets. Authors also present two techniques, inference via optimization, and pairwise distance distributions for generating diverse samples.

[113] and [114] introduced a regularized penalty for G and D, respectively, to achieve better convergence in the GANs algorithm. [115] analyzed the convergence of GANs training for finding the main reason(s) of the

mode collapse. Then proposed gradient penalty approach, called DRAGAN to avoid the local equilibria which causes sharp D's gradients nearby some real data points.

**Further discussion.** The selection of appropriate distance measure plays a vital role in the training of generative networks. The major difference between these distances is the impact on the convergence of sequences of probability distributions. As discussed earlier, several GANs variants have been proposed for minimizing the probability distances/divergences between real and generated data distribution, such as JSD, f-divergence, maximum mean discrepancy (MMD) and Wasserstein distance. Some approaches also used probability measure for distance calculation. Cramer distance, a family of IPMs, calculates the distance between probability distributions of generated and real data in D's activations. Divergences based strategies try to ensure the existence of Nash-equilibria.

In most of the works, objective functions are derived from the Wasserstein distance instead of f-divergences for improving the stability in the GANs training. Wasserstein distance does not support the theoretical guarantee, but it is computationally efficient. In addition, moment matching based approaches have handled the issue of vanishing gradients, but these approaches are computationally expensive than training a classifier. While, the use of gradient penalty [107] and spectral normalization [110] are beneficial in the high-capacity architectures. Several solutions have been proposed to regularize the GANs for handling the issue of vanishing gradients and non-convergence where most of the works are theoretically formulated. Gradient-norm based regularization enhance local stability. However, regardless of the progress, stable GANs training is an open challenge which requires a careful balance during the adversarial optimization.

# 7. ALTERNATIVE OPTIMIZATION ALGORITHM

Previous approaches in GANs used gradient descent techniques which mainly focus on finding low value of cost function instead of Nash equilibrium of a game. These algorithms may fail to converge when find the Nash equilibrium [126]. Introducing a stable algorithm for GANs training is a long-awaited problem and numerous solutions have been introduced. In this section, we shall discuss optimization approaches introduced to address the issues in GANs using Simultaneous Gradient Descent.

[32] analyzed the main reasons associated with simultaneous gradient ascent algorithm failure in finding local Nash-equilibria of smooth games. Authors have shown through theoretical analysis that the main factors are the presence of eigenvalues of the Jacobian of the associated gradient vector field with zero real-part and eigenvalues with a large imaginary part. The second case makes saddle-point problems more challenging than local optimization problems. To handle these problems, [32] proposed a robust algorithm using consensus optimization to find Nash-equilibria of smooth two-player games. The proposed algorithm is orthogonal to other strategies making the GANs framework better, such as using new distances, or regularization. These solutions assume the existence of Nash-equilibria while proposed algorithm works with their computation and the numerical difficulties occurring in practice. The proposed solution is an approximation to the implicit Euler method for integrating the gradient vector field where implicit Euler method has strong stability properties [154] which can be translated into convergence theorems for local Nash-equilibria. However, in implicit Euler method, solution of a nonlinear equation in each iteration is required. Moreover, it can be extended by having better approximations to the implicit Euler method. Alternatively, the proposed work can be seen as second order method where it can be extended by revisiting second order optimization methods [155] in the context of saddle point problems. Authors claimed that it is a first step to understand main factors of GANs training and different objective functions. The main aim of the work is to stabilize GANs training on different architectures and divergence functions. The results show that the proposed algorithm achieves stability in the training and handles the issue of mode collapse successfully. However, stability suffers in the case of deeper architectures as gradients can have different scales in such architectures. Also, if regularization parameter set to high, proposed method can make unstable stationary points of the gradient vector field stable and may lead to poor solutions.

Simultaneous Gradient Descent is similar to use no-regret dynamics for each player where in game theory, it is assumed that this limits the oscillatory behavior in zero sum games. In the case of convex-concave zero-sum games, average of the weights of the two players establishes an equilibrium and not the last-iterate. But, average the neural nets weights is not allowed as zero-sum game defined by training two deep networks against each other is not a convex-concave zero-sum game. Therefore, it is important to get training algorithms making the last iterate of the training be very close to the equilibrium, instead of only the average. [116] proposed to use a variant of gradient descent, called Optimistic Mirror Descent (OMD) for WGAN training which achieves faster convergence rate to equilibrium. Gradient Descent (GD) dynamics are bound to cycle while the last iterate of OMD dynamics converges to an equilibrium. OMD considers that in zero-sum game, another player is also training through similar algorithm; OMD predicts the strategy of another player to achieve faster regret rates. Results show that OMD gets smaller KL divergence w.r.to the true underlying distribution compared to GD variants. Experiments have also shown that GD suffers from limit cycles even tested in a simple distribution learning setting (learning the mean of a multi-variate distribution). In addition, OMD converges pointwise. an optimistic variant of Adam is also introduced which shows better performance than Adam.

In recent times, actor-critic learning has been applied for stochastic approximation. [117] used the two time-scale update rule (TTUR) which guarantees that training achieves a local Nash equilibrium when the critic learns faster than the actor. [33] used the same approach and train the GANs by TTUR to reach the local Nash equilibrium where D and G both have different learning rates. The main idea is that D converges to a local minimum when G is fixed. If changes in G are quite slow, still D still converges, since G perturbations are small. In addition, performance also enhances as D must first learn new patterns before transferring to G. On the other hand, Adam stochastic approximation is used to handle the issue of mode collapse where Adam stochastic optimization is introduced as a heavy ball with friction (HBF) dynamics in which Adam try to get flat minima and avoids small local minima. TTUR enhances learning of DCGANs and WGAN-GP and outperforms basic GANs training for various image datasets.

GANs is difficult to train as optimal weights of the loss functions in GANs relate to saddle points, and not minimizers. Alternative SGD methods used for zero-sum games either do not reliably converge to saddle points or if converges, they are highly sensitive to learning rates. Standard alternating SGD methods generally moves between minimization and maximization steps. It is possible that minimization step overpower maximization step where iterates will "slide off" the edge of saddle and it will lead to instability. To handle this issue, [35] proposed modified stochastic gradient descent, called *prediction step* to stabilize adversarial networks in which maximization step can exploit information about minimization step. Authors have also shown that proposed algorithm converges to saddle points and is stable with a wider range of training parameters than plain SGD methods. This supports faster training with larger learning rates. A theoretical proof is also provided to show that prediction step is asymptotically stable for solving saddle point problems.

# 8. Summary

In Table 2, we summarize GANs design and optimization solutions proposed for handling two main GANs challenges, mode collapse (MC), non-convergence and instability (NC&I). The first and second column state the challenges addressed in the paper. The rest columns show the proposed solutions in the paper for handling the addressed GANs challenge(s) (as shown in Table 1).

## Table 2 Summary of addressed GANs challenges and their proposed solutions

| **Models** | **MC** | **NC&I** | **$S_{11}$** | **$S_{12}$(i)** | **$S_{12}$(ii)** | **$S_{12}$(iii)** | **$S_{13}$(i)** | **$S_{13}$(ii)** | **$S_{14}$** | **$S_{15}$** | **$S_{16}$** | **$S_{21}$** | **$S_{22}$** |
|---|---|---|---|---|---|---|---|---|---|---|---|---|---|
| **cGANs [28]** | √ | | √ | | | | | | | | | | |
| **LAPGAN [5]** | √ | | √ | | | | | | | | | | |
| **SGAN [43]** | √ | | √ | | | | | | | | | | |
| **BiCoGAN** [50] | √ | | √ | | | | | | | | | | |
| **MatAN** [51] | √ | | √ | | | | | | | | | | |
| [52] | √ | | √ | | | | | | | | | | |
| **AC-GAN** [53] | √ | | √ | | | | | | | | | | |
| **TripleGAN** [54] | √ | | √ | | | | | | | | | | |
| **KDGAN** [55] | √ | | √ | | | | | | | | | | |
| **ControlGAN** [56] | √ | | √ | | | | | | | | | | |
| **DCGANs [6]** | | √ | | √ | | | | | | | | | |
| **ProgressGAN [8]** | √ | √ | | √ | | | | | | | | | |
| **PacGAN [37]** | √ | | | √ | | | | | | | | | |
| **BayesianGAN** [57] | √ | | | √ | | | | | | | | | |
| **CapsNets** [58] | √ | | | √ | | | | | | | | | |
| **SAGAN** [60] | | √ | | √ | | | | | | | | | |
| **cGAN [44]** | √ | | | | √ | | | | | | | | |
| **AdaGAN [45]** | √ | √ | | | √ | | | | | | | | |
| **MAD-GAN** [61] | √ | | | | √ | | | | | | | | |
| **MGAN** [62] | √ | | | | √ | | | | | | | | |
| **MPM GAN** [63] | √ | | | | √ | | | | | | | | |
| **Fictitious GAN** [64] | √ | | | | √ | | | | | | | | |
| **MIX+GAN** [65] | √ | | | | √ | | | | | | | | |
| **D2GAN** [66] | √ | | | | | √ | | | | | | | |
| **GMAN** [67] | √ | | | | | √ | | | | | | | |
| **StabGAN** [68] | | √ | | | | √ | | | | | | | |
| **Dropout GAN** [69] | √ | | | | | √ | | | | | | | |
| **SGAN** [71] | √ | √ | | | | √ | | | | | | | |
| **VAE-GAN** [72] | √ | | | | | | √ | | | | | | |
| **AAE** [73] | √ | | | | | | √ | | | | | | |
| **AVB** [74] | √ | | | | | | √ | | | | | | |
| **ASVAE** [75] | √ | | | | | | √ | | | | | | |
| **MDGAN [40]** | √ | | | | | | √ | | | | | | |
| **Dist-GAN** [76] | √ | | | | | | √ | | | | | | |
| **α-GAN [25]** | √ | | | | | | √ | | | | | | |

| **Models** | **MC** | **NC&I** | $S_{11}$ | $S_{12}$**(i)** | $S_{12}$**(ii)** | $S_{12}$**(iii)** | $S_{13}$**(i)** | $S_{13}$**(ii)** | $S_{14}$ | $S_{15}$ | $S_{16}$ | $S_{21}$ | $S_{22}$ |
|---|---|---|---|---|---|---|---|---|---|---|---|---|---|
| **ALI** [77]/ **BiGAN** [78] | | | | | | | | √ | | | | | |
| **HALI** [81] | | | | | | | | √ | | | | | |
| **AGE** [82] | √ | | | | | | | √ | | | | | |
| **VEEGAN** [83] | √ | | | | | | | √ | | | | | |
| **MGGAN** [84] | √ | | | | | | | √ | | | | | |
| **EBGAN** [85] | √ | √ | | | | | | | √ | | | | |
| **BEGAN** [86] | √ | √ | | | | | | | √ | | | | |
| **MAGAN** [87] | | √ | | | | | | | √ | | | | |
| **Max-Boost-GAN** [89] | √ | √ | | | | | | | √ | | | | |
| **MemoryGAN** [90] | | √ | | | | | | | | √ | | | |
| **DeliGAN** [91] | √ | | | | | | | | | | √ | | |
| **NEMGAN** [92] | √ | | | | | | | | | | √ | | |
| **DE-GAN** [94] | | √ | | | | | | | | | √ | | |
| **InfoGAN** [95] | | √ | | | | | | | | | √ | | |
| **WGAN [29]** | √ | √ | | | | | | | | | | √ | |
| **LS-GAN** [96] | | √ | | | | | | | | | | √ | |
| **RWGAN [31]** | | √ | | | | | | | | | | √ | |
| **f-GAN** [97]/**b-GAN** [98] | | √ | | | | | | | | | | √ | |
| **χ2-GAN** [99] | √ | √ | | | | | | | | | | √ | |
| **LSGAN [34]** | | √ | | | | | | | | | | √ | |
| **SoftmaxGAN** [101] | | √ | | | | | | | | | | √ | |
| **OT-GAN** [100] | | √ | | | | | | | | | | √ | |
| **IGAN [36]** | √ | √ | | | | | | | | | | √ | |
| **McGAN** [104]**,** | | √ | | | | | | | | | | √ | |
| **MMD GAN [46]** | | √ | | | | | | | | | | √ | |
| **CramerGAN** [106] | | √ | | | | | | | | | | √ | |
| **WGAN-GP** [107] | | √ | | | | | | | | | | | √ |
| **BWGAN [30]** | | √ | | | | | | | | | | | √ |
| **CT-GAN** [109] | | √ | | | | | | | | | | | √ |
| **SN-GAN** [110] | | √ | | | | | | | | | | | √ |
| [111] | | √ | | | | | | | | | | | √ |
| **FisherGAN** [112] | | √ | | | | | | | | | | | √ |
| **[32]** | | √ | | | | | | | | | | | √ |
| **Unrolled GANs [39]** | √ | | | | | | | | | | | | √ |
| [113] | | √ | | | | | | | | | | | √ |
| [114] | | √ | | | | | | | | | | | √ |
| **DRAGAN** [115] | √ | √ | | | | | | | | | | | √ |

# 9. Applications of GANs

Research on GANs is rapidly growing and several GANs variants have been proposed focusing on various aspects of deep learning. These variants have demonstrated the potential of GANs and have shown promise for developing broad number of interesting and useful applications in several research domains, such as image generation, domain transfer, data generation, ethics in AI, etc. In this section, we shall point out main GANs applications within these major research domains.

## 9.1. Image Generation

Recently, GANs has gained more and more momentum for generating naturalistic images through adversarial training. [156] studied the problem of image generation from different points of view. Authors claimed that visual appearance of objects is significantly influenced by their shape geometry which has not been considered by any of the existing works for image generation. To consider this into account, a novel Geometry-Aware GANs model, called GAGAN is proposed which incorporates geometric information into the image generation process. [157] proposed an adversarial image generation model, called LR-GAN to generate sharp images by considering both scene structure and context. Unlike existing models for image generation, LR-GAN combines foregrounds on the background in a contextually relevant style for generating more realistic images. Results have shown that LR-GAN outperforms DCGAN. [158] proposed Style and Structure GAN (S2-GAN) model in which a GANs is used to produce the image structure and then output is fed into the second GANs for considering the image style. Generation of realistic images has wide range of practical applications, such as anime character generation [159]–[164], image synthesis [165]–[168], super resolution [10], [124], [169]–[177], image editing and blending [178], [179], inpainting [125], [180]–[182], interactive image generation [183], [184], human pose estimation [185], [186], face aging [187], [188], 3D object detection [189]–[192], etc.

## 9.2. Domain Transfer

Image-to-image translation maps an image from one domain to a corresponding image in another domain. Existing researches for image-to-image translation have used supervised setting as shown in Figure 26 (left side), paired images ($x_i$, $y_i$) [158], [193]–[196]. But, availability of the paired data is either difficult or expensive. Only few datasets are available for some tasks, like semantic segmentation [197], but they are quite small. Therefore, in recent times, several researchers have proposed to translate between domains without paired dataset (see Figure 26 (right side), unpaired data (X, Y)) by exploring the underlying relationship between the domains. Image-to-image translation is categorized into two parts: unsupervised image-to-image translation, and semi-supervised image-to-image translation.

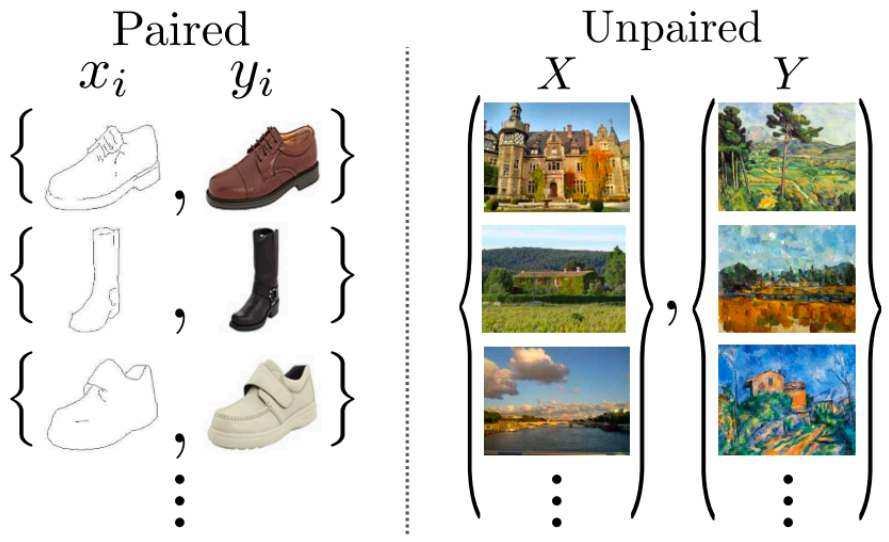

**Figure 26. Paired data vs Unpaired data**

Unsupervised Image-to-image translation problem is quite challenging as unpaired data has no corresponding images. Basic GANs is designed to learn the data distribution of a domain, but image-to-image translation requires matching the joint distribution of image-image pairs. Conditional GANs can only be used to produce images of different style rather than producing images in two different domains, such as color and depth image domains. In recent times, several researchers have explored both unsupervised [7], [198]–[207] and semi-supervised image-to-image translation [208]–[210].

### 9.3. Sequential Data Generation

It is very beneficial to have large labelled dataset for better predictive learning through neural networks. But, labelling large datasets is time consuming and expensive. One possible solution proposed by Simulated + Unsupervised (S+U) learning approach [201], in which it suggests to generate synthetic images from unlabelled real data. Moreover, S+U preserves annotation information for training of machine learning models.

In addition, the task of generating music is quite different than image and video generation as it is only temporal data. Moreover, music is composed of various instruments with different temporal dynamics. [211] proposed a GANs based framework for symbolic multi-track music generation. [212] introduced a GANs based model to produce melodies either from scratch, by following a chord sequence, or by conditioning on the melody of previous bars. Moreover, authors also claimed that proposed model can be expanded to produce music with multiple MIDI channels (i.e., tracks).

On the other hand, enhancing speech is one of the possible solutions for improving speech quality in noisy environments. Recently, deep neural networks have been used to learn complex functions from large example sets and have performed well in compared to traditional approaches for speech enhancements. [213] explored the basic feats of conditional GANs to learn mapping between spectrogram of noisy speech and its enhanced counterpart. [214] proposed to operate at the waveform level and trained an end-to-end model for 28 speakers and with 40 different noise conditions.

### 9.4. GANs for Ethics in AI

Ethics in AI has become one of the significant areas of research. Deep learning based software can cause threats to privacy, democracy and national security [215]. Recent developments in generative data modeling have raised concern about several ethical issues, such as generating fake content (i.e., deepfake), privacy breaching, biased outcomes (i.e., fairness), etc. GANs has also been explored for identifying the fake contents, handling privacy issues in the data generation, and handling bias in the data. In this sub-section, we shall point out several GANs based solutions introduced to handle mentioned ethical issues.

#### 9.4.1. Deepfakes

GANs generates high-quality data (text, image, video, speech) which can be used to generate fake information, i.e., deepfakes. Deepfakes algorithms generate fake images and videos whose authenticity cannot be distinguished from the real data. Deepfakes mainly have been explored for facial manipulation which can be classified into three categories, (1) face synthesis is about creating non-existent realistic faces using GANs [216][217]; (2) face swap is about swapping faces [218]; (3) facial attributes and expression [199] is about manipulating attributes of the face, such as color tone, age, gender, etc. There is a need of algorithms that can detect fake content automatically. [215] proposed a novel solution to detect deepfakes by identifying subtle visual artifacts in the image.

#### 9.4.2. Handling privacy issues in data generation

GANs has also been used for generating synthetic data which can be made available publicly instead of real data. However, an adversary can get the training set membership through the GANs model and generated synthetic data. A model memorizes the training samples which leads the issue of privacy and makes the GANs model vulnerable. [219] have shown that it is easy to identify the training samples through the observation that D is more likely to classify training samples as real rather than samples not present in the training data. To achieve the privacy, records in the data can be de-identified, but now de-identified records can be re-identified by relating them to other identifiable datasets [220]. [221] proposed Differentially Private Generative Adversarial Network (DP-GAN) to guarantee differential privacy of D by introducing noise during the model optimization. But, proposed model does not maintain trade-off between sample quality and diversity while supporting differential privacy. Hence, model is not useful for practical applications. [222]–[224] proposed an architecture design maintaining trade-off between privacy and sample quality.

#### 9.4.3. Fairness

Recently, achieving fairness in learning models have gained momentum in machine learning for several applications, such as granting loan, shortlisting candidate for interview, etc. Decision making models still may suffer from unwanted discrimination against sensitive attributes, such as gender, race, age, etc. To achieve fairness in the outcome, it is important to get the unbiased training data. How to get fair training data, it is an important research problem.

[225] proposed an GANs-based approach to generate fair dataset w.r.to sensitive attributes in allocative decision making from the real dataset for model training. [226] proposed reinforcement learning based race balance network to handle the bias in the data. [227] pointed out that existing deep learning models for face recognition encode gender information implicitly. To handle this, an algorithm is introduced to reduce gender information from the face descriptors. [228] introduced a new direction for handling the issue of missing fairness in the outcome. Authors proposed to achieve fairness through representation learning as it removes the semantics of the sensitive attributes. [229] identified that existing deep learning classifiers trained to generate diagnostic labels from X-ray images are biased w.r.to sensitive attributes. [230] introduced a framework based on conditional GANs for generating synthetic fair data with selective properties from real data. [231] studied the fairness from the perspective of causal relations in the data.

### 9.5. Others

Steganalysis allows to find any payload hidden in the message and retrieving it if possible. An efficient classifier is required to classify the presence of hidden payloads. [232] proposed a DCGAN based approach to produce more setganalysis-secure message embedding. Detection of small objects is a challenging task because of object's low resolution and noisy representation. Several researches have proposed to identify small objects by learning representations of all objects at multiple scales. But, these solutions are computationally expensive. To handle this issue, [233] proposed a new Perceptual Generative Adversarial Network [234] model for detecting small objects by reducing the representation difference between small objects and large ones. [235] proposed to use GANs for the information retrieval. [236] used GANs for automatically generating accurate driving scenes for testing the reliability of DNN-based autonomous driving systems. [237] proposed to produce GANs based malware which can sidestep the black-box machine learning based detection models. [238] proposed to use GANs for predicting the crystal structure.

## 10. DISCUSSION AND RESEARCH DIRECTIONS

According to the proposed taxonomy (in Table 1), we can see that the rapidly growing body of research focused on GANs now comprises several novel solutions for handling the challenges of GANs. Here, we look at what could be explored in future research to further improve the existing works.

### 10.1. GANs Design and Optimization Solutions

GANs has shown its potential for generating natural images, still it goes through the problem of mode collapse as discussed earlier, i.e., G collapses and could capture only limited varieties of modes in the data. GANs commonly fails to learn some of the modes trained on the multi-modal distribution data. Most of the time, GANs could not converge to the true equilibrium and settles for sub-optimal local solutions. Due to mode collapse, samples produced by the trained generative model often lack diversity. In addition, we can say that mode collapse is connected to the instable GANs training which leads to another main challenge in GANs.

Despite the promising achievements, GANs is still hard to train due to several common problematic unstable training and convergence behaviors, such as vanishing gradients, mode collapse, and diverging or oscillatory behavior. These issues in GANs training often hinders further research and applicability in this area. In recent times, to alleviate the above-mentioned issues, existing studies have proposed several solutions, such as designing more stable network architectures, modifying the learning objectives, regularize objectives, training strategies, tuning hyperparameters, etc. However, in most of the cases, their achievement is often the result of sacrificing the image quality and image diversity where these issues have trade-off relationships.

Most of the existing works have focused either on image quality or on image diversity. One possible research direction can be to work on image quality without suffering from the low image diversity. Furthermore, to handle the issue of GANs training instability, existing methods still depend on heuristics which are very sensitive to amendments. This is one of the main reasons that restricts applicability of these approaches in new domains. On the other hand, we observe that most of the existing works have proposed to solve only one training issue at a time and often do not include theoretical analysis. Another important research direction is to have a theoretical framework/analysis for handling issues in GANs training process with the aim of exploring more tractable formulations and to make training stable and straightforward.

In addition, proposed solutions differ in the training improvement scale. Most of the models can achieve similar results by improving the hyperparameters settings and computational resources. Therefore, we can say that majority of related works mainly emphasized on achieving state-of-the-art accuracy instead of state-of-the-art efficiency. Developing solutions having the algorithmic improvements over existing works can be a future direction.

To handle the GANs challenges, existing approaches are focused on three major directions: re-engineered network architecture, new objective functions and optimization algorithms as discussed in Section 5, 6 and 7. Objective function GANs variants often show more improvement in training than architectural GANs but still they cannot boost mode diversity in the generated visual samples. Several GANs design and optimization solutions have been proposed in these three directions where proper selection of architecture, objective functions and optimization techniques have improved GANs training stability. In addition, objective functions are sensitive to the use of optimization strategies, hyperparameter selection, and number of training steps which can be explored in the future research for the different GANs.

On the other hand, research on using other technologies for improving the GANs training, such as online learning [239], game theory variants, etc., is still at the early stage. A combination of proper architecture, loss function and optimization techniques can prove to yield superior results and can be a future research direction to explore.

### 10.2. Applications

Generation of realistic images has broad range of practical applications, such as face aging, face editing, pose estimation, entertainment, etc. Recently, GANs has gained momentum for generating naturalistic images through adversarial training. Apart from it, GANs has also shown its potential for several other applications, such as Steganalysis [240], information retrieval [235], spatio-temporal data prediction, such as transportation [241], autonomous driving [236], [242], [243], speech enhancement [214], single-cell RNA-sequence imputation [244], etc. Exploring GANs applicability for new application domains can be a future direction. Moreover, the support of GANs for maintaining ethics in AI is new research direction.

### 10.3. Evaluation Metrics

Generative models were evaluated based on the model likelihood which is intractable. Basic GANs used a proxy for log-likelihood, i.e., Parzen window estimate for evaluation [245]. Theis, et al. [24] presented that likelihood estimation is often incorrect for Parzen window estimate. GANs lacks the meaningful measure for evaluating the GANs output. Due to this issue, it is very challenging to compare different GANs variants and still based on the visual assessment of generated images.

Furthermore, because of the lack of robust and consistent metrics, it is difficult to evaluate which GANs algorithm(s) outperforms other algorithms. A good evaluation is needed as it will allow to choose appropriate algorithm from a very large set. Also, to have better algorithms and their understanding, like which modifications are critical, and which algorithms cannot make a significant difference in practice [245]. To overcome above-mentioned issues, researchers have proposed several evaluation methods for GANs [154][155]. Moreover, different applications choose different evaluation metrics as different applications require different trade-offs between different measures. It is better to know a combination of training and evaluation metrics for the target application.

# 11. SUMMARY AND CONCLUSION

Recently, GANs has gained significant attention for generating realistic images and has become important in modern world applications, such as image generation, domain adaptation, etc. However, GANs is hard to train and training faces two main challenges, mode collapse, non-convergence and stability. The possible solutions to handle these GANs challenges are to design an efficient model by choosing appropriate network architecture, by using suitable objective function or by selecting proper optimization techniques. Within these solutions, many different GANs variants have already been proposed with diverse characteristics, but still some issues remained unsolved.

Research on GANs is quite broad and several designing and training solutions of GANs handling these challenges lay ahead. In this paper, we recapitulate the basic GANs framework and survey the developments of solutions for better design and optimization of GANs. More concretely, we proposed a novel taxonomy of GANs design and optimization techniques based on re-engineered network architectures, new objective functions and alternative optimization algorithms and discussed how existing works deal with these challenges. We mapped the existing works to the taxonomy for finding the research gaps in GANs. Our work provides a panorama of current progress and an in-depth analysis of the reviewed methods to serve both novices and experienced researchers. In addition, the new taxonomy aims to build a problem-solution structure with a hope to suggest a guideline when readers are selecting their research topics or developing their approaches. Based on insights gained, we proposed promising directions that the researchers can pursue in the future.

# SUPPLEMENTARY MATERIAL

## Generative Adversarial Networks (GANs): Challenges, Solutions, and Future Directions

# 5. Re-engineered Network Architecture

## 5.1 Conditional Generation

**Table 1 Loss function of conditional generation**

| Model | Loss Type | Loss Function | Description |
|---|---|---|---|
| **cGANs** [1] | Conditional Loss | $\mathbb{E}_{x\sim p_{data}}\left[log\left(D(x,c)\right)\right] + \mathbb{E}_{z\sim p_z}\left[\log\left(1 - D(G(z,c),c)\right)\right]$ | $c$ is a condition. |
| **IRGAN** [5] | Conditional loss + Regularization | $\mathbb{E}_{x\sim p_{data}(x)}\left[log\left(D(x,c)\right)\right] + \mathbb{E}_{z\sim p_z, c\sim p_{data}(c)}\left[\log\left(1 - D(G(z,c),c)\right)\right] + R(G)$<br>$R(G) = -\lambda \mathbb{E}_{c\sim p_{data(c)}, x\sim G(z,c)}[\log Q(c\mid x)]$ | $R(G)$ is defined by the lower bound [37] of the mutual information $I\left(c;\ G(z,c)\right)$ where $c$ is sampled from the data distribution rather than a pre-defined distribution. Approximator $Q(c\mid x)$ measures $p(c\mid x)$ from $x$ |
| **IcGAN** [2] | Squared reconstruction loss | $L_{Ez} = \mathbb{E}_{z\sim p_z, y'\sim p_y}\left\lVert z - E_z\left(G(z,y')\right)\right\rVert_2^2$<br>$L_{Ey} = \mathbb{E}_{x,y\sim p_{data}}\left\lVert y - E_y(x)\right\rVert_2^2$ | $E$ is composed of two independent sub-encoders: $E_z$ for encoding an image to $z$, and $E_y$ for encoding an image to $y$. $P_y$ is density model to sample generated labels y' for generated data x'. |
| **BiCoGAN** [3] | Extrinsic factor loss (EFL) | $\mathbb{E}_{x\sim p_{data(x)}}\left[\log\left(D(E(x),x)\right)\right] + \gamma\, \mathbb{E}_{(x,c)\sim p_{data(x,c)}}\left[EFL\left(c, E(x)\right)\right] + \mathbb{E}_{z\sim P_{\tilde{z}}(\tilde{z})}\left[\log\left(1 - D\left((\tilde{z}), G(\tilde{z})\right)\right)\right]$ | $\tilde{z}$=[z c]<br>EFL is an explicit mechanism used to guide BiCoGANs for encoding extrinsic factors efficiently. |
| **MatAN** [4] | MatAN loss | $\mathbb{E}_{y_1,y_2\sim p_{data(x,y,t)}} \log D(T_1(\hat{y}), T_2(\hat{y}), \theta_M, b) + \mathbb{E}_{y_1,x\sim p_{data(x,y,t)}} log\left(1 - D\left(T_1(\hat{y}), T_g\left(G(x,\theta_G)\right), \theta_M, b\right)\right)$ | $T_1$, $T_2$, $T_g$ might be the identity transformation, depending on the $T_i()$ configuration, $\theta_M$ id for Siamese network and $b$ is a trained bias. |
| **AC-GAN** [6] | Conditional loss | $L_S = \mathbb{E}_{x\sim p_{data}}[\log D(x)] + \mathbb{E}_{z\sim p_z}\left[\log\left(1 - D\left(G(\mathrm{z})\right)\right)\right]$<br>$L_C = \mathbb{E}_{x\sim p_{data}}[\log D(x,c)] + \mathbb{E}_{z\sim p_z}\left[\log\left(1 - D(G(\mathrm{z},\mathrm{c}),c)\right)\right]$ | Log-likelihood of the correct source, $L_S$, and the log-likelihood of the correct class, $L_C$.<br>D is trained to maximize $L_S + L_C$ while G is trained to maximize $L_C - L_S$. |
| **TripleGAN** [61] | Standard supervised loss (i.e., cross-entropy loss) | $E_{(x,y)\sim p_{data}(x,y)}[logD(x,y)] + \alpha E_{(x,y)\sim p_c(x,y)}\left[log\left(1 - D(x,y)\right)\right] + (1 - \alpha)E_{(x,y)\sim p_z(x,y)}\left[\log\left(1 - D(G(y,z),y)\right)\right] + R_L$<br>$R_L = E_{(x,y)\sim p_{data}(x,y)}[-\log p_c(y\mid x)]$ | $\alpha \epsilon (0,1)$ is a constant that controls the relative importance of generation and classification. |
| **KDGAN** [62] | Distillation loss | $\mathbb{E}_{y\sim p_{data}}[\log D(x,y)] + \alpha \mathbb{E}_{y\sim p_c}\left[log\left(1 - D(x,y)\right) + (1-\alpha)\mathbb{E}_{y\sim p_t}\left[log\left(1 - D(x,y)\right) + \beta L_{DS}^c\left(C(y\mid x), T(y\mid x)\right) + \gamma L_{DS}^t\left(T(y\mid x), C(y\mid x)\right)\right.\right.$ | $L_{DS}^t$ and $L_{DS}^c$ are distillation loss.<br>$\alpha \in (0,1), \beta \in (0,+\infty), \gamma \in (0,+\infty)$ |
| **ControlGAN** [9] | Classification loss | $\theta_D = argmin\left\{\alpha\, L_D\left(t_D, D(x;\theta_D)\right) + (1 - \alpha)\, L_D\left((1-t_D), D(G(z,l;\theta_G);\theta_D)\right)\right\}$<br>$\theta_G = argmin\left\{\gamma_t . L_C\left(l, G(z,l;\theta_G)\right) + L_D\left(\left(t_D, D(G(z,l;\theta_G),\theta_D)\right)\right)\right\}$<br>$\theta_C = argmin\left\{L_C(l,x;\theta_C)\right\}$ | $l$ is the binary representation of labels of sample $x$ and input data for the G, $t_D$ is the label for D and set to 1, and α denotes a parameter for the D.<br>$\gamma_t$ is a learning parameter. |

**Further discussion.** Table 1 shows loss function of conditional GANs. Conditional GANs mainly handling the mode collapse challenge by conditioning the model on additional information. In addition, generally cGANs are based on the supervised learning [1][2][3][4]. We also find that conditional GANs has been used for the several applications, such as image generation (IG), image tag recommendation, compression, etc., which shows that vast number of applications require that data is conditionally generated from certain data mode(s).

We have already discussed in Section 3 that selection of right evaluation metric is critical factor for making the right conclusion just like selection of a training method is crucial for accomplishing high performance in an application. [1][5] have used Log-likelihood (or equivalently KL divergence), a de-facto standard for training and evaluating generative models. It is considered that model with maximum likelihood generates better samples. [6] have explored most widely adopted Inception Score (IS) [7] and Multi-Scale Structural Similarity Index (MS-SSIM) [8] metric for GANs evaluation. IS mainly captures the two main properties of generated samples: highly classifiable and diverse w.r.to class labels. This score shows correlation with the quality and diversity of generated samples. MS-SSIM is one of the image quality measures which only consider visual fidelity and do not consider diversity of samples. Precision, Recall and F1-score are applied for computing degree of overfitting in GANs. [6] has used SSIM as evaluation measure which only consider visual fidelity instead of diversity of generated samples. While, [9] used human evaluation as evaluation measure which cannot completely measure the capacity of models.

## 5.2 Generative-discriminative Network Pair

### 5.2.1 Training of Single Generator

**Table 2. Loss function of single Generator based GANs**

| Model | Loss Type | Loss Function | Description |
|---|---|---|---|
| **LAPGAN** [63] | -- | $\mathbb{E}_{x,l\sim p_{data}}[\log D(x,l)] + \mathbb{E}_{z\sim p_z, l\sim p_l}\left[\log\left(1 - D\big(G(z,l)\big)\right)\right]$ | Used cGANs approach at each level $l$ of the pyramid.<br>$p_l$ is the prior distribution over classes |
| **DCGANs** [13] | Cross Entropy Loss | $\mathbb{E}_{x\sim p_{data}}[\log D(x)] + \mathbb{E}_{z\sim p_z}\left[\log\left(1 - D\big(G(z)\big)\right)\right]$ | -- |
| **SGAN** [64] | Conditional Loss (G) + Entropy Loss (G) | $\mathbb{E}_{h_i\sim p_{data,E}}\left[-\log D_i(h_i)\right] + \mathbb{E}_{z_i\sim p_{z_i}, h_{i+1}\sim p_{data,E}}\left[-\log\left(1 - D_i\big(G_i(h_{i+1}, z_i)\big)\right)\right] +$<br>$\mathbb{E}_{z_i\sim p_{z_i}, h_{i+1}\sim p_{data,E}}\left[f\big(E_i\big(G_i(h_{i+1}, z_i)\big), h_{i+1}\big)\right] +$<br>$\mathbb{E}_{z_i\sim p_{z_i}}[\mathbb{E}_{\hat{h}_i\sim G_i(\hat{h}_i \mid z_i)}\left[-\log Q_i\big(z_i \mid \hat{h}_i\big)\right]$ | $h$ is hierarchical level<br>$f$ is a Euclidean distance measure for intermediate representations and cross-entropy for labels.<br>SGAN used an auxiliary distribution $Q_i\big(z_i \mid \hat{h}_i\big)$ to approximate the true posterior $P_i\big(z_i \mid \hat{h}_i\big)$. |
| **ProgressGAN** [14] | WGAN-GP loss and LSGAN loss | -- | -- |
| **PacGAN** [15] | Total Variation (TV) Distance | $L_D = \log\left(1 + \exp\big(-\mathrm{D}(x)\big)\right) + \log\left(1 + \exp\big(\mathrm{D}(x')\big)\right)$<br>$L_G = \log\left(1 + \exp\big(\mathrm{D}(x)\big)\right) + \log\left(1 + \exp\big(-\mathrm{D}(x')\big)\right)$ | $x'$ is generated data.<br>Packed samples as a single sample that is drawn from the product distribution. |
| **BayesianGAN** [17] | -- | $\theta_D^{j,m} = \theta_D^{j,m} + v_D$<br>$\theta_G^{j,m} = \theta_G^{j,m} + v_G$<br>$v_D = (1-\alpha)v + \eta\left(\sum_{i=1}^{J_D}\sum_{k=1}^{J_G}\frac{\partial\, logp\left(\theta_D \mid z^i, x, \theta_G^{k,m}\right)}{\partial\theta_D}\right) + \mathrm{n}$<br>$v_G = (1-\alpha)v + \eta\left(\sum_{i=1}^{J_G}\sum_{k=1}^{J_D}\frac{\partial\, logp\left(\theta_G \mid z^i, x, \theta_D^{k,m}\right)}{\partial\theta_G}\right) + \mathrm{n}$<br>$\mathrm{n} \sim \mathcal{N}(0, 2\alpha\eta I)$ | α is the friction term for Stochastic Gradient Hamiltonian Monte Carlo (SGHMC), η is the learning rate.<br>$J_G$ and $J_D$ are simple MC samples for the G and D, respectively, and $m$ is SGHMC samples for each simple MC sample. |
| **CapsNets** [18] | Margin Loss | $L_D = \mathbb{E}_{x\sim p_{data}}\left[-L_M(D(x), T=1)\right] + \mathbb{E}_{z\sim p_z}\left[-L_M\big(D\big(G(z)\big), T=0\big)\right]$<br>$L_M = \sum_{k=1}^{K} T_k \max(0, m^+ - \lVert V_k\rVert)^2 + \lambda(1 - T_k)\max(0, \lVert V_k\rVert - m^-)^2$<br>$L_G = \mathbb{E}_{z\sim p_z}\left[-L_M\big(D\big(G(z)\big), T=1\big)\right]$ | $T_k$ represents target labels.<br>$m^+$, $m^-$ and λ are down-weighting factor to prevent initial learning from shrinking lengths of the capsule outputs in the final layer.<br>$V_k$ is vector outputs of the final layer. |
| **SAGAN** [19] | Hinge version of the adversarial loss | $L_D = -\mathbb{E}_{(x,y)\sim p_{data}}\left[min\big(0, -1 + D(x,y)\big)\right] - \mathbb{E}_{z\sim p_z, y\sim p_{data}}\left[min\big(0, -1 - D(G(z), y)\big)\right]$<br>$L_G = \mathbb{E}_{z\sim p_z, y\sim p_{data}}[D(G(z), y)]$ | -- |

**Further discussion.** Normalization of D is valuable from both the optimization (more efficient gradient flow, a more stable optimization) and representation perspective. In GANs literature, several techniques have been introduced from the optimization perspective, namely batch normalization (BN) [10] and layer normalization (LN) [11]. DCGAN popularized the use of Batch normalization in the GANs framework, while Layer normalization is used in GANs framework by [12].

Table 2 shows loss function of single generator based GANs. Current approaches have mainly targeted the development of solutions for the mode collapse challenge using different architectural layers and networks. In addition, the scale at which each of the proposed architecture improves the stability of the GANs is quite different. We find that most of the works [13][14][15][16]–[19] of the developing new architectural GANs has mainly focused on the image generation (IG) and classification applications. DCGAN has demonstrated better training stability and is one of the most popular GANs variants in the literature. DCGANs has used the classification error rate for the evaluation where this metric deeply depends on the selection of a classifier. DCGANs has make the use of nearest neighbor classifiers where Euclidean distance is not a suitable dissimilarity measure for images. [16] have proposed a new evaluation metric, Generative Adversarial Metric (GAM), to evaluate the model quantitatively.

### 5.2.2 Training of Multiple Generators

**Table 3. Loss function of multiple Generators based GANs**

| Model | Loss Type | Loss Function | Description |
|---|---|---|---|
| **cGAN** [65] | Cross Entropy loss | $Q(x_k) = \begin{cases} 1 & D(x) > t_r \\ 0 & else \end{cases}$ | Used gate-function Q for redirecting data to the next GANs, $Q(x_k) = 1$ means transfer the data to the next GANs, otherwise not. |
| **AdaGAN** [66] | Binary Cross Entropy loss | $\mathbb{E}_{x \sim p_{data}}[logD(x)] + \mathbb{E}_{z_1 \sim p_{z_1(z_1)}}\left[\log\left(1 - D\left(G_1(z_1)\right)\right)\right] + \mathbb{E}_{z_2 \sim p_{z_2(z_2)}}\left[\log\left(1 - D\left(G_2(z_2)\right)\right)\right]$ <br> $\mathbb{E}_{x \sim p_z(z), enc[msg(G_2(z_2, m_1)), z_2, m_1]]}\left[\log\left(1 - D\left(G_1(x)\right)\right)\right] - f\left(D\left(G_1(x)\right)\right) - D\left(G_2(x)\right)$ | For $G_1$, and similar version for $G_2$. |
| **MAD-GAN** [67] | Multi-label cross entropy loss | $\mathbb{E}_{x \sim p_{data}}\left[\phi\left(D_v(x)\right)\right] + \mathbb{E}_{x \sim p_z}\left[\phi\left(1 - D_v(x)\right)\right]$ | $\phi$ is a measuring function $\phi: [0,1] \rightarrow \mathbb{R}$, $u \in U$ and $v \in V$ |
| **MGAN** [68] | Standard Softmax loss | $\mathbb{E}_{x \sim p_{data}}\left[\log D_{k+1}(x; \theta_d)\right] + \mathbb{E}_{z \sim p_z}\left[\log\left(1 - D_{k+1}\left(G_i\left(z; \theta_g^i\right); \theta_d\right)\right]$ | Dj(x; $\theta_d$) is *j*th index of D(*x*; $\theta_d$) for which δ(*j*) = 1. A standard softmax loss is used for a multi-classification to maximize the entropy for the classifier. |
| **MPM GAN** [69] | -- | $\mathbb{E}_{x \sim p_{data}}[\log\ D(x)] + \mathbb{E}_{x \sim p_z}\left[\log\left(1 - D(x)\right)\right] - \beta\left\{\sum_{k=1}^{K} \pi_k \mathbb{E}_{x \sim p_{G_k}}[\log C_k(x)\right\}$ | Each $G_k$ maps *z* to *x* = $G_k(z)$, thus inducing a single distribution $p_{G_k}$. $C_k(x)$ is the probability that *x* is generated by $G_k$ and $\beta > 0$ is the diverse hyperparameter. |
| **Fictitious GAN** [70] | -- | $\mathbb{E}_{x \sim p_{data}}\left[f_0\left(D(x)\right)\right] + \mathbb{E}_{x \sim p_z}\left[f_1\left(D\left(G(x)\right)\right)\right]$ | $f_0(\cdot)$ and $f_1(\cdot)$ are some quasi-concave functions depending on the GANs variants. |
| **MIX+GAN** [71] | -- | $\mathbb{E}_{x \sim p_{data}}\left[\log\left(D_i(x)\right)\right] + \mathbb{E}_{z \sim p_z}\left[\log\left(1 - D_i\left(G(\text{z})\right)\right)\right]$ | -- |

**Further discussion.** Table 3 shows loss function of multiple generators based GANs. Most of the current approaches of the multiple Gs have proposed solutions for mitigating the mode collapse problem. Majority of the work have evaluated the proposed multi-agent architecture for the image generation application. Proposed solutions have shown promising results but not state-of-the-art results. Most of the works have used the well-known evaluation metric, Inception Score (IS).

### 5.2.3 Training of Multiple Discriminators

**Table 4. Loss function of multiple Discriminators based GANs**

| Model | Loss Type | Loss Function | Description |
|---|---|---|---|
| **D2GAN** [21] | Symmetric KL divergence and Wasserstein distance | $\alpha\ \mathbb{E}_{x \sim p_{data}}[\log D_1(x)] + \mathbb{E}_{z \sim p_z}[-D_1(G(z))] + \mathbb{E}_{x \sim p_{data}}[-D_2(x)] + \beta\ \mathbb{E}_{z \sim p_z}[log\ D_2(G(z))]$ | 0< $\alpha, \beta \leq 1$ |
| **GMAN** [20] | Cross Entropy loss | $\mathbb{E}_{x \sim p_{data}}[\log D(x)] + \mathbb{E}_{z \sim p_z}\left[\log\left(1 - D(G(\mathrm{z}))\right)\right]$ | -- |
| **StabGAN** [72] | -- | $\sum_{i=k}^{K} \mathbb{E}_{x \sim p_{data}}[\log D_k(W_k^T x)] + \mathbb{E}_{Z \sim pZ}\left[\log\left(1 - D_k\left(W_k^T\ G(z)\right)\right)\right]$ | $W_k, k \in \{1, \ldots, K\}$ is a randomly chosen matrix in $\mathbb{R}^{d*m}$ with $m < d$. |
| **Dropout GAN** [73] | -- | $\sum_{i=k}^{K} \delta_k \left(\mathbb{E}_{x \sim p_{data}(x)}[logD_k(x)] + \mathbb{E}_{z \sim p_z(z)} log\left(1 - D_K(G(z))\right)\right)$ | $\delta_k$ is a Bernoulli variable ($\delta_k$ ~ Bern (1 − d)) and $\{D_k\}$ is the set of $K$ total Ds. |
| **SGAN** [22] | Wasserstein loss | $L_G = \mathbb{E}_{z \sim p_z}\left[\log\left(D(G(\mathrm{z}))\right)\right]$ | Objective function for $L_D$ is not defined. |

**Further discussion.** Table 4 shows loss function of multiple discriminators based GANs. Most of the works have used the well-known evaluation metric, Inception Score (IS). [20] has proposed a new metric, called Generative multi-adversarial metric (GMAM). GMAM is an extension of GAM [16] to train multiple Ds. [21] used mode score which is quite similar to IS metric. Mode score also consider prior distribution of the labels over training data. [22] have used Fréchet Inception Distance (FID), a metric to calculate the Wasserstein-2 distance between the generated and real distributions to find the generated samples quality.

## 5.3 Joint Architecture

### 5.3.1 Data space Autoencoders

**Further discussion**. Recently, three classes of algorithms, (Variational) Auto Encoders (AEs and VAEs), autoregressive approaches and deep GANs have shown potential for learning deep directed generative models. Autoregressive approaches model the relationship between input variables directly and produce outstanding samples. However, autoregressive models [23]–[25] suffer from the slow sampling speed. GANs can learn a unidirectional mapping to generate samples from the data distribution without foregoing sampling speed and also makes use of a latent representation in the generation process while VAEs includes both generation and inference as it learns a bidirectional mapping between a complex data distribution and simple prior distribution. The combination of VAE and GANs come with several benefits, such as solutions can be used to reconstruct data, inference network supports representation learning, etc. Models based on the combination of VAE and GANs can be used for unsupervised, supervised and reinforcement learning.

## Table 5. Loss function of data space autoencoders

| Model | Loss Type | Loss Function | Description |
|---|---|---|---|
| **VAE-GAN** [27] | Prior regularization term + reconstruction error (D) | $L_{GAN} = \log\big(D(x)\big) + \log\Big(1 - D\big(Dec(z)\big)\Big) +$<br>$\log\Big(1 - D\Big(Dec\big(E(x)\big)\Big)\Big)$<br>$L_{prior} = D_{KL}\big(q(z \mid x) \Vert p(z)\big)$<br>$L_{llike}^{D} = -\mathbb{E}_{q(z \mid x)}[\log p(D_l(x) \mid z)]$ | $D_l(x)$ denote the hidden representation of the *l*th layer of the D. |
| **AAE** [28] | Reconstruction error + adv. training | -- | Adversarial training procedure in which q(z) is matched to the whole distribution of p(z) |
| **AVB** [29] | -- | $L_D = \max_D \mathbb{E}_{p_d(x)} \mathbb{E}_{q_\phi(Z \mid x)} \log \sigma\big(D(x,z)\big) +$<br>$\mathbb{E}_{p_d(x)} \mathbb{E}_{p(z)} log\Big(1 - \sigma\big(D(x,z)\big)\Big)$ | Objective for the $D(x,z)$ for a given $q_\phi(z \mid x)$ |
| **ASVAE** [31] | -- | $\min_{\theta,\phi} \max_{\psi_1,\psi_2} -\mathcal{L}_{VAE_{xz}}\big(\theta,\phi,\psi_1,\psi_2\big)$<br>$\mathcal{L}_{VAE_{xz}}(\theta,\phi,\psi_1,\psi_2) =$<br>$\mathbb{E}_{x \backsim q(x), \epsilon \backsim \mathrm{N}(0,I)}\Big[f_{\psi_1}\Big(x, z_\phi(x,\epsilon)\Big) -$<br>$log_{p_\theta}\Big(x \mid z_\phi(x,\epsilon)\Big)\Big] +$<br>$\mathbb{E}_{z \backsim p(z), \epsilon \backsim \mathrm{N}(0,I)}\Big[f_{\psi_2}(x_\theta(z,\xi), z) - log_{q_\phi}\big(z \mid x_\theta(z,\xi)\big)\Big]$ | The expectations are approximated via samples drawn from $q(x)$ and $p(z)$, as well as samples of $\epsilon$ and ξ |
| **MDGAN** [26] | Regularization with loss | $L_G = -\mathbb{E}_z\big[\log D\big(G(z)\big)\big] +$<br>$\mathbb{E}_{x \backsim p_d}[\lambda_1 d\big(x, G\ o\ E(x)\big) + \lambda_2 \log D\big(G\ o\ E(x)\big)$<br>$L_E = \mathbb{E}_{x \backsim p_d}[\lambda_1 d\big(x, G\ o\ E(x)\big) +$<br>$\lambda_2 \log D\big(G\ o\ E(x)\big)$ | G ∘ E is an autoencoder |
| **Dist-GAN** [30] | Latent- data distance constraint + discriminator-score distance constraint | $L_G = \Big\vert \mathbb{E}_x \sigma\big(D(x)\big) - \mathbb{E}_z \sigma\Big(D\big(G(z)\big)\Big)\Big\vert$<br>$L_D = -\Big(\mathbb{E}_x log\sigma\big(D(x)\big) - \mathbb{E}_z log\Big(1 -$<br>$\sigma\Big(D\big(G(z)\big)\Big)\Big) + \mathbb{E}_x log\sigma\Big(D\Big(G\big(E(x)\big)\Big)\Big) - L_p\Big)$<br>$L_p = \lambda_p \mathbb{E}_{\hat{x}}(\lVert \nabla_{\hat{x}} D(\hat{x}) \rVert_2^2 - 1)^2$ | Applied the gradient penalty $L_p$ for the $D_\gamma$ objective, where $\lambda_p$ is penalty coefficient, and $\hat{x} = \epsilon x + (1 - \epsilon)G(z)$, $\epsilon$ is a uniform random number $\epsilon \in U[0, 1]$ |
| **α-GAN** [32] | Hybrid loss | $\mathcal{L}(\theta,\eta) = \mathbb{E}_{q_\eta(Z \mid x)}\Big[-\lambda \lVert x - \mathcal{G}_\theta(z) \rVert_1 +$<br>$\log \frac{\mathcal{D}_\phi(\mathcal{G}_\theta(z))}{1 - \mathcal{D}_\phi(\mathcal{G}_\theta(z))} + log\ \frac{\mathcal{C}_\omega(z)}{1 - \mathcal{C}_\omega(z)}\Big]$ | $\mathcal{L}(\theta,\eta)$ is the hybrid loss function where $\mathcal{D}_\phi(x)$ is a classifier and $\mathcal{C}_\omega$ is a latent classifier |

Table 5 shows loss function of data space autoencoders. [26][31][32][75] autoencode data points. Due to this, the selection of a good loss function over natural images is a challenging task. In data-space autoencoders, only [30] has tried to handle both gradient vanishing and mode collapse problems in the GANs training, while remaining works have handled only the mode collapse problem. The combination of VAE and GANs provide promising results in term of handling mode collapse, but still it has not been explored much.

These research works are mainly focusing on the image generation. [31] has used a metric negative log-likelihood (NLL), estimated via the variational lower bound for natural images. [31] has evaluated proposed solution on both reconstruction and generation. [32] has used independent Wasserstein critic which calculates both mode collapse and overfitting in which if G remembers the training data, the critic trained on the validation set can differentiate between generated samples and data and if mode collapse strikes, critic can easily separate generated samples from the data.

### 5.3.2 Latent space Autoencoders

**Table 6. Loss function of latent space autoencoders**

| Model | Loss Type | Loss Function | Description |
|---|---|---|---|
| **ALI** [33] | -- | $\mathbb{E}_{q(x)}\left[\log\left(D(x, G_z(x))\right)\right] + \mathbb{E}_{p(x)}\left[\log\left(1 - D(G_x(z), z)\right)\right]$ | ALI's objective is to match the two joint distributions. |
| **BiGAN** [34] | -- | $\mathbb{E}_{q(x)}\left[\log\left(D(x, E(x))\right)\right] + \mathbb{E}_{p(x)}\left[\log\left(1 - D(G(z), z)\right)\right]$ | Optimize this minimax objective using the same alternating gradient based optimization as [74] |
| **HALI** [75] | -- | $L_D = \frac{1}{M}\sum_{i=1}^{M}\log\left({p_q}^{(i)}\right) - \frac{1}{M}\sum_{i=1}^{M}\log\left(1 - {p_p}^{(i)}\right)$<br>$L_G = \frac{1}{M}\sum_{i=1}^{M}\log\left(1 - {p_q}^{(i)}\right) - \frac{1}{M}\sum_{i=1}^{M}\log\left({p_p}^{(i)}\right)$ | $p_q$ is used to get D's predictions on E's distribution and $p_p$ is used to get D's predictions on D's distribution. |
| **AGE** [76] | -- | $\Delta\left(E(G(z))\right) \Vert E(x)$ | Generate a distribution G($z$) in data space that is close to the real data distribution $x$. |
| **VEEGAN** [35] | logistic regression loss | $-\mathbb{E}_{\gamma}\left[\log\left(D_\omega(z, x)\right)\right] - \mathbb{E}_{\theta}\left[\log\left(1 - D_\omega(z, x)\right)\right]$ | $E_\gamma$ is expectation w.r.to the joint distribution $q_\gamma(x \mid z)p(x)$ and $E_\theta$ w.r.to $p_\theta(z \mid x)p(x)$ |
| **MGGAN** [77] | -- | $\mathbb{E}_{x \sim p_{data}}\left[\log D_x(x) + \log D_m(E(x))\right] + \mathbb{E}_{z \sim p_z}\left[\log\left(1 - D_x(G(z))\right) + \log\left(1 - D_m\left(E(G(z))\right)\right)\right]$ | *Dm* is a discriminator for the guidance network and *m* means manifold space. Two Ds, *Dx* and *Dm*, do not explicitly affect each other, but both of them influence G |

**Further discussion.** Table 6 shows loss function of latent space autoencoders. In basic GANs, efficient inference mechanism does not exist. [33] proposed to learn generation and inference network jointly in an adversarial way. Moreover, [34] training has shown that *inverse* objective delivers more strong gradient signal to E and G and makes the training stable.

Most of the works in latent space autoencoder is focusing on the diversity of the generated images. [35] has used a new inference via optimization metric (IvOM) metric in which samples from the test data are compared to the nearest generated sample. If G undergoes mode collapse, then for some images distance is large.

## 5.4 Improved Discriminator

**Table 7. Loss function of Improved Discriminator based GANs**

| Model | Loss Type | Loss Function | Description |
|---|---|---|---|
| **EBGAN** [78] | Hinge loss | $L_D = D(x) + \left[m - D(G(z))\right]^+$<br>$L_G = D(G(z))$ | here [·]+ = max(0, ·) |
| **BEGAN** [79] | -- | $L_D = L(x) - k_t \alpha\left(G(z_D)\right)$<br>$L_G = L\left(G(z_G)\right)$ | $k_t \in [0, 1]$ to control $L\left(G(z_D)\right)$ during gradient descent |
| **MAGAN** [80] | Adaptive hinge loss | $L_D = D(x) + \max\left(0, m - D(G(z))\right)$<br>$L_G = D(G(z))$ | D($x$) is a deep auto-encoder function |
| **Max-Boost-GAN** [81] | Margin loss | $L_D = D(G(x)) + \left[\max\left(D(G(z_1)), D(G(z_2))\right) - m\right]^+$<br>$L_G = \max\left(D(G(z_1)), D(G(z_2))\right)$ | *m* is a positive margin |

**Further discussion.** Table 7 shows loss function of improved discriminator based GANs. Energy-based GANs has explored the GANs framework from the energy-based perspective. Energy-based GANs supposed to increase the quality and variety of generated images. Moreover, these GANs have shown good convergence pattern and scalability for generating higher-resolution images. However, in these solutions visual modes are not boosted. Energy-based GANs works for both image quality and diversity of the images. Apart from it, all current approaches of the energy-based GANs have mainly target the instability challenge. Energy-based GANs model has focused on the image generation application. Also, most of the works have used Inception Score (IS) as evaluation metric.

## 5.6 Latent Space Engineering

**Table 8. Loss function of latent space engineering based GANs**

| Model | Loss Type | Loss Function | Description |
|---|---|---|---|
| **DeliGAN** [38] | L2 regularization | $L_{GAN} = \mathbb{E}_{x \sim p_{data}}[\log D(x)] + \mathbb{E}_{z \sim p_z}\left[log\left(1 - D\left(G(z)\right)\right)\right]$ $\mathbb{E}_{z \sim p_z}\left[log\left(1 - D\left(G(z)\right)\right)\right] + \lambda \sum_{i=1}^{N} \frac{(1-\sigma_i)}{N}$ | $N$ is the number of Gaussian Component and $\sigma_i = 0.2$ |
| **NEMGAN** [36] | Categorical cross-entropy loss | $\mathrm{L}(G, h_1, h_2, D) = \mathbb{E}_{x \sim p_{data}}[log\ D(x)] +$ $\mathbb{E}_{z \sim p_z}[log\left(D\ o\ G(z)\right) + \|z - h_1\ o\ G(z)\|p] + D_{KL}(P_{\hat{Y}} \| P_Y)$ $L_{GAN} = \mathrm{L}(G, h_1, h_2, D) + \min_h L_{CC} + \min_\alpha D_{KL}\left(P_{\hat{Y}} \| \hat{P}_{\hat{Y}}\right)$ | The inversion network h2(h1(g(z))) inverts the generation process to ensure the matching of modal properties of generating and latent distributions. $\mathbb{E}_z[h(x)] = P_{\hat{Y}}$ |
| **DE-GAN** [82] | Hidden-space loss | $L_{GAN} = \mathbb{E}_{x \sim p_{data}}[\log D(x)] + \mathbb{E}_{z \sim p_z}\left[log\left(1 - D\left(G(z)\right)\right)\right]$ Hidden space loss $= \frac{1}{N} \sum_{i=1}^{N} \left\| h^i(X_{real}) - h^i\left(X_{gen}\right) \right\|$ | $h(X_{real})$ represents the activation map of a high convolutional layer when a real image is sent to D, while $h\left(X_{gen}\right)$ denotes a generated one |
| **InfoGAN** [37] | Information-theoretic regularization | $\mathbb{E}_{x \sim p_{data}}[\log D(x)] + \mathbb{E}_{z \sim p_z}\left[\log\left(1 - D\left(G(z)\right)\right)\right] -$ $\lambda I\left(c; G(z, c)\right)$ | -- |

**Further discussion.** Table 8 shows loss function of latent space engineering based GANs, i.e., noise-engineered GANs. They have primarily handled the issue of mode collapse while [36] has focused for the convergence of the training to make the model more robust. [37] proposed a solution for minimizing the mutual information between a subset *c* of the latent code and *x* by using an auxiliary distribution Q(*c*|*x*). However, InfoGAN does not support full inference on *z*, i.e., only the value for *c* is inferred.

Noise-engineered GANs models have tackled a variety of applications. Furthermore, [38] has introduced modified Inception Score (m-IS) by adding cross-entropy style score in the original IS metric. [36] introduced Frechet Classification Distance (FCD) to compute quality of generated images. Authors have also provided an analysis that a low value of FCD and high values of the clustering metrics (clustering accuracy (ACC), normalized mutual information (NMI) and adjusted rand index (ARI)) show that generated data is near to the real data and generated modes are matched with real data modes.

# 6 NEW LOSS FUNCTION

To highlight the contribution of solutions proposed under this section, we provide a summary including concerns raised in the paper, approaches used to handle those concerns, strengths and limitations of the proposed solution. We believe this summary will be useful for the future researchers to understand the progress of loss functions for GANs well. Table 9 and 10 show the summary of proposed probability distance and divergence, and regularization schemes for GANs, respectively.

**Table 9 Summary of Probability Distance and Divergence for GANs**

| Models | Concerns Raised | Approach | Strengths | Limitations |
| --- | --- | --- | --- | --- |
| **WGAN** [39] | Vanishing Gradient<br>Require D to learn 1-Lipschitz functions | Minimizes a reasonable and efficient approximation of the EM distance or Wasserstein-1<br>Enforce the Lipschitz constrained by weight clipping | A meaningful loss metric that correlates with the G's convergence and sample quality<br>An improved stability of the optimization process<br>It does not require a careful design of the network architecture either.<br>The mode dropping phenomenon that is typical in GANs is also drastically reduced | Unstable when gradients of the loss function are large<br>Slow training<br>Tuning weight clipping and hyperparameters is a tedious task<br>It do not consider to increase the variety of mode of generated visual samples and the variety of semantics of visual samples |
| **LS-GAN** [40] | G suffers from vanishing gradient as D can be optimized very quickly<br>Assuming infinite capacity for convergence which leads to mode collapse | Uses a Lipschitz constraint but reason is independent of WGAN's Lipschitz condition<br>Uses a weight-decay regularization technique<br>Loss having a data-dependent margin with gradients everywhere<br>Convergence proof without the assumption of infinite capacity. Generalization bounds | Due to Lipschitz regularity, LS-GAN can generalize well to produce new data from training examples | It does not increase the variety of mode of generated visual samples and the variety of semantics of visual samples |
| **RWGAN** [41] | Lack of robustness and efficiency during the learning process | Proposed a new class of statistical divergence, RW divergence, a combination of Bregman divergence and Wasserstein divergence<br>Asymmetric clipping | Due to Asymmetric clipping, avoids the low-quality samples and the failure of convergence<br>Wasserstein-L2 distance of order 2 improves the speed of convergence | -- |
| **f-GAN** [42] | Generalize GANs objective to variational divergence minimization | Minimizes the variational estimate of f-divergence<br>Simplify the saddle-point optimization [] and provide a theoretical justification | Learning objective is effective and computationally inexpensive than GANs | Use of Generative neural samplers is limited because after training they are unable to provide inferences |
| **b-GAN** [43] | GANs are sensitive to datasets, the form of the network and hyperparameters<br>In GANs, the value function derived from the two-player minimax game does not match the objective function | Objective function derived from the original motivation is not changed for learning the generative model<br>Learn a deep generative model from a density ratio estimation perspective<br>Pearson divergence | Improve the stability of GANs learning | -- |
| **χ2-GAN** [44] | Address the problem of simultaneous matching (SM) of multiple distributions | Optimizes the divergence between $\mu(x, y)$ and $\mu(x)\mu(y)$, which allows easy generalization beyond matching two distributions<br>Fully exploit the learned critic function | Stable at training and embraces sample diversity during generation<br>Formulation generalizes to problems requiring SM of multiple distributions | -- |

| Models | Concerns Raised | Approach | Strengths | Limitations |
|---|---|---|---|---|
| **LSGAN** [45] | Sigmoid cross entropy loss function may lead to the vanishing gradients problem<br>Basic GANs cause almost no loss for samples that lie in a long way on the correct side of the decision boundary | Adopted the least squares loss function for D<br>Penalizes the samples lying a long way to the decision boundary | LSGANs is more stable than original GANs during the learning process | Do not increase the variety of mode of generated visual samples and the variety of semantics of visual samples |
| **SoftmaxGAN** [46] | As the D utilizes a logistic loss, original GANs suffer from the vanishing gradients | Softmax cross-entropy loss<br>The target is to assign all probability to real data for D and to assign probability equally to all samples for G | The stability of GANs is improved through the usage of a Softmax cross-entropy loss in the sample space | Do not increase the variety of mode of generated visual samples and the variety of semantics of visual samples |
| **OT-GAN** [47] | In WGAN, it is not possible to optimize over all possible 1-Lipschitz functions leading to imperfect critic<br>Sinkhorn distance has biased sample gradients | A new distance metric Mini-batch energy distance does not require Lipschitz assumption<br>Mini-batch energy distance uses Sinkhorn distance along with Generalized energy distance hence has unbiased estimator | Mini-batch energy distance remains a valid training objective even when we stop training the critic | It requires large amounts of computation and memory |
| **IGAN** [7] | Overtraining of D<br>Mode collapse of G<br>Gradient descent may not converge<br>Vulnerable to adversarial examples<br>GAN outputs depend on the inputs | Present a variety of new architectural features and training procedures for GANs<br>• Feature Matching<br>• Mini-batch Discrimination<br>• Historical Averaging (Fictitious play)<br>• Label-smoothing<br>• Virtual Batch normalization | Feature matching's objective performs well for classification<br>Minibatch discrimination works well for generating realistic images<br>The historical average of the parameters can be updated in an online fashion so this learning rule scales well to long time series | Feature matching could not generate indistinguishable samples<br>Minibatch discrimination is computationally complex and highly sensitive to the selection of hyperparameters<br>VBN has high computational complexity |
| **McGAN** [48] | Impact of the distance choice on the stability of the optimization | Mean and covariance measure of distance for a critic function | Stable to train, have a reduced mode dropping and the IPM loss correlates with the quality of the generated samples | The use of clipping ends up restricting the capacity of the model<br>Requires matrix (tensor) decompositions which is hard to scale to higher order moment matching |
| **MMD GAN** [49] | GMMN is not efficient as GANs on challenging and large benchmark datasets<br>GMMN requires large batch size during the training | Introduce adversarial kernel learning techniques, as the replacement of a fixed Gaussian kernel in the original GMMN | New distance measure in MMD GAN is a meaningful loss that enjoys the advantage of weak topology and can be optimized via gradient descent with relatively small batch sizes | Computational complexity of MMDGAN increases as number of sample increases |
| **CramerGAN** [50] | Non-convergence of WGAN due to biased gradient estimator<br>Powerful critic is needed and also should not overfit the empirical distribution | Propose Cramer distance with unbiased sample gradients<br>Cramer distance enables learning without perfect critic | It measures energy distance indirectly in the data manifold but with a transformation function *h* | -- |

**Table 10 Summary of Regularization Schemes**

| Models | Concerns Raised | Approach | Strengths | Limitations |
|---|---|---|---|---|
| **WGAN-GP** [51] | Weight clipping in WGAN causes vanishing and exploding gradients and capacity underuse | WGAN except no weight clipping<br>Introduced a data dependent constraint namely a gradient penalty to enforce the Lipschitz constraint on the critic<br>Added a regularization term to encourage theoretical guarantee | Makes training more stable than WGAN and converge better<br>Produces high quality images | Gradient penalty adds computational complexity |
| **BWGAN** [52] | Extended WGAN-GP concept to any separable complete normed space | l2 norm is replaced with a dual norm<br>Generalized the WGAN-GP theory to Banach spaces to allow features selection for G | Introduced a generalization of WGANs with gradient norm penalization to Banach spaces, allowing to easily implement WGANs for a wide range of underlying norms on images | -- |
| **CT-GAN** [53] | WGAN-GP needs lot of iterations to ensure Lipschitz constraint | Add a regularization through Consistency term by perturbing the real data sample itself, twice. | Better photo-realistic samples than the previous methods<br>Achieves state-of-the-art semi-supervised learning results | -- |
| **SN-GAN** [12] | Weight clipping reduced the rank of the weight matrix<br>WGAN-GP introduces regularization based on unreliable model samples | Regularization which performs spectral normalization of weight matrix and does not affect the rank | Not dependent on model samples and less computationally complex | -- |
| [54] | Dimensionality misspecification<br>Variance due to noise | Adding high dimensional noise<br>Noise-induced regularization scheme | Regularization turns GAN models into reliable building blocks for deep learning | -- |
| **FisherGAN** [55] | Weight clipping reduces the capacity of D<br>WGAN-GP has high computational cost | Introduce data-dependent regularization which maintains the capacity of the critic while ensuring stability | It reduces the distance between two distributions as well as in-class variance<br>It does not add any weight clipping or gradient penalty | -- |
| [56] | Revisited the basic GANs algorithm for finding the Nash-equilibrium<br>Non-convergence of SGD | Identify the cause based on the Jacobian of gradients<br>Propose consensus optimization based on regularization w.r.to $\varphi$, $\theta$<br>Introduced a new design for the GANs training<br>Prove its convergence | Enables stable training of GANs on a variety of architectures and divergence measures | -- |
| **Unrolled GANs** [57] | Mode collapse as D cannot be trained till optimality at every iteration<br>G moves mass to a single point and D assigns lower probability to it | Introduce a surrogate loss which in limit equals the optimal D<br>G is updated based on the future update of D hence reducing mode collapse | Unrolling effectively increases the capacity of the D | Computational Complexity due to k-step D's updates<br>Only considered a small fraction of the design space |

| **Models** | **Concerns Raised** | **Approach** | **Strengths** | **Limitations** |
| --- | --- | --- | --- | --- |
| [58] | GAN is not convex-concave objective hence gradient descent may not converge<br>WGAN has non-convergent limit cycles<br>Local instability in GANs | Use ODE method to prove that GAN objective is locally asymptotically stable under certain conditions<br>Propose regularization on gradients of D for stability | Regularization term guarantee local stability for both the WGAN and the traditional GANs<br>Speeding up convergence and addressing mode collapse | -- |
| [59] | Non-convergence of unregularized GANs and WGAN-GP on non-overlapping manifolds | Noise induced regularization (Roth et al., 2017) converges<br>Propose simplified version of the above and prove convergence | (Unregularized) Gradient based GAN optimization is not always locally convergent.<br>WGANs and WGAN-GP do not always lead to local convergence whereas instance noise and zero-centered gradient penalties do | -- |
| **DRAGAN** [60] | Mode Collapse due to non-convex loss function<br>Non-convergence of alternative gradient descent<br>GANs learn swiss roll distribution despite vanishing gradients<br>WGAN-GP does not follow from KR duality as WGAN does | View the GANs optimization as regret minimization.<br>Prove convergence for the convex-concave case.<br>Converge to $\epsilon$- approximate equilibrium in non-convex case | Avoid the local equilibria which causes sharp D's gradients nearby some real data points | Do not improve diverse image generation with real datasets |